\documentclass[12pt]{gatechthesis}
\usepackage{wrapfig}
\usepackage{amsmath}
\usepackage{bm}
\usepackage{algorithmic}
\usepackage{amssymb}
\usepackage{amsfonts}
\usepackage{multirow}
\usepackage{makecell}
\usepackage[ruled,linesnumbered,vlined]{algorithm2e}

\title{Learning Control Policies for Fall prevention and safety in bipedal locomotion}
\author{Visak Kumar}
\approvaldate{September 7, 2021}
\school{School of Mechanical Engineering}
\department{Institute for Robotics and Intelligent Machines}
\bibliography{references}

\begin{document}

\makeTitlePage{December}{2021} 

\begin{approvalPage}{6}


\committeeMember{Dr. Karen Liu}{Computer Science Department}{Stanford University}
\committeeMember{Dr. Sehoon Ha}{School of Interactive Computing}{Georgia Institute of Technology}
\committeeMember{Dr. Gregory Turk}{School of Interactive Computing}{Georgia Institute of Technology}
\committeeMember{Dr. Gregory Sawicki}{School of Mechanical Engineering}{Georgia Institute of Technology}
\committeeMember{Dr. Ye Zhao}{School of Mechanical Engineering}{Georgia Institute of Technology}

\end{approvalPage}

\begin{frontmatter}
    
\begin{acknowledgments}

First, I would like to thank my parents and sister, Vijay Kumar, Gayithri and Apoorva, whose unconditional support and love have helped me immensely along this journey in graduate school.   

Next, I want to thank the members of my committee: Greg Turk, Gregory Sawicki and Ye Zhao. I have learned a lot through our interactions and greatly appreciate your valuable comments and suggestions.

I would be remiss to not express my gratitude towards my labmates and friends I have made along the way:  Wenhao Yu, Yifeng Jiang, Alex Clegg, Nitish Sontakke, Maksim Sorokin and many others. I had to endure completing the final components of this doctoral dissertation amidst a global pandemic. During this challenging time, without the additional support and inspiration from my dear friends Sophia, Aakash and Laura obtaining this degree would have been extremely challenging. I would also like to thank Stan Birchfield, Johnathan Tremblay and Katsu Yamane for valuable guidance during past internships.

Finally, I want to express my deepest gratitude to my advisors, Karen Liu and Sehoon ha. I am fortunate to have you both as my advisors and to have learned from you. Thank you for your patience and support as I navigated the ups and downs of this research journey.  

\end{acknowledgments}

    \makeTOC
    \makeListOfTables
    \makeListOfFigures
    \begin{summary}


The ability to recover from an unexpected external perturbation is a fundamental motor skill in bipedal locomotion. An effective response includes the ability to not just recover balance and maintain stability but also to fall in a safe manner when balance recovery is physically infeasible. For robots associated with bipedal locomotion, such as humanoid robots and assistive robotic devices that aid humans in walking, designing controllers which can provide this stability and safety can prevent damage to robots or prevent injury related medical costs. This is a challenging task because it involves generating highly dynamic motion for a high-dimensional, non-linear and under-actuated system with contacts. Despite prior advancements in using model-based and optimization methods, challenges such as requirement of extensive domain knowledge, relatively large computational time and limited robustness to changes in dynamics still make this an open problem.      

In this thesis, to address these issues we develop learning-based algorithms capable of synthesizing push recovery control policies for two different kinds of robots : Humanoid robots and assistive robotic devices that assist in bipedal locomotion. Our work can be branched into two closely related directions : 1) Learning safe falling and fall prevention strategies for humanoid robots and 2) Learning fall prevention strategies for humans using a robotic assistive devices. To achieve this, we introduce a set of Deep Reinforcement Learning (DRL) algorithms to learn control policies that improve safety while using these robots. To enable efficient learning, we present techniques to incorporate abstract dynamical models, curriculum learning and a novel method of building a graph of policies into the learning framework. We also propose an approach to create virtual human walking agents which exhibit similar gait characteristics to real-world human subjects, using which, we learn an assistive device controller to help virtual human return to steady state walking after an external push is applied.  

Finally, we extend our work on assistive devices and address the challenge of transferring a push-recovery policy to different individuals. As walking and recovery characteristics differ significantly between individuals, exoskeleton policies have to be fine-tuned for each person which is a tedious, time consuming and potentially unsafe process. We propose to solve this by posing it as a transfer learning problem, where a policy trained for one individual can adapt to another without fine tuning.

\end{summary}
\end{frontmatter}

\begin{thesisbody}
    \chapter{Introduction}

We often find ourselves startled by unexpected incidents like a push, slip or a trip while walking, yet we can successfully recover from these without having to put much thought into our actions. This ability to generate a recovery motion from external perturbations is  
one of the most important motor skills humans learn to prevent fall related injuries, these could be motion that either prevents a fall or enables us to fall in a safe manner. Studies have shown that we develop these abilities from a very young age, as infants transition from crawling into walking, an important step towards walking efficiently \cite{adolph2012you} is to learn how to fall. The quick reflexes and balance we develop over time help us avoid harmful injuries that can be sustained during a fall. 

Understanding the neuro-muscular control strategies that humans employ to achieve these exemplary motor skills has been a long standing quest in the fields of computer graphics, bio-mechanics and robotics. For bipedal robots, providing these abilities will be a step closer towards ensuring safety during operation. A safety layer in the control framework could open doors for these robots to work in unstructured outdoor environments or perhaps even enable learning more challenging dynamic locomotion skills like gymnastics or yoga. This would fulfill some of the initial motivations for designing bipedal robots such as being deployed in a disaster management scenario \cite{wang2007review} or perform labor intensive work such as delivery \cite{lee2016technological}. From a healthcare stand point, a deeper understanding of human motion can have lasting benefits for society. For example, methods used for helping disabled or injured people with gait rehabilitation can be vastly improved. In addition, better assistive devices can be designed for locomotion leading to improved the quality of life for many frail individuals.

Planning an effective response for a bipedal system is a challenging task because it requires generating highly dynamic motion for a high-dimensional system with unstable, nonlinear and discontinuous dynamics due to contacts. To alleviate some of these issues researchers have drawn inspiration from human strategies to 
design control algorithms. Prior work in this area are typically built on model-based and optimization techniques \cite{Stephens2007,Ha2015,Goswami2014,Mcgreavyetal} which require significant domain knowledge, large computational time and have limited robustness to changes in dynamics leaving plenty of opportunities for improvement. 


In this dissertation, we 
focus on developing learning based methods to teach robots safe locomotion skills. More specifically, we aim to teach effective push recovery response for two different kinds of robots 1) Humanoid robots and 2) Assistive robotic devices that aid in human walking. Our work with these two kinds of robots is along two closely related directions: 1) Learning safe falling and fall prevention strategies for bipedal robots and 2) Learning fall prevention strategies for humans using robotic assistive devices.    


As bipedal robots are increasingly being tested in unstructured real-world environment rather than in a laboratory setting, it is paramount that they have similar human-like ability of safe locomotion to prevent damages to the robot or to its immediate surroundings. In the first few chapters of this thesis we focus on developing a set of reinforcement learning based algorithms that enables bipedal robots learn these skills efficiently. While healthy adults have remarkable motor skills, people with disability or older adults have a reduced ability to prevent falls. According the Center for Disease Control (CDC) \cite{web2013centers}, over 3 million older adults are treated for fall related injuries resulting in millions of dollars in medical bills \cite{florence2018medical}. One in five falls result in head injury \cite{sterling2001geriatric,alexander1992cost} which makes falls the leading cause of traumatic brain injuries \cite{jager2000traumatic} among geriatric patients. In latter chapters of this thesis, we shift our focus towards investigating the effectiveness of reinforcement learning methods in learning control policies for assistive devices such as hip exoskeleton to prevent falls.

\section{Thesis overview}

This dissertation is organized as follows - In chapter 2 we provide a comprehensive overview of the most relevant prior work in designing control policies for humanoid robots and assistive devices. We highlight some prior reinforcement learning and transfer learning methods we draw inspiration from. Next, we present our work on designing safe falling and fall prevention control policies for bipedal robots in Chapter 3. We then describe our approach to design assistive device control policies for fall prevention in humans (Chapter 4). In Chapter 5, we present a novel method to enable transfer of policies from one human agent to another in a zero-shot manner. Finally, we conclude by summarizing our contributions and providing some exciting future directions this work can be extended to.

For bipedal robots, In our first work \cite{kumar2017learning} we addressed the challenge of optimizing a safe falling motion for a humanoid robot. Falling in a safe manner involves minimizing the impulse when a body part makes contact with a surface. Ideally, if a robot could turn itself into a ball during a fall, the resulting rolling motion would produce minimal impulse, similar to how gymnasts or martial artists fall when they lose balance. Since this has physical constraints for a robot we solve for two quantities, planning the next contacting body location as well as the corresponding joint motion to minimize the impulse of contact. We designed an actor-critic DRL algorithm that learns $n$ actors and critics simultaneously, where each critic corresponds to selecting the next contact location while the corresponding actor provides the necessary continuous joint motion in order to fall safely. We showed that using our approach we are able to solve for falling motion that generated lesser impulse on contact compared to prior approaches \cite{Ha2015} while significantly speeding up the required computational time. In our next work \cite{AdaptSample}, we developed a curriculum learning method to learn control policies for humanoid push recovery and balancing. Often solving a challenging task for high-dimensional robot requires a curriculum to enable efficient learning. During training, we maintain as estimate of the current push-magnitude the policy is capable of handling and provide pushes around those magnitudes. We showed that a policy learned using our approach outperforms other DRL baselines by handling larger push magnitudes.    
In \cite{expandmotor}, we introduce a novel framework to simplify learning complex tasks, such as 2D and 3D walking without falling, by building a directed graph of local control policies which automatically decomposes a complex task into sub-tasks. Using this approach, we show that control policies perform better while consuming lesser samples compared to naive DRL baselines.

For assistive devices, we introduce an algorithm in \cite{kumar2019learning} to learn a hip exoskeleton policy that helps a virtual human return to steady state walking after an external push is applied. To achieve this, we first model steady state human walking in simulation using DRL and open source bio-mechanical motion capture data. We demonstrate that the virtual human exhibits similar dynamical behaviour to real humans by comparing joint kinematics, kinetics \cite{Winter} and foot-step lengths to real world data. Next, we learn a recovery policy for exoskeleton to help the virtual human overcome external pushes applied to the hip. A thorough analysis is presented on the efficacy of the exoskeleton policy to help recovery and we also showed that our method can provide insights into the sensory and actuation capability required by exoskeletons to perform optimally. 

In our final work, we design an algorithm to address the issue of transferring the learned exoskeleton policy to different individuals. As walking gait characteristics differ significantly between individuals, exoskeleton policies have to be designed for each person. This process can be tedious and time consuming, we propose to solve this by posing it as a transfer learning problem. More specifically, we extend our prior work \cite{kumar2019learning} and take the following steps 1) Train multiple virtual human walking agents , each of them varying in physical characteristics like height, mass, etc. as well as ability to recover. 2) Learn an error function that predicts the difference in dynamics between them and 3) Train a policy which is explicitly aware of the difference enabling adaptation and efficient transfer.

\comment{
Balancing is an essential skill for bipedal robots to ensure safe operation by not just preventing damage to the robot itself but also preventing damages to environment around the robot, so naturally, it has been a topic of research interest for quite sometime. While there has been a lot of progress in   
designing controllers inspired by human balancing such as ankle and hip strategies \cite{Stephens2007,Aftab2012}, recent advancements in Deep Reinforcement Learning (DRL) \cite{schulman2015trust,schulman2017proximal,VanHasselt2007,Lillicrap2015} has shown promising results in overcoming 
some of the shortcomings of prior methods such as ability to handle high-dimensional state and action spaces as well solving under-actuated control problems with contacts, as is the case of bipedal locomotion.    

In this proposal, we leverage DRL approaches to teach robots fall prevention strategies and in the worst case scenario, fall in a safe manner.   
The proposed work primarily tackles two problems : 1) Strategies to learn safe falling and fall prevention for bipedal robots and 
2) Fall prevention for humans using assistive devices such as exoskeletons and approaches that could potentially enable transfer of policies among different individuals. As DRL methods often require a large sample budget to learn, physics based simulation plays a key role in our proposed work to generate data.

For bipedal robots, In our first work \cite{kumar2017learning} we introduced an algorithm to minimize damage while falling down. The algorithm leverages an abstract inverted pendulum model of the robot to plan a sequence of contacts to minimize impulse. Then in \cite{AdaptSample}, we present a curriculum learning method that aimed to improve model-based methods for humanoid push recovery and balancing. We show that our approach outperforms is capable of handling a wide range of pushes. Next \cite{expandmotor}, we introduce an algorithm to build a directed graph of local control policies which automatically decomposes a complex task into sub-tasks making learning easier. Using this approach, we show that control policies perform better for 2D and 3D walking, which requires balance, tasks compared to naive DRL baselines.

For assistive devices, we introduce an algorithm in \cite{kumar2019learning} to learn a hip exoskeleton policy that helps a virtual human return to steady state walking after an external push is applied. To achieve this, we first model steady state human walking in simulation using DRL and open source bio-mechanical \cite{} motion capture data. We demonstrate that the virtual human exhibits similar dynamical behaviour to real human by comparing joint kinematics, kinetics, ground reaction forces and foot-step lengths to real world data. Next, we learn a recovery policy for exoskeleton to help the virtual human overcome external pushes. A thorough analysis is presented on the efficacy of the exoskeleton policy to help recovery and we also showed that our method can provide insights into the sensory and actuation capability required by exoskeletons to perform optimally. 

In our last proposed work, we design an algorithm to address the issue of transferring the learned exoskeleton policy to different individuals. As walking gait characteristics differ significantly between individuals, exoskeleton policies have to be designed for each person. This process can be tedious and time consuming, we propose to solve this by posing it as a transfer learning problem. More specifically, we will take the following steps 1) Train multiple virtual human walking agents , each of them varying in physical characteristics like height, mass, etc. as well as ability to recover. 2) Learn a function that predicts the difference in dynamics between them and 3) Train a policy conditioned on this prediction to enable efficient transfer. 

}

    \chapter{Related work}

In this chapter, we give an overview of the most relevant prior work. Our contributions can be broadly categorised into two topics - First, algorithms to teach safe falling and balance recovery for humanoid robots and second, fall prevention using assistive devices for humans. In the first section, we  highlight prior work done at the intersection of human inspired control strategies for bipedal robots and reinforcement learning. Often, reinforcement learning suffers from the requirement of a large sample budget, we also present a summary of earlier work that addressed this shortcoming. Then, we provide an overview of common approaches to design control algorithms for assistive devices such as exoskeletons for humans. Finally, we outline prior work which focused on developing algorithms that enable transfer of control policies for robots from training to testing domains.

\section{Control of bipedal robots}

Researchers in the field of robotics, computer graphics and bio-mechanics have worked on designing control policies for bipedal robots for a long time. Due to the inherent instability of bipedal locomotion, falling and loss of balance are a common occurrence for these robots, naturally, researchers have spent extensive efforts on developing methods to maintain stability or fall safely to prevent damage. Due to the challenging nature of designing these controllers, a common approach that has worked well is to draw inspiration from human strategies by first understanding human motion and then developing control algorithms that replicate the desired behaviour. There is a vast body of research in this area ranging from deriving linear controllers from simple inverted pendulum dynamical models to more complex model predictive controllers. We highlight some relevant ones in this section.

\subsection{Balance recovery strategies}

As a response to external perturbations, humans employ multiple strategies to maintain balance. For example when pushed, bio=mechanical studies on postural balance control have shown that we regain balance by controlling the center of pressure on the foot using torques generated at the ankle and hip \cite{nashner1985organization,gordon1991analysis,kuo1993human}. Inspired by such strategies, \cite{Stephens2007,Stephens,macchietto2009momentum,kuo1995optimal} have derived control algorithms that are built strategies seen in humans. For larger perturbations, controllers that combine ankle, hip and foot-placement strategies have been proven successful in the humanoid robot community \cite{aftab2012ankle,Perrin2013,Komura2005,pratt2006capture,Pratt2012}. Many of these controllers rely on simplified dynamical models, such as Linear Inverted Pendulum (LIP) or 3D LIP, to compute control signals. While simplified models are beneficial for computational efficiency, they come at the cost of accuracy and inability to adapt to novel situations. Further, they also fail to capture arm motions that are sometimes crucial in maintaining balance. On the other hand, after the pioneering work of Mnih \etal~\cite{mnih2015human} deep reinforcement learning methods such as \cite{lillicrap2015continuous} and \cite{schulman2015trust} have shown promising results in learning effective control policies for high-dimensional systems. However, for challenging tasks, such as balance recovery, reinforcement learning methods can require a large sample budget to learn the skill. Our work \cite{AdaptSample} aims to combine the best of both worlds by building on top of these simplified models while leveraging the benefits of reinforcement learning to learn a control policy for a humanoid robot to recover from an external push. To reduce the required sample budget, we present a curriculum approach that simplifies learning process by providing external pushes of adequate difficulty during training.    

\subsection{Safe falling}
Recovery from an unstable state is important, but sometimes it is inevitable to prevent a fall, in such situations a controller which plans a safe falling motion is necessary to prevent damage to the robot and potentially to the surrounding environment as well. A plethora of motion planning and optimal control algorithms have been proposed to reduce the damage of humanoid falls. These algorithms have also been largely inspired by human strategies, for example rolling as one falls is an effective strategy. One direct approach to generating falling motion is to use knee and torso flexion to generate a squatting motion and backward to minimize impulse upon contact with the ground \cite{ma2014bio,hsiao1997common,robinovitch2004effect,tan2006wrist,robinovitch2000impact}, this strategy works best when the robot is falling backwards.
Another heuristic based approach is to design a few falling motion sequences for a set of expected scenarios. When a fall is detected, the sequence designed for the most similar falling scenario is executed \cite{Fujiwara2002,Ruiz-del-solar2010}. This approach, albeit simple, can be a practical solution in an environment in which the types of falls can be well anticipated. To handle more dynamic environments, a number of researchers cast the falling problem to an optimization which minimizes the damage of the fall. To accelerate the computation, various simplified models have been proposed to approximate falling motions, such as an inverted pendulum \cite{Fujiwara2006, Fujiwara2007}, a planar robot in the sagittal plane \cite{Wang2012}, a tripod \cite{Yun2014}, and a sequence of inverted pendulums \cite{Ha2015}. In spite of the effort to reduce the computation, most of the optimization-based techniques are still too slow for real-time applications, with the exception of the work done by \cite{Goswami2014}, who proposed to compute the optimal stepping location to change the falling direction. In contrast, our work takes the approach of policy learning using deep reinforcement learning techniques. Once trained, the policy is capable of handling various situations with real-time computation.

Our work is also built upon recent  advancement in deep reinforcement learning (DRL). Although the network architecture used in this work is not necessarily "deep", we borrow many key ideas from the DRL literature to enable training a large network with $278976$ variables. The ideas of "experience replay" and "target network" from Mnih \etal \cite{Mnih2015} are crucial to the efficiency and stability of our learning process, despite that the original work (DQN) is designed for learning Atari video games from pixels with the assumption that the action space is discrete. Lilicrap \etal \cite{Lillicrap2015} later combined the ideas of DQN and the deterministic policy gradient (DPG) \cite{Silver2014} to learn actor-critic networks for continuous action space and demonstrated that end-to-end (vision perception to actuation) policies for dynamic tasks can be efficiently trained. 

The approach of actor-critic learning has been around for many decades \cite{Sutton1988}. The main idea is to simultaneously learn the state-action value function (the Q-function) and the policy function, such that the intractable
optimization of the Q-function over continuous action space can be avoided. van Hasselt and Wiering introduced CACLA \cite{VanHasselt2007, VanHasselt2012} that represents an actor-critic approach using neural networks. Our work adopts the update scheme for the value function and the policy networks proposed in CACLA. Comparing to the recent work using actor-critic networks \cite{Lillicrap2015,Hausknecht2016}, the main difference of CACLA (and our work as well) lies in the update scheme for the actor networks. That is, CACLA updates the actor by matching the action samples while other methods follow the gradients of the accumulated reward function. Our work is mostly related to the MACE algorithm introduced by Peng \etal  \cite{Peng2016}. We adopt their network architecture and the learning algorithm but for a different purpose: instead of using multiple actor-critic pairs to switch between experts, we exploit this architecture to solve an MDP with a mixture of discrete and continuous action variables. 

\subsection{Overview of methods which address sample efficiency in reinforcement learning}
Recently, many successful model-free reinforcement learning (RL) algorithms have been developed which show remarkable promise in learning challenging motor skills \cite{Mnih2015,mnih2016asynchronous,schulman2015trust,schulman2017proximal,lillicrap2015continuous,Silver2014,VanHasselt2007,Hausknecht2016}. However, one of the biggest shortcomings of these methods is requirement of a large sample budget. Additionally, policies trained using RL are typically effective only from a small set of initial states defined during training. To address these issues, one approach researchers have taken is to cast a large-scale task as a hierarchical reinforcement learning algorithm.  This approach decomposes problems using \emph{temporal abstraction} which views sub-policies as macro actions, or \emph{state abstraction} which focuses on certain aspects of state variables relevant to the task. Prior work \cite{sutton1999between,daniel2012hierarchical,kulkarni2016hierarchical,heess2016learning,peng2017deeploco} that utilizes temporal abstraction applies the idea of parameterized goals and pseudo-rewards to train macro actions, as well as training a meta-controller to produce macro actions. One notable work that utilizes state abstraction is MAXQ value function decomposition which decomposes a large-scale MDP into a hierarchy of smaller MDP's \cite{Dietterich:2000:HRL,Bai:2012:OPL,Grave:2014:BEI}. MAXQ enables individual MDP's to only focus on a subset of state space, leading to better performing policies. Our relay networks can be viewed as a simple case of MAXQ in which the recursive subtasks, once invoked, will directly take the agent to the goal state of the original MDP. That is, in the case of relay networks, the Completion Function that computes the cumulative reward after the subtask is finished always returns zero. As such, our method avoids the need to represent or approximate the Completion Function, leading to an easier RL problem for continuous state and action spaces.

Yet another approach is to use a set of policies chained together to solve a challenging task. Tedrake \cite{tedrake2009lqr} proposed the LQR-Tree algorithm that combines locally valid linear quadratic regulator (LQR) controllers into a nonlinear feedback policy to cover a wider region of stability. 
Borno \etal \cite{borno2017domain} further improved the sampling efficiency of RRT trees \cite{tedrake2009lqr} by expanding the trees with progressively larger subsets of initial states. However, the success metric for the controllers is based on Euclidean distance in the state space, which can be inaccurate for high dimensional control problems with discontinuous dynamics. Konidaris \etal \cite{konidaris2009skill} proposed to train a chain of controllers with different initial state distributions. The rollouts terminate when they are sufficiently close to the initial states of the parent controller. They discovered an initiation set using a sampling method in a low dimensional state space. In contrast, our algorithm utilizes the value function of the parent controller to modify the reward function and define the terminal condition for policy training. There also exists a large body of work on scheduling existing controllers, such as controllers designed for taking simple steps \cite{coros2009robust} or tracking short trajectories \cite{liu2017learning}.
Inspired by the prior art, our work demonstrates that the idea of sequencing a set of local controllers can be Manipulating the initial state distribution during training has been considered a promising approach to accelerate the learning process. \cite{kakade2002approximately} studied theoretical foundation of using  ``exploratory'' restart distribution for improving the policy training.
Recently, \cite{popov2017data} demonstrated that the learning can be accelerated by taking the initial states from successful trajectories.
\cite{florensa17a} proposed a reverse curriculum learning algorithm for training a policy to reach a goal using a sequence of initial state distributions increasingly further away from the goal. We also train a policy with reversely-ordered initial states, but we exploited chaining mechanism of multiple controllers rather than training a single policy.realized by learning policies represented by the neural networks.

\section{Control of assistive devices}

\subsection{Simulation of human motion}
To study human motion during walking and recovery, biomechanics researchers often adopt an experimental approach. First, data is collected in the real-world, then control policies are synthesized using the real-world data. Winters et al \cite{Winter} was among the first in the field to study human gait, and the data published in this work remains relevant to this day. Wang et al \cite{Wang2014} and Hof et al \cite{hof2010balance} performed perturbation experiments and identified important relationships between COM velocity, step-lengths, center of pressure, stepping vs ankle strategy, etc.. We aim to leverage the finding of this research to validate some of the results. In Joshi et al \cite{Joshi2019}, a balance recovery controller was derived using the results reported in \cite{Wang2014}, however, they use a 3D Linear Inverted pendulum model to approximate the human dynamics. A 3D LIPD does not capture the dynamics fully, for example, angular momentum about the center of mass. Most relevant to our work, Antoine et al \cite{antoine2019}, used a direct-collocation trajectory optimization to synthesize a walking controller for a 3D musculo-skeletal model in OpenSim (OpenSim gait2392 model). The gait generated by the controller closely matched experimental data. The proposed method, relies on understanding the basic principles that lead to walking, such as minimizing metabolic cost, muscle activations,etc. However, the proposed solution enforces left-right symmetry, which works for walking, but is not ideal for disturbance recovery. Hence its unclear how well this approach will perform when there is an external disturbance to the human.

\subsection{Design of control algorithms}
Many researchers have developed control algorithms for robotic assistive walking devices. However, majority of the work can be classified into tracking controllers \cite{Wangpre2015,JezPre2004,BlayaPre} and model-based controllers \cite{WickModel,KazModel,vanderkooijModel}. 
There has been limited work at the intersection of DRL and control of assistive devices, especially for push recovery. Hamaya et al \cite{HAMAYA201767} presented model-based reinforcement learning algorithm to train a policy for a handheld device that takes muscular effort measured by electromyography signals (EMGs) as inputs. This method requires collecting user interaction data on the real-world device to build a model of the user's EMG patterns. However, collecting a large amount of data for lower-limb assistive devices is less practical due to the safety concerns.
Another recent work, Bingjing et al \cite{Bingjing2019}, developed an adaptive-admittance model-based control algorithm for a hip-knee assistive device, the reinforcement learning aspect of this work focused on learning parameters of the admittance model. 
Our method \cite{kumarexo} is agnostic to the assistive walking devices and can be used to augment any device that allows for feedback control.

\subsection{Transfer of RL policies}

A popular approach to transfer control policies is Domain randomization (DR). DR methods   \cite{openai2018learning,TobinFRSZA17,PintoDSG17,Pengdr,rajeswaran2017epopt} propose to train policies that are robust to variations in the parameters that affect the system dynamics. Although some of these methods have been validated in the real world \cite{openai2018learning,Pengdr}, DR often requires manual engineering of the range in which the parameters are varied to make sure that the true system model lies within the range of variation. For a complex robotic system, it is often challenging to estimate the correct range of all the parameters because a large range of variation could lead to lower task performance, whereas a smaller range leads to less robust policies. To address the demanding sample budget issue with domain randomization, \cite{muratore2021dataefficient} presented a data-efficient domain randomization algorithm based on bayesian optimization. The algorithm presented in Mehta et al \cite{mehta2019active} actively adapts the randomization range of variation to alleviate the need for exhaustive manual engineering. Ramos et al \cite{ramos2019bayessim} proposed an approach to infer the distribution of the dynamical parameters and showed that policies trained with randomization within this distribution can transfer better.

Careful identification of parameters using data from the real world, popularly known as system identification, has also shown promising results in real-world robots.
Tan et al \cite{Jieetal} and Hwangbo et al \cite{Hwangboeaau5872} carefully identified the actuator dynamics to bring the source environment closer to the target, Xie et al \cite{pmlr-v100-xie20a} also demonstrated that careful system identification techniques can transfer biped locomotion policies from simulation to real-world. Jegorova et al \cite{jegorova2020adversarial} presented a technique that improves on existing system identification techniques by borrowing ideas from generative adversarial networks (GAN) and showed improved ability to identify the parameters of a system. Similarly, Jiang et al \cite{jiang2021simgan} presented a SimGAN algorithm that identifies a hybrid physics simulator to match the simulated trajectories to the ones from the target domain to enable policy adaptation.
Yu et al \cite{yu2017preparing} developed a method that combines online system identification and universal policy to enable identifying dynamical parameters in an online fashion. Citing the difficulty in obtaining meaningful data for system identification, \cite{zhou2019environment} developed an algorithm that probes the target environment to provide more information about the dynamics of the environment. A few model based approaches have also been successful in transferring policies to a target domain \cite{song2020provably,desai2020stochastic,DataEffLeg}. 

Another popular approach of transferring policies includes utilizing data from the target domain to improve the policy. Chebotar et al \cite{chebotar2018closing} presented a method that interleaves policy learning and system identification, however this requires deploying the policy in the target domain every few iterations. This method would be impractical for a system that interacts closely with a human because of safety concerns. Yu et al \cite{yu2019simtoreal} and Peng et al \cite{RoboImitationPeng20} presented latent space adaptation techniques where the policy is adapted in the target domain by searching for a latent space input to the policy that enables successful transfer. Exarchos et al \cite{exarchos2020policy} also presented an algorithm that achieved policy transfer using only kinematic domain randomization combined with policy adaptation in the target domain, similar to \cite{yu2019simtoreal}. 

Yu et al~\cite{yu2020learning} proposed Meta Strategy Optimization, a meta-learning algorithm for training policies with latent variables that can quickly adapt to new scenarios with a handful of trials in the target environment.
Among the methods that use data from the target domain also include meta-learning approaches like  Bhelkale et al \cite{belkhale2020modelbased}, in which a model-based meta-reinforcement learning algorithm was presented to account for changing dynamics of an aerial vehicle carrying different payloads. In this approach, the parameters causing the variations in the dynamics are inferred by deploying the policy in the target domain, which in turn helps improve the policy's performance. In Ignasi et al. \cite{ignasi}, the idea of model-agnostic meta-learning \cite{finn2017modelagnostic} was extended to modelling dynamics of a robot. The authors presented an approach to quickly adapt the model of the robot in a new test environment while using a sampling-based controller MPPI to compute the actions. \cite{zeroshotdriving} developed a zero-shot transfer for policy by combining reinforcement learning and a robust tracking controller with a disturbance observer in the target environment. The validated the approach on a vehicle driving task. Similarly, \cite{fan2019bayesian,pmlr-v120-gahlawat20a} presented an approach to combine bayesian learning and adaptive control by learning model error and uncertainty.  

For tasks such as assistive device control for human locomotion, it is potentially unsafe and prohibitive to collect sufficient task-relevant data in the real world which prevents us from using methods such as system identification or transfer learning approaches that need data in the target environment. In addition to this, human dynamics exhibit large variations due to many unobserved parameters, this makes it challenging to define the right parameters for the system model in simulation and also in finding the right range of parameter variation for an approach like DR.

\subsection{Adaptation for Assistive Devices}

Assistive devices such as exoskeletons provide unique challenges for domain adaptation due to the large variations between individuals who pilot the device. Zhang et al \cite{Zhang1280} reported a human-in-the-loop optimization approach for ankle exoskeletons to account for this variability, however, this approach takes a few hours per individual to find the optimal control law. Jackson et al \cite{steve1} presented a unique heuristic-based approach to design a control law that adapts to the person's muscle activity. While these methods work well for steady-state walking, the large number of data required to optimize for in the case of \cite{Zhang1280} and the complex muscle responses involved during push recovery make it an infeasible application.
Several recent works have incorporated a learning-based approach to tackling the problem of adaptation, Peng et al \cite{peng2020data} adopted a reinforcement learning approach to learn assistive walking strategies for Hemiplegic patients, which was tested on real human patients and showed robustness and adaptability. However, it requires online data to update the actor-critic network. This process involves deploying a policy on a patient to collect data, for a task like push recovery it might be challenging to collect relevant data required for updating the policy without compromising the patient's safety. Both \cite{Huang2016AdaptiveIC} and \cite{DMPbasedRL} combined dynamic motion primitives (DMPs) and learning approaches to adapt control strategies for different individuals. Majority of the work with assistive devices have primarily focused on walk assistance and not on push-recovery.

    \chapter{Safe falling and fall prevention of Bipedal Robots}


\section{Learning control policies for safe falling }

\subsection{Motivation}
\begin{figure}[!ht]
\centering
\includegraphics[width=0.45\linewidth]{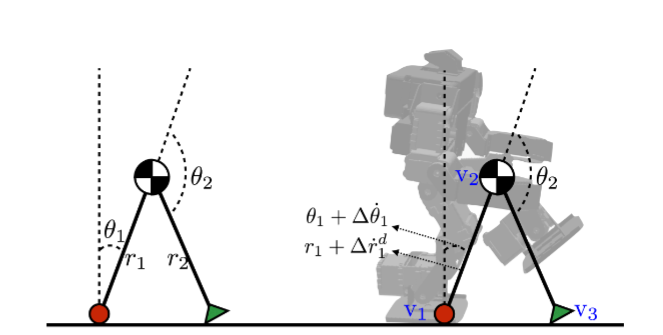}
\caption{Abstract model of the humanoid used in policy training.}
\label{AbstractFigure}
\end{figure}

As research efforts to push bipedal robots out of a laboratory setting and into the real world are increasing, the ability for robots to tackle unexpected situations in unstructured environments becomes crucial. For bipedal robots, due to the inherently unstable nature of dynamics, falling and losing balance are situations where being equipped with control policies that can prevent damages to the robot are paramount for success in the real world. Unlike walking , which is periodic in nature, falling is more challenging problem as it involves reasoning about both discrete and continuous quantities.

\begin{figure}[!ht]
\centering
\includegraphics[width=0.45\linewidth]{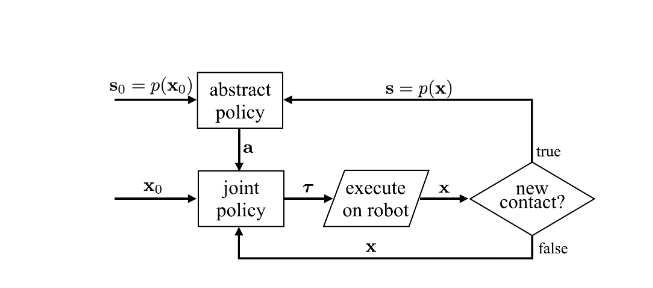}
\caption{Illustration of the method to compute control signal to execute a safe fall }
\label{Workflow}
\end{figure}

First, the policy must decide which body part should make contact with the ground, the location on the ground as well as the timing of the contact. The policy must also output the corresponding joint trajectory which minimizes impact upon contact. For example, humans employ a very effective strategy of rolling in order to reduce impact. Prior methods reported in generating walking motions which leverage the periodic nature of walking using finite-state machines (FSM) \cite{yin2007simbicon,hodgins1995animating,jain2009optimization,coros2010generalized} can be challenging to apply to the task of falling. Even approaches that mimic human motion using strong priors such as motion capture data \cite{da2008interactive,ye2010optimal,muico2009contact,sok2007simulating} are not entirely suitable due to lack of motion capture data of a wide variety of falls.

To this end, we developed a policy optimization approach \cite{kumar2017learning} for minimizing the damage to a humanoid robot during a fall.
In our approach, we first decide $n$ candidate contacting body part such as hands, feet, or knees designed to be a contact point with the ground. Our algorithm trains $n$ control policies (actors) and the corresponding value functions (critics) in a single interconnected neural network. Each policy and its corresponding value function are associated with a candidate contacting body part. During policy execution, the network is queried every time the robot establishes a new contact with the ground, allowing the robot to re-plan throughout the fall. Based on the current state of the robot, the actor corresponding to the critic with the highest value will be executed while the associated body part will be the next contact with the ground. With this mixture of actor-critic architecture, the discrete contact sequence planning is solved through the selection of the best critics while the continuous control problem is solved by the optimization of actors. To simplify learning , the policy uses an abstract model representation of a humanoid robot, shown in \ref{AbstractFigure}, consisting of an inverted pendulum and a telescopic rod to plan actions which are then mapped back to a full joint space using inverse kinematics.  
\ref{Workflow} illustrates the sequence of operations involved in planning a safe fall.
We use CACLA policy learning algorithm \cite{VanHasselt2007, VanHasselt2012} to optimize the control policy. We show that our proposed approach is able to generate motions that lead to less impact while falling down compared to prior approaches \cite{Ha2015} while also significantly speeding up the computational time (~50 to 400 times faster).

\subsection{Mixture of actor-critic experts}

\ref{Workflow} illustrates the workflow of the overall policy. We first define a function $p: \mathcal{X} \mapsto \mathcal{S}$ which maps a state of robot ($\vc{x} \in \mathcal{X}$) to a state of abstract model ($\vc{s} \in \mathcal{S}$). The mapping can be easily done because the full set of joint position and velocity contains all the necessary information to compute $\vc{s}$. As the robot receives an external force initially, the state of the robot is projected to $\mathcal{S}$ and fed into the abstract-level policy. The action ($\vc{a}$) computed by the abstract-level policy is passed into the joint-level policy to compute the corresponding joint torques ($\boldsymbol{\tau}$). If no new contact is detected after executing $\boldsymbol{\tau}$, the new state of the robot will be fed back into the joint-level policy and a new $\boldsymbol{\tau}$ will be computed (the lower feedback loop in Figure \ref{Workflow}). If the robot encounters a new contact, we re-evaluate the contact plan by querying the abstract-level policy again  (the upper feedback loop in \ref{Workflow}).


\begin{figure}[!t]
\centering
\includegraphics[width=0.4\columnwidth]{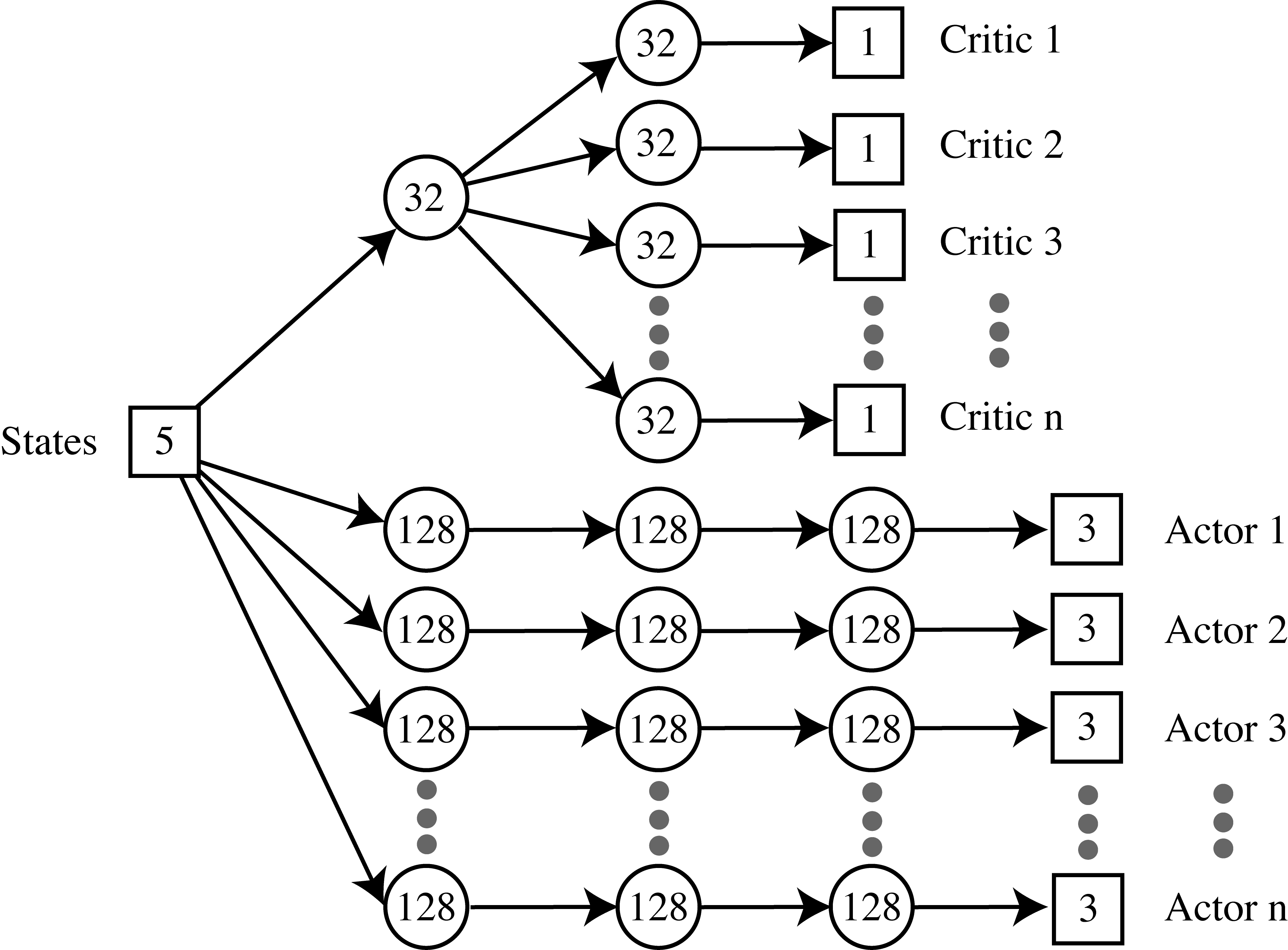}
\caption{A schematic illustration of our deep neural network that consists of $n$ actor-critics. The numbers indicate the number of neurons in each layer.}
\label{fig_NN}
\end{figure}

\subsubsection{Abstract-level Policy}
To overcome the challenge of optimizing over both discrete and continuous action variables, we introduce a new policy representation based on a neural network architecture inspired by MACE \cite{Peng2016}. The policy takes as input the state of the abstract model and outputs the action, as well as the next contacting body part. The state space is defined as $\vc{s} = \{c_1, r_1, \theta_1, \dot{r}_1, \dot{\theta}_1\}$, where the total elapsed time $t$ is removed from the state space previously defined by Ha and Liu \cite{Ha2015}. As a side benefit of a feedback policy, we no longer need to keep track of the total elapsed time. For the action space, we remove the discrete action variable, $c_2$ so that the new action is a continuous vector in $\mathbb{R}^3$: $\vc{a} = \{\theta_2, \Delta, \dot{r}_1^d\}$. The network combines $n$ pairs of actor-critic subnets, each of which is associated with a contacting body part. Each actor, $\Pi_i(\vc{s}): \mathbb{R}^5 \mapsto \mathbb{R}^3$, represents a control policy given that the $i$-th contacting body part is chosen to be the next contact. Each critic, $V_i(\vc{s}): \mathbb{R}^5 \mapsto \mathbb{R}$, represents the value function that evaluates the return (long-term reward) of using the $i$-th contacting body part as the next contact and taking the action computed by $\Pi_i(\vc{s})$ as the next action. We fuse all $n$ actor-critic pairs in one single network with a shared input layer (\ref{fig_NN}).

At each impact moment when a new contact is established, the network evaluates all the $V_i(\vc{s}), \;1 \leq i \leq n$, and selects the policy corresponding to the highest critic. This arrangement allows us to train $n$ experts, each specializes to control the robot when a particular contacting body part is selected to be the next contact. As a result, we cast the discrete contact planning into the problem of expert selection, while simultaneously optimizing the policy in continuous space.

\subsubsection{Reward}
Since our goal is to minimize the maximal impulse, we define the reward function as:
\begin{equation}
r(\vc{s}, \vc{a}) = \frac{1}{1 + j}, 
\end{equation}
where $j$ is the impulse induced by the contact. Suppose the COM of the pendulum and the tip of the stopper are $(x_1, y_1)$ and $(x_2, y_2)$ respectively at the impact moment, the impulse can be computed as: 
\begin{equation}
j = -\frac{\dot{y}_2^-}{\frac{1}{M} + \frac{1}{I}(x_2 - x_1)^2},
\end{equation}
where $M$ is the mass and $I$ is the inertia of the abstract model (see details in \cite{Ha2015}). 

With this definition of the reward function, the objective of the optimization is to maximize the minimal reward during the fall.

\subsubsection{Learning Algorithm}
Algorithm 1 illustrates the process of learning the abstract-level policy. We represent the policy using a neural network consisting $n$ pairs of actor-critic subnets with a shared input layer (\ref{fig_NN}). Each critic has two hidden layers with 32 neurons each. The first hidden layer is shared among all the critics. Each actor has 3 hidden layers with 128 neurons each. All the hidden layers use tanh as the activation functions. We define weights and biases of the network as $\theta$, which is the unknown vector we attempt to learn.

\begin{algorithm} 
\label{alg1}
\caption{Learning abstract-level policy}
\begin{algorithmic}[1]                    
\STATE Randomly initialize $\theta$ 
\STATE Initialize training buffer with tuples from DP 
\WHILE{not done}
\STATE EXPLORATION:
\FOR{$k=1 \cdots K$}
\STATE $\vc{s} \sim \mathcal{N}_0$ 
\WHILE{$\vc{s}.\dot{\theta}_1 \geq 0$}
\STATE c $\leftarrow$  Select actor stochastically using Equation \ref{boltzman}
\STATE \vc{a} $\leftarrow \Pi_{c}(\vc{s}) + \mathcal{N}_{t}$
\STATE Apply $\vc{a}$ and simulate until next impact moment
\STATE $\vc{s}' \leftarrow$ Current state of abstract model 
\STATE $r \leftarrow r(\vc{s}, \vc{a})$
\STATE Add tuple $\tau \leftarrow (\vc{s}, \vc{a}, \vc{s}', r, c)$ in training buffer
\STATE $\vc{s} \leftarrow \vc{s}'$
\ENDWHILE
\ENDFOR
\newline
\STATE UPDATE CRITIC: 
\STATE Sample a minibatch $m$ tuples $\{ \tau_{i} = (\vc{s}_{i}, \vc{a}_i, \vc{s}'_{i}, r_i, c_{i})\}$ 
\STATE $y_{i}$ $\leftarrow$ $\min(r_{i},\gamma \max_j$ $\hat{V}_{j}(\vc{s}'_{i}))$ for each $\tau_{i}$   
\STATE $\theta \leftarrow \theta + \alpha \sum_i (y_i -  V_{c_i}(\vc{s})) \nabla_\theta V_{c_i}(\vc{s})$
\newline
\STATE UPDATE ACTOR: 
\STATE Sample a minibatch $m$ tuples $\{ \tau_{i} = (\vc{s}_{i}, \vc{a}_i, \vc{s}'_{i}, r_i, c_{i})\}$
\STATE $y_{i} = \max_j V_j(\vc{s})$ 
\STATE $y'_{i} \leftarrow \min(r_i,\gamma \max_j \hat{V}_{j}(\vc{s}'_{i}))$ 
\IF {$y'_{i} > y_{i}$}
\STATE $\theta \leftarrow \theta  + \alpha (\nabla_\theta \Pi_{c_i}(\vc{s}))^T(\vc{a}_i - \Pi_{c_i}(\vc{s}))$

\ENDIF     
\ENDWHILE
\end{algorithmic}
\end{algorithm}
The algorithm starts off with generating an initial set of high-reward experiences, each of which is represented as a tuple: $\tau = (\vc{s}, \vc{a}, \vc{s}', r, c)$, where the parameters are the starting state, action, next state, reward, and the next contacting body part. To ensure that these tuples have high reward, we use the dynamic-programming-based algorithm described in \cite{Ha2015} (referred as DP thereafter) to simulate a large amount of rollouts from various initial states and collect tuples. Filling the training buffer with these high-reward experiences accelerates the learning process significantly. In addition, the high-reward tuples generated by DP can guide the abstract-level policy to learn "achievable actions" when executed on a full body robot. Without the guidance of these tuples, the network might learn actions that increase the return but unachievable under robot's kinematic constraints. 


In addition to the initial experiences, the learning algorithm continues to explore the action space and collect new experiences during the course of learning. At each iteration, we simulate $K (=10)$ rollouts starting at a random initial state sampled from a Gaussian distribution $\mathcal{N}_0$ and terminating when the abstract model comes to a halt. A new tuple is generated whenever the abstract model establishes a new contact with the ground. The exploration is done by stochastically selecting the critic and adding noise in the chosen actor. We follow the same Boltzmann exploration scheme as in \cite{Peng2016} to select the actor $\Pi_i(\vc{s})$ based on the probability defined by the predicted output of the critics:
\begin{equation}
\label{boltzman}
\mathcal{P}_i(\vc{s}) = \frac{e^{V_i(\vc{s}|\theta)/T_t}}{\sum_j e^{V_j(\vc{s}|\theta)/T_t}},
\end{equation}
where $T_t (=5)$ is the temperature parameter, decreasing linearly to zero in the first $250$ iterations. While the actor corresponding to the critic with the highest value is most likely to be chosen, the learning algorithm occasionally explores other actor-critic pairs, essentially trying other possible contacting body parts to be the next contact. Once an actor is selected, we add a zero-mean Gaussian noise to the output of the actor. The covariance of the Gaussian is a user-defined parameter.

After $K$ rollouts are simulated and tuples are collected, the algorithm proceeds to update the critics and actors. In critic update, a minibatch is first sampled from the training buffer with $m (=32)$ tuples, $\tau_i = (\vc{s}_i, \vc{a}_i, \vc{s}'_i, r_i, c_i)$. We use the temporal difference to update the chosen critic similar to \cite{Mnih2015}:
\begin{eqnarray}
&y_i = \min(r_i, \gamma \max_j \hat{V}_j(\vc{s}'_i)) \\ \nonumber
&\theta \leftarrow \theta + \alpha \sum_i (y_i -  V_{c_i}(\vc{s})) \nabla_\theta V_{c_i}(\vc{s})
\end{eqnarray}
where $\theta$ is updated by following the negative gradient of the loss function $\sum_i (y_i - V_{c_i}(\vc{s}))^2$ with the learning rate $\alpha (=0.0001)$. The discount factor is set to $\gamma = 0.9$. Note that we also adopt the idea of target network from \cite{Mnih2015} to compute the target, $y_i$, for the critic update. We denote the target networks as $\hat{V}(\vc{s})$.

The actor update is based on supervised learning where the policy is optimized to best match the experiences: $\min_\theta \sum_i \|\vc{a}_i - \Pi_{c_i}(\vc{s}_i)\|^2$. We use the positive temporal difference to decide whether matching a particular tuple is advantageous:
\begin{eqnarray}
&y = \max_j V_j(\vc{s}_i) \\ \nonumber
&y' = \min(r_i, \gamma \max_j \hat{V}_j(\vc{s}'_i)) \\ \nonumber
&\mathrm{if\;\;} y' > y, \;\;\;\theta \leftarrow \theta  + \alpha (\nabla_\theta \Pi_{c_i}(\vc{s}))^T(\vc{a}_i - \Pi_{c_i}(\vc{s})).
\end{eqnarray}

\subsection{Results}

\begin{figure}
\centering
\includegraphics[width=0.5\columnwidth]{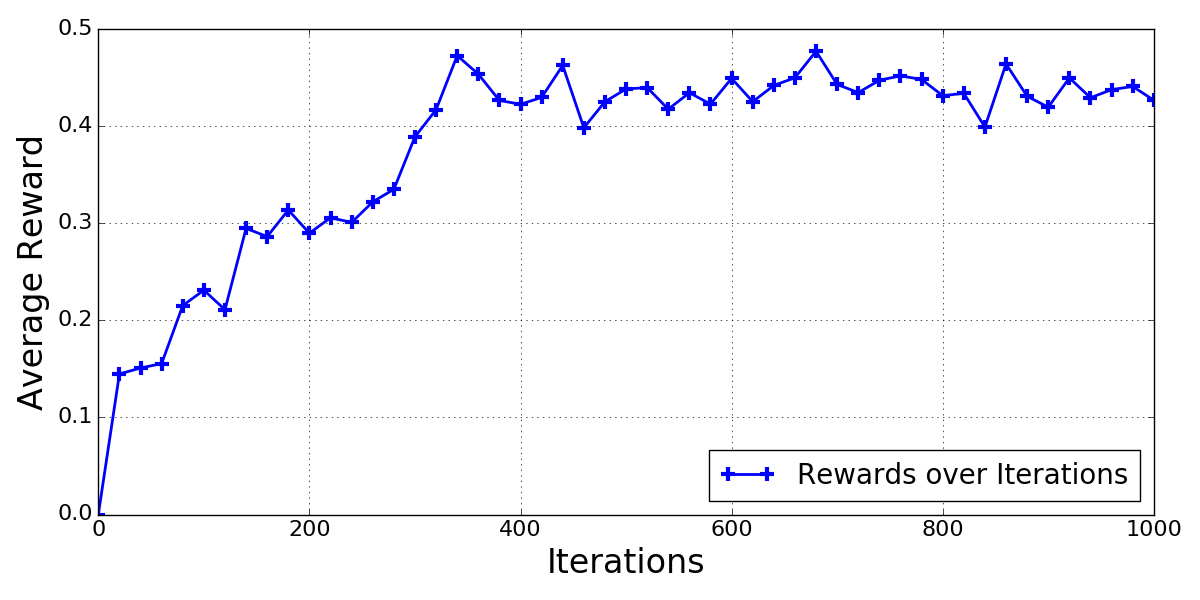}
\caption{The average reward for $10$ test cases.}
\label{fig:reward_iter}
\end{figure}

We validate our policy in both simulation and on physical hardware. The testing platform is a small humanoid, BioloidGP \cite{BioloidGP} with a height of $34.6$~cm, a weight of $1.6$~kg, and 16 actuated degrees of freedom. We also compare the results from our policy against those calculated by the dynamic-programming (DP) based method proposed by Ha and Liu \cite{Ha2015}. Because DP conducts a full search in the action space online, which is 50 to 400 times slower than our policy, in theory DP should produce better solutions than ours. Thus, the goal of the comparison is to show that our policy produces comparable solutions as DP while enjoying the speed gain by two orders of magnitude.


We implement and train the proposed network architecture using PyCaffe \cite{jia2014caffe} on Ubuntu Linux. For simulation, we use a Python binding \cite{PyDart} of an open source physics engine, DART \cite{Dart}.

\subsubsection{Learning of Abstract-Level Policy}
In our experiment, we construct a network with $8$ pairs of actor-critic to represent $8$ possible contacting body parts: right toe, right heel, left toe, left heel, knees, elbows, hands, and head. During training, we first generate $5000$ tuples from DP to initialize the training buffer. The learning process takes $1000$ iterations, approximately $4$ hours on $6$ cores of 3.3GHz Intel i7 processor. \figref{reward_iter} shows the average reward of $10$ randomly selected test cases over iterations. Once the policy is trained, a single query of the policy network takes approximately $0.8$~milliseconds followed by $25$~milliseconds of the inverse kinematics routine. The total of $25.8$~milliseconds computation time is a drastic improvement from DP which takes $1$ to $10$ seconds of computation time.


\begin{figure}
\centering
\includegraphics[width=0.5\columnwidth]{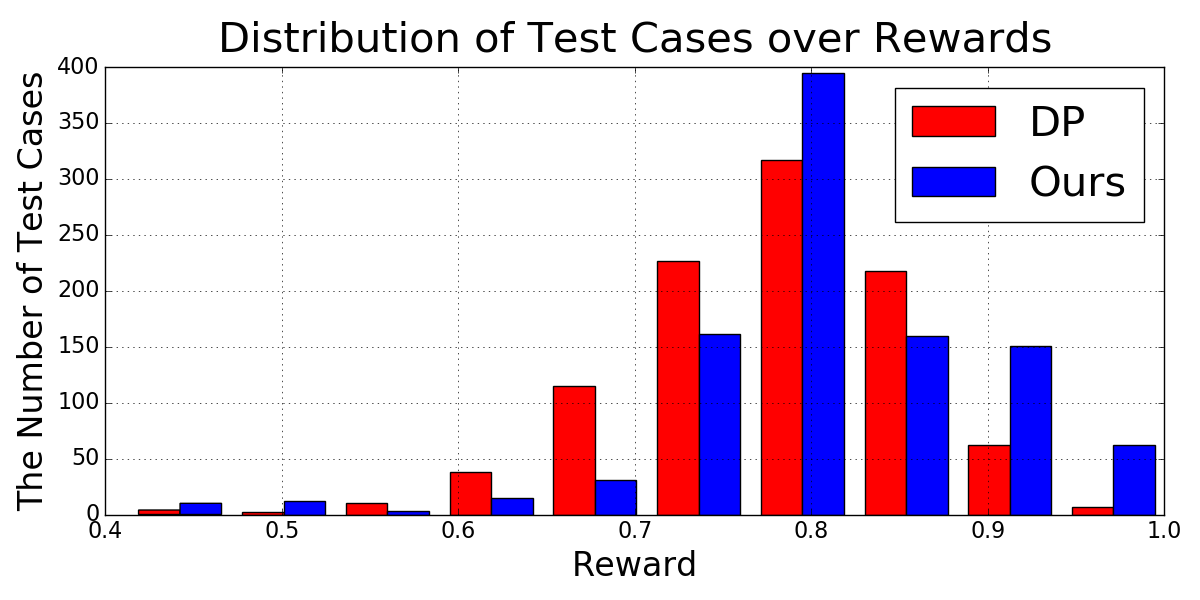}
\caption{The histogram of rewards for the 1000 test cases. Our policy outperforms DP in $65\%$ of the tests.}
\label{fig:reward_distribution}
\end{figure}

\begin{figure*}
\center
\setlength{\tabcolsep}{1pt}
\renewcommand{\arraystretch}{0.5}
\begin{tabular}{c c c c c}
\includegraphics[width=0.195\textwidth]{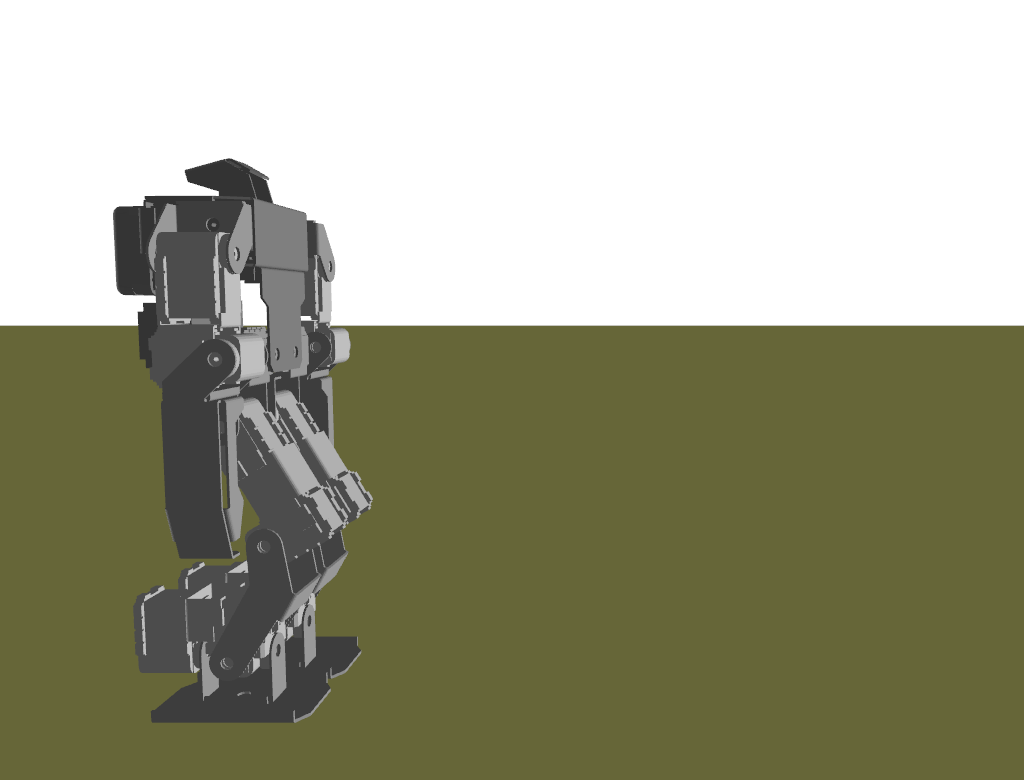}&
\includegraphics[width=0.195\textwidth]{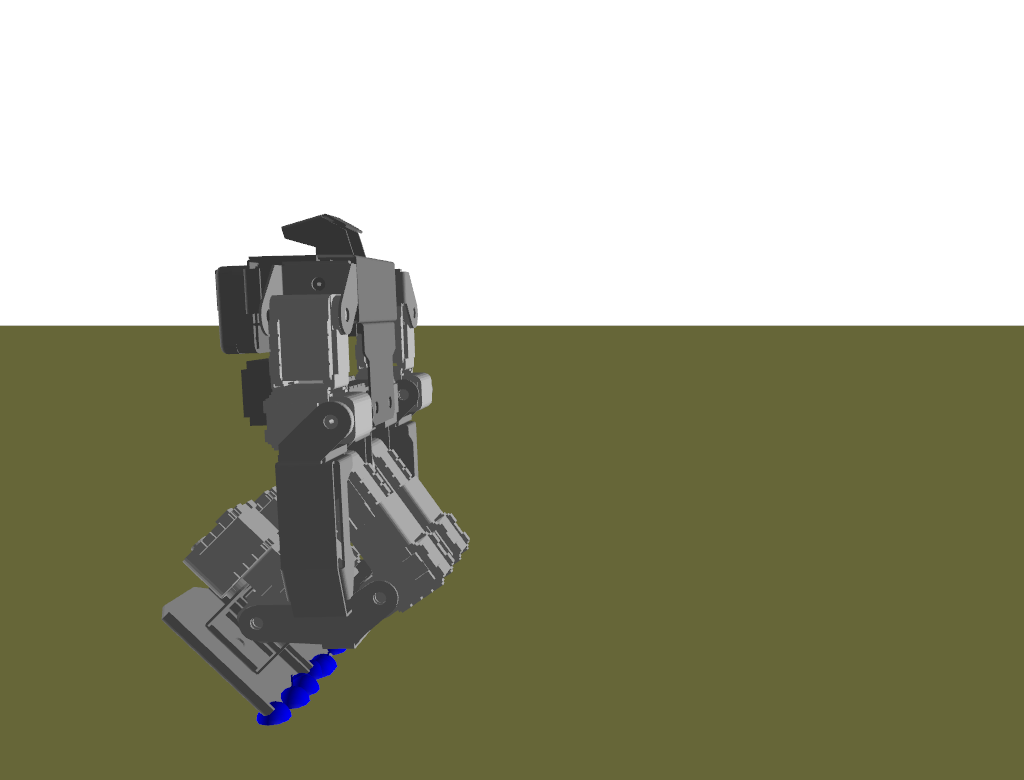}&
\includegraphics[width=0.195\textwidth]{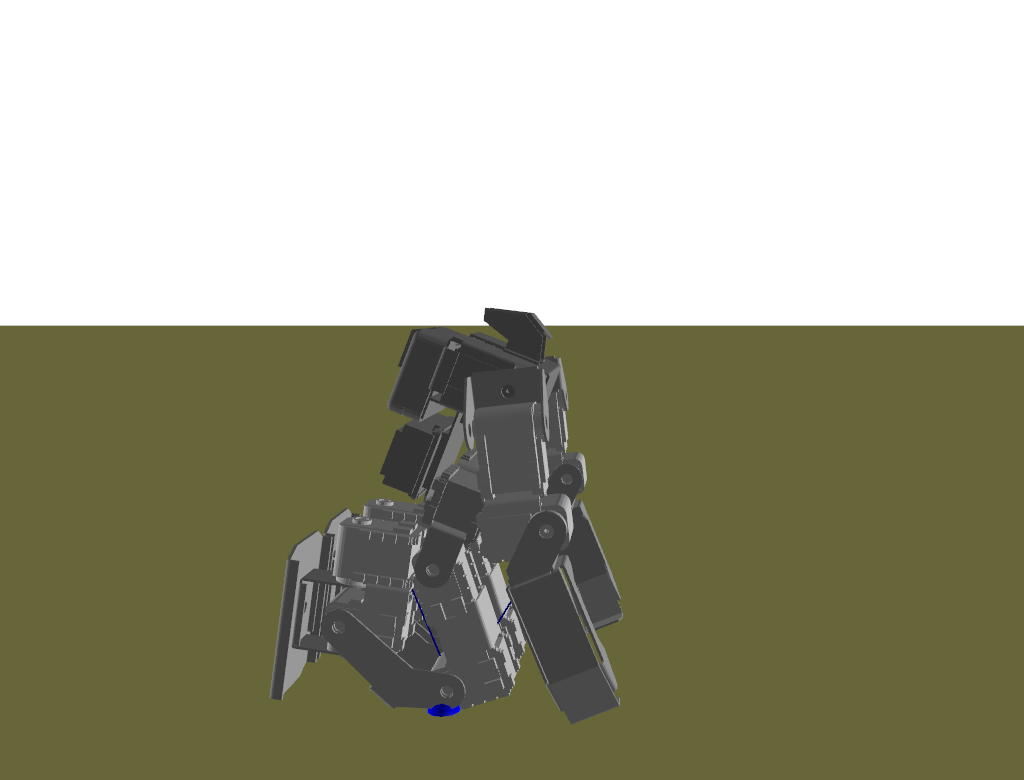}&
\includegraphics[width=0.195\textwidth]{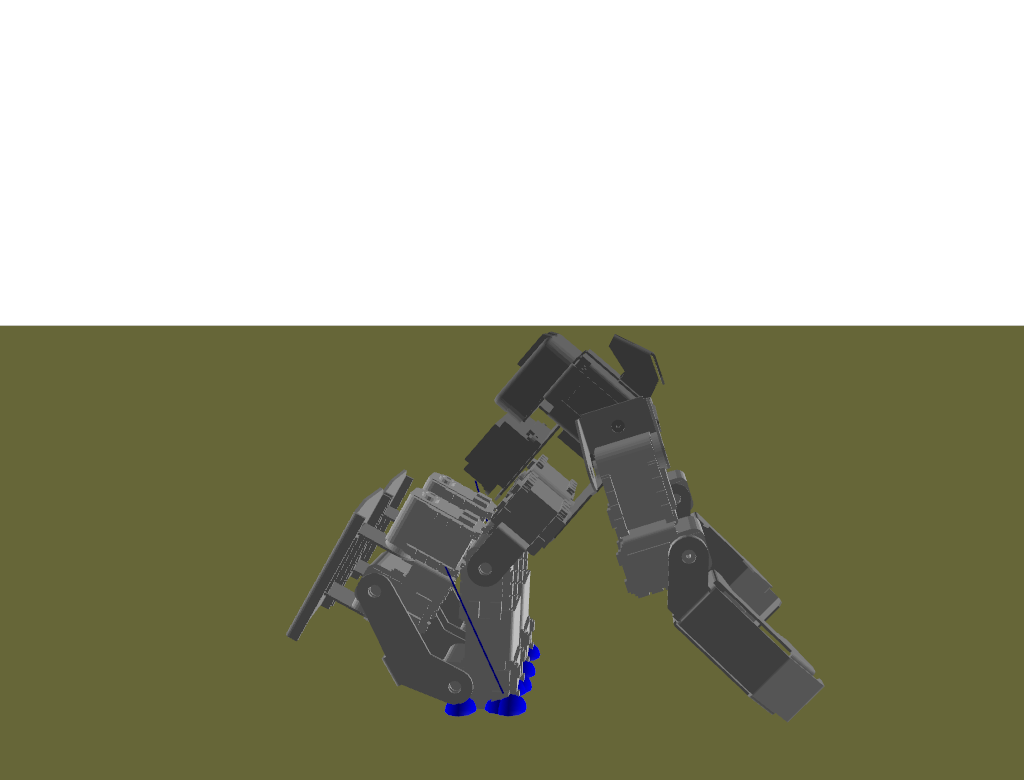}&
\includegraphics[width=0.195\textwidth]{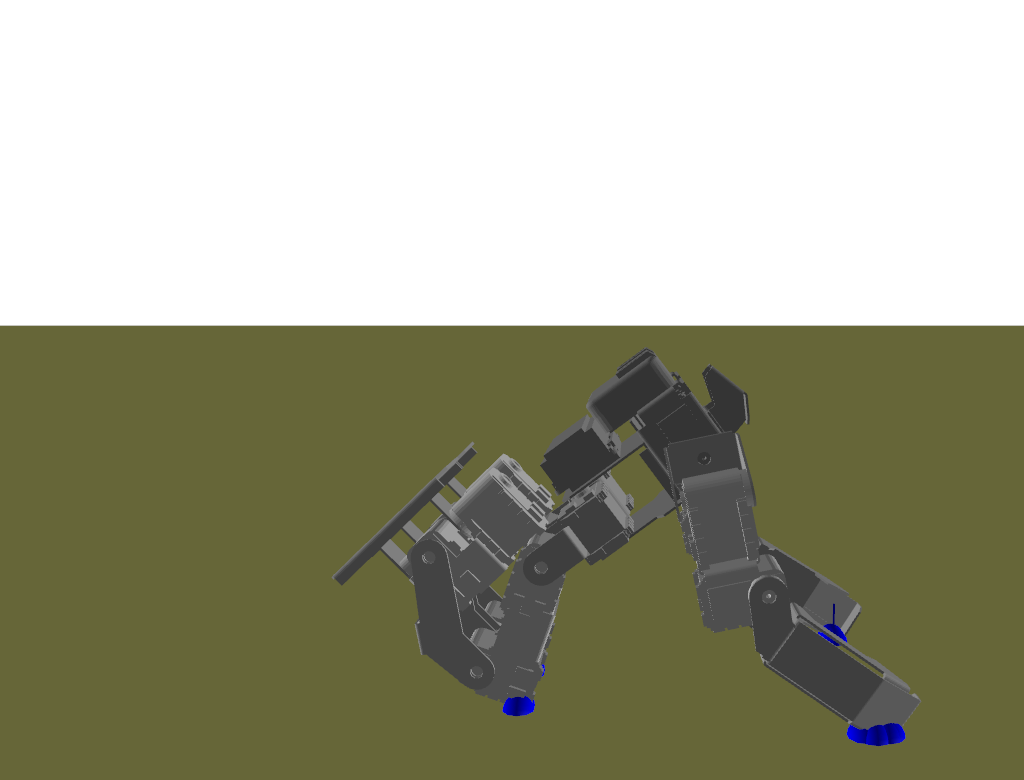} \\
\includegraphics[width=0.195\textwidth]{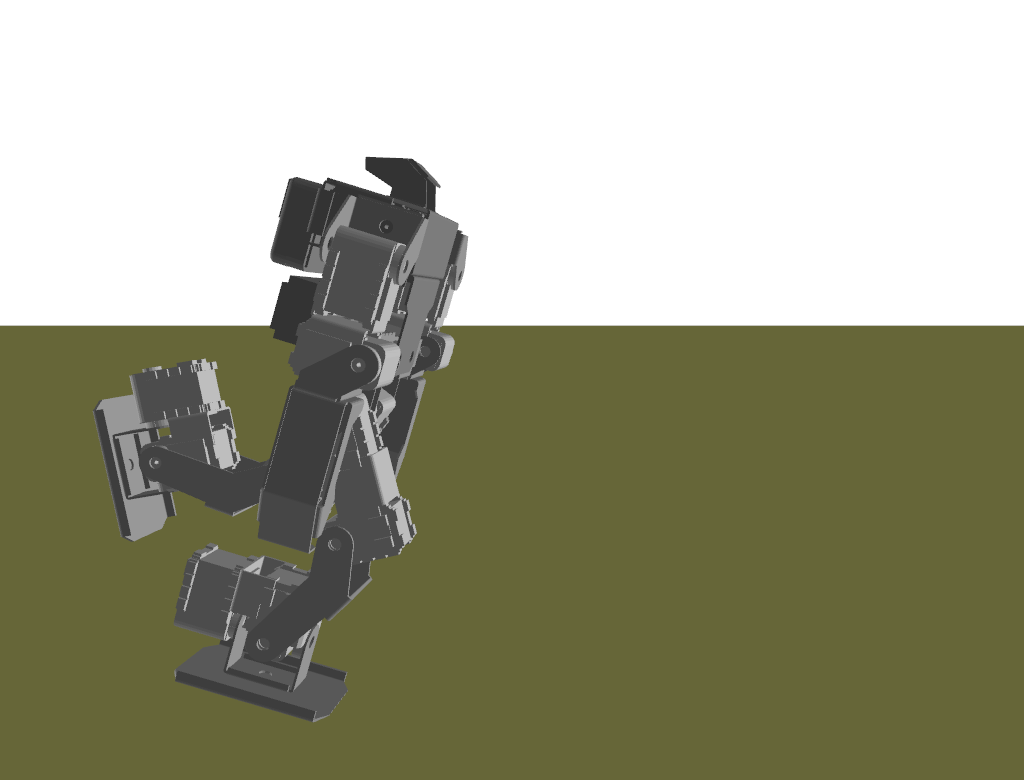}&
\includegraphics[width=0.195\textwidth]{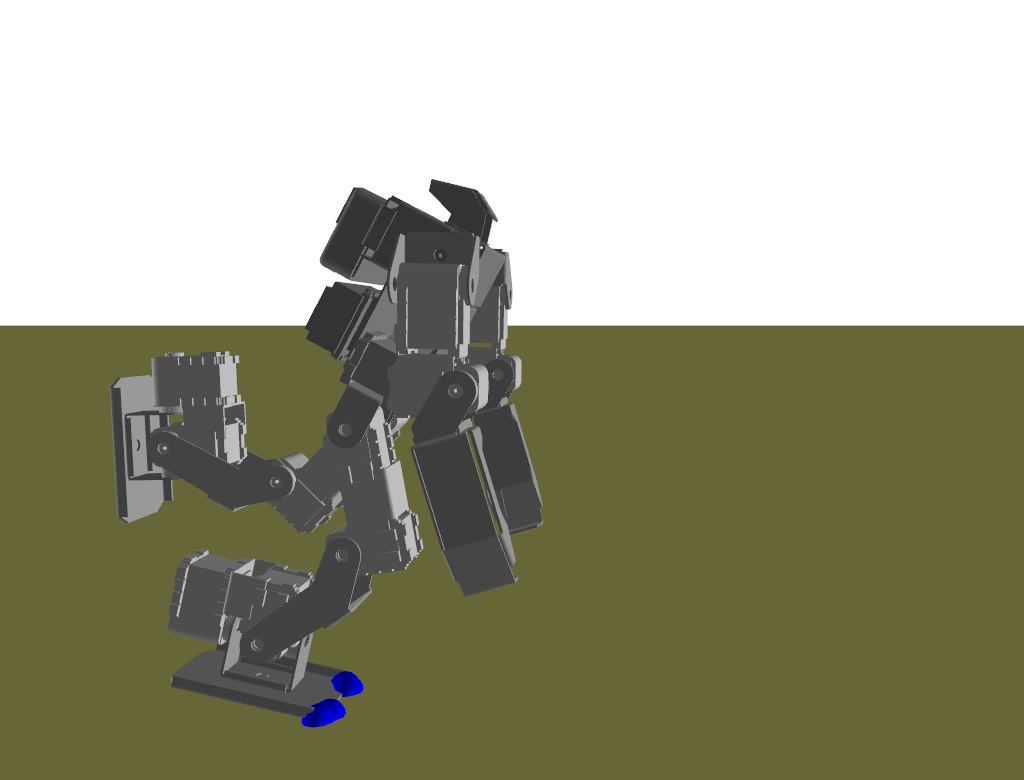}&
\includegraphics[width=0.195\textwidth]{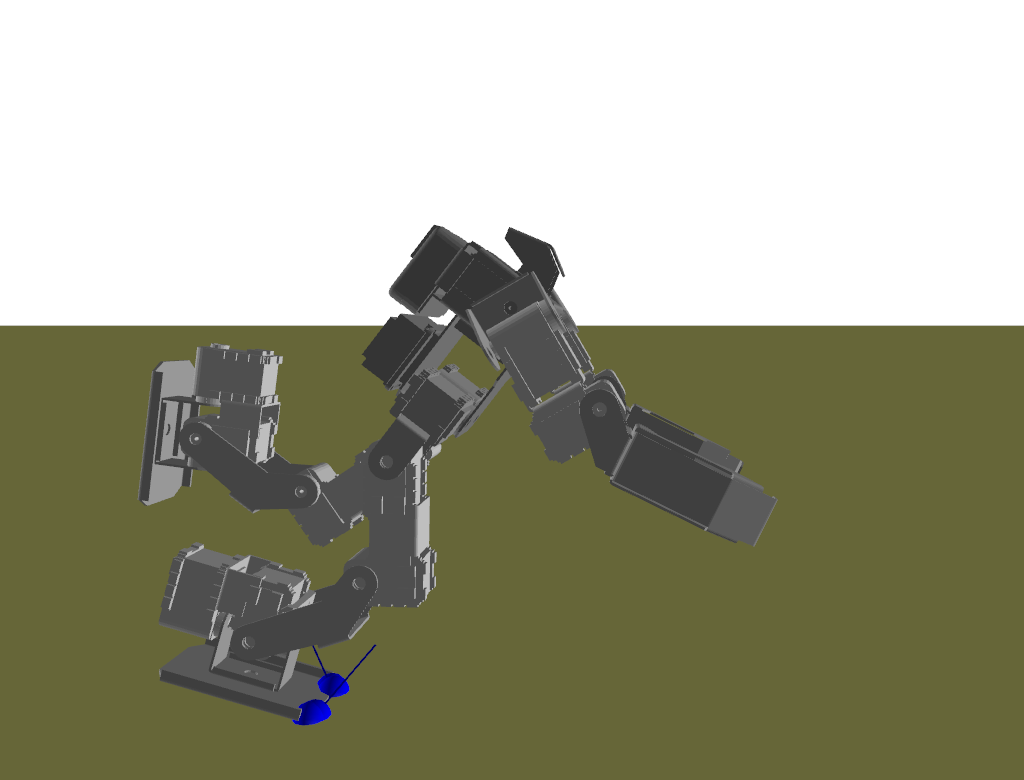}&
\includegraphics[width=0.195\textwidth]{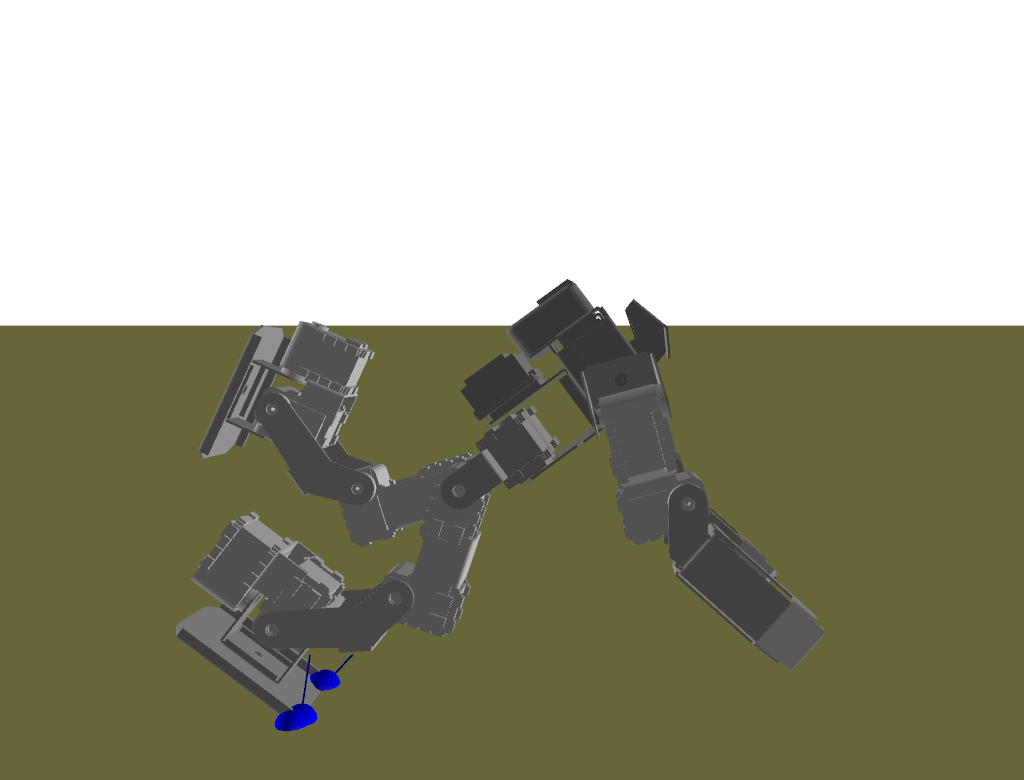}&
\includegraphics[width=0.195\textwidth]{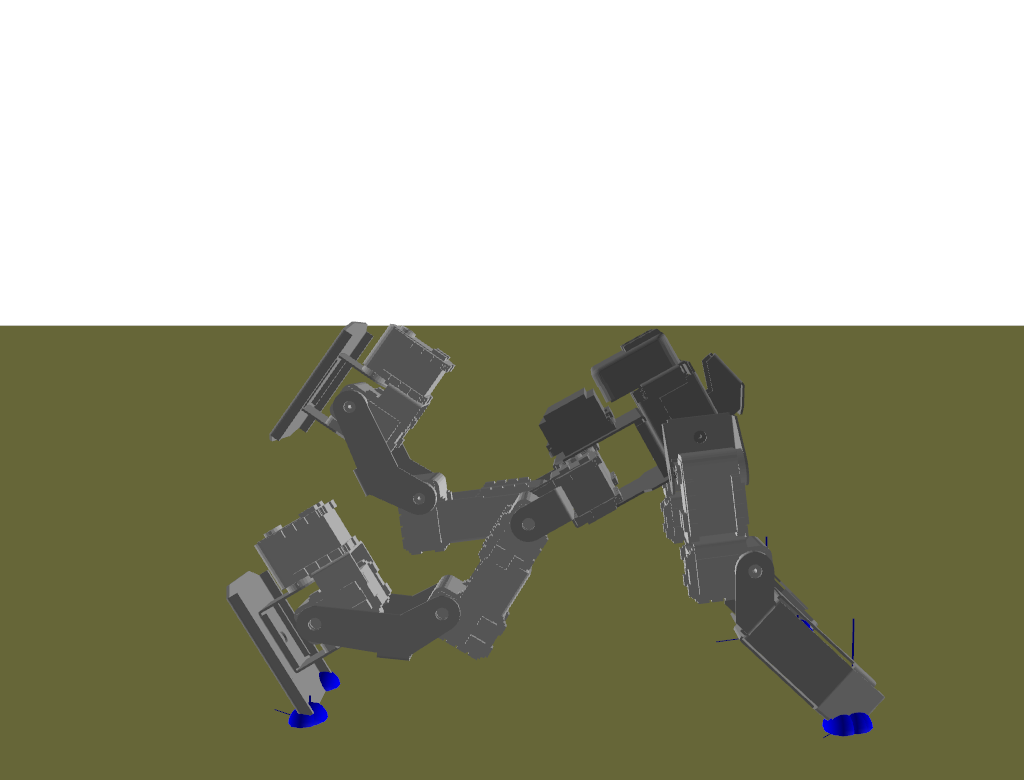} \\
\includegraphics[width=0.195\textwidth]{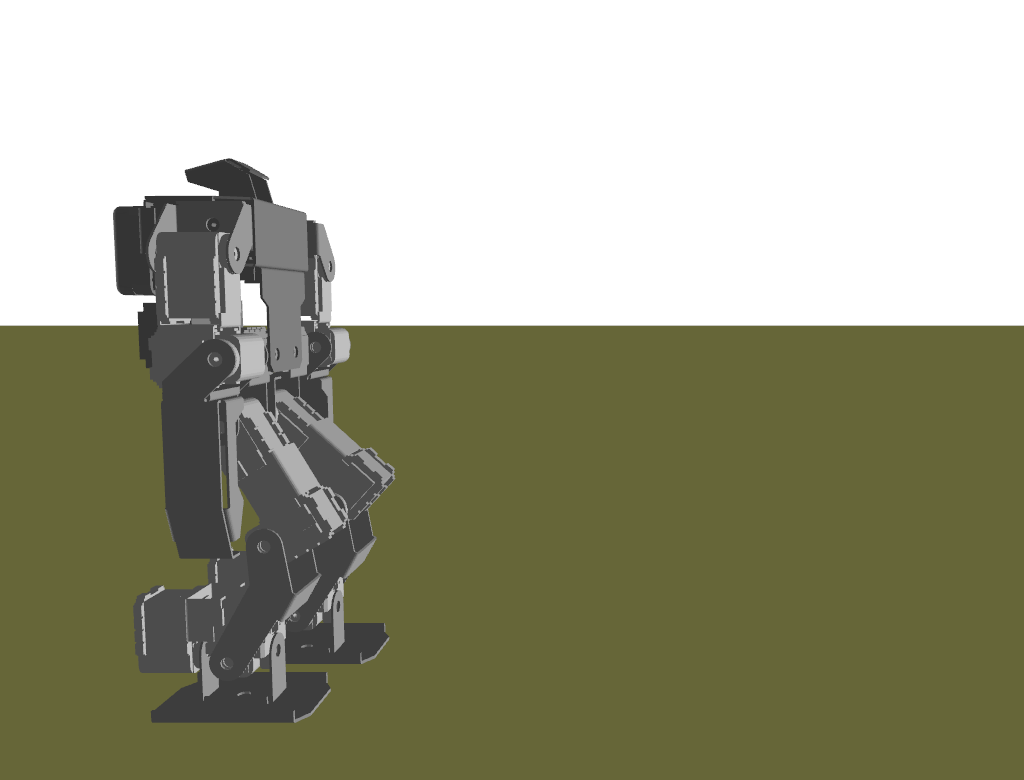}&
\includegraphics[width=0.195\textwidth]{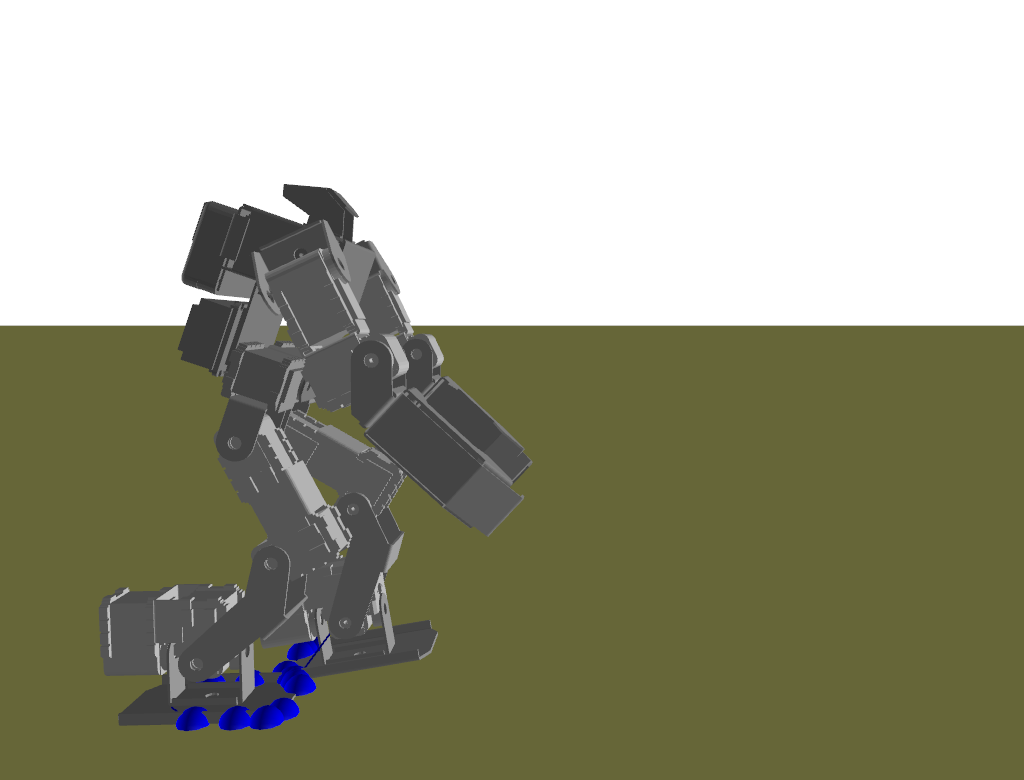}&
\includegraphics[width=0.195\textwidth]{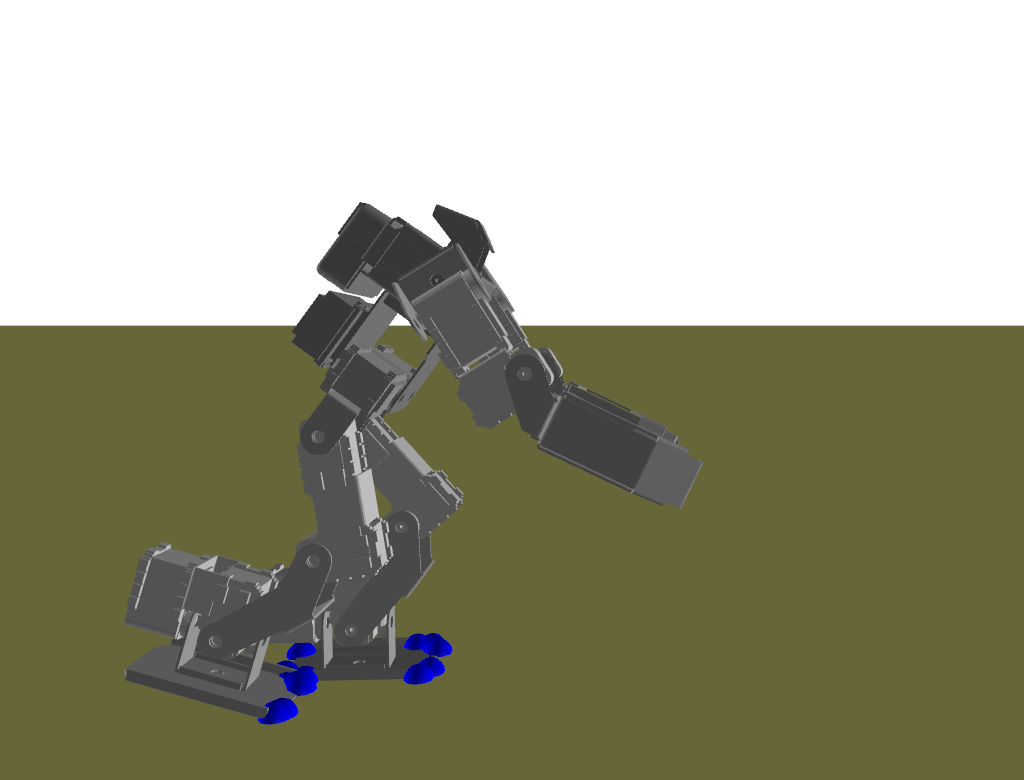}&
\includegraphics[width=0.195\textwidth]{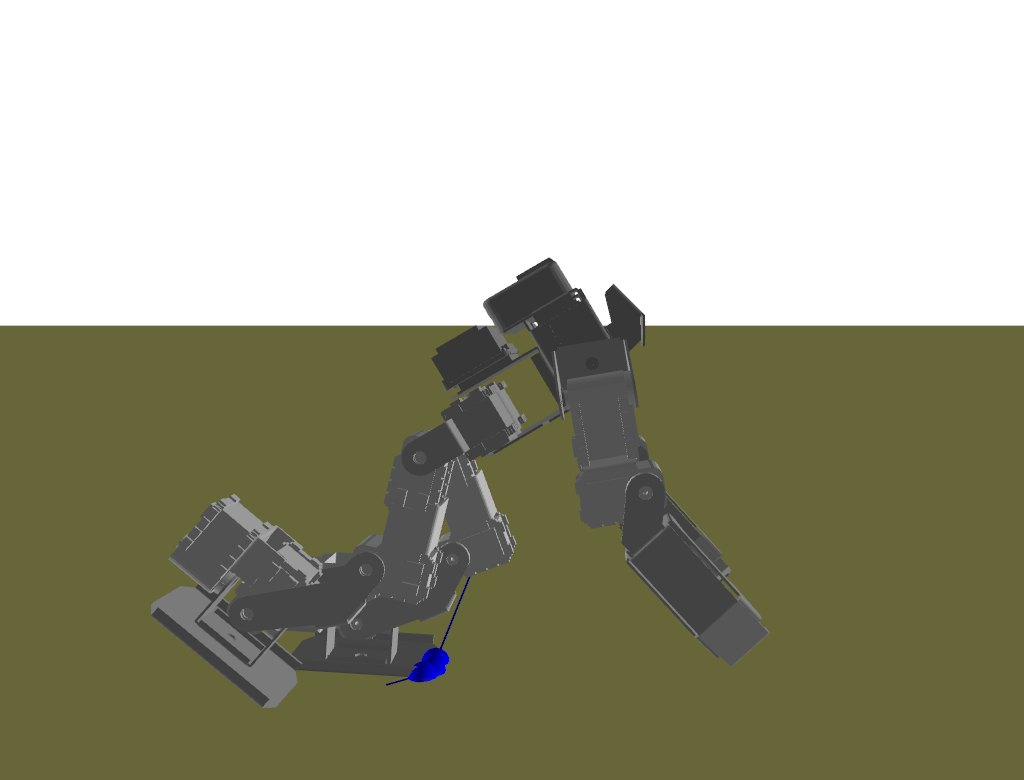}&
\includegraphics[width=0.195\textwidth]{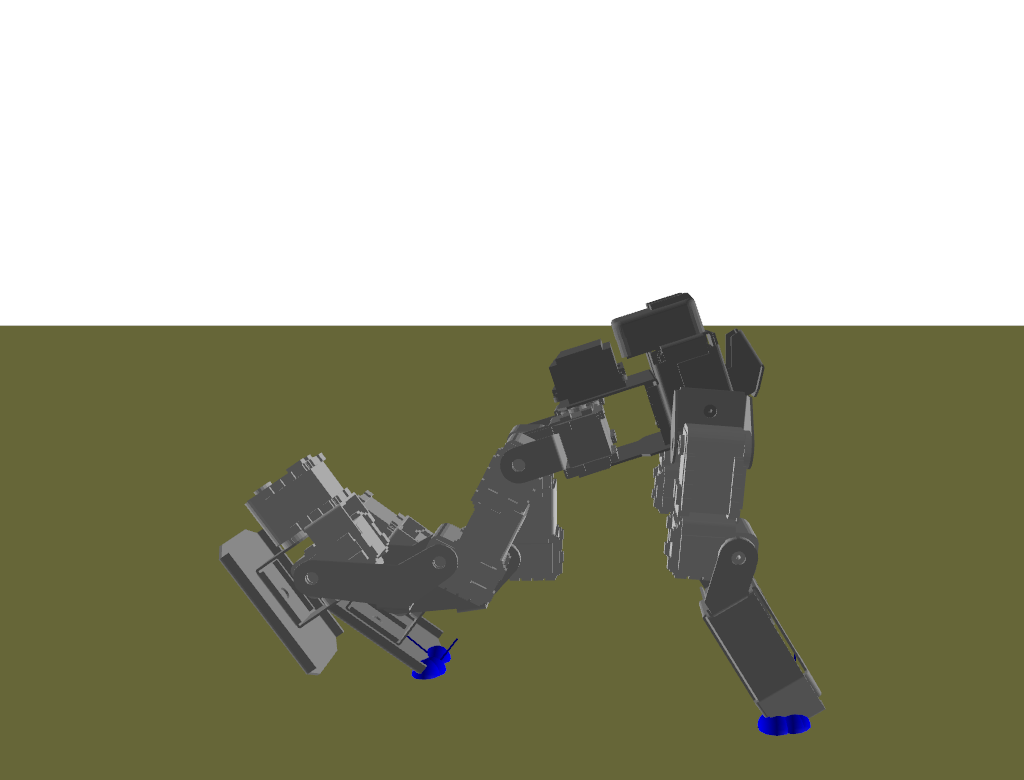} \\
\end{tabular}
\caption{\textbf{Top:} A fall from a two-feet stance due to a $3$~N push (Two-feet). 
         \textbf{Middle:} A fall from an unbalanced stance due to a $5$~N push (Unbalanced). 
         \textbf{Bottom:} A fall from a one-foot stance due to a $6$~N push (One-foot). }
  \label{fig:motions}
\end{figure*}

We compare the rewards of the trained policy with those of DP by running $1000$ test cases starting from randomly sampled initial states. The results are shown in \figref{reward_distribution}, a histogram of rewards for the $1000$ tests computed by our policy and by DP. Our policy achieves not only comparable rewards, it actually outperforms DP in $64$~\% of the test cases. The average reward of our policy is $0.8093$, comparing to $0.7784$ of DP. In terms of impulse, the average of maximum impulse produced by our policy is $0.2540$, which shows a $15$~\% improvement from $0.2997$ produced by DP.

Theoretically, DP, which searches the entire action space for the given initial state, should be the upper bound of the reward our policy can ever achieve. In practice, the discretization of the state and action spaces in DP might result in suboptimal plans. In contrast, our policy exploits the mixture of actor and critic network to optimize continuous action variables without discretization. The more precise continuous optimization often results in more optimal contact location or timing. In some cases, it also results in different contact sequences being selected (45 out of 641 test cases where our policy outperforms DP).

\subsubsection{Different falling strategies}

\begin{table}
\setlength{\tabcolsep}{3pt}
\renewcommand{\arraystretch}{1.1}
\scriptsize
\center
{
\caption{Different falling strategies.}
\begin{tabular}{| c | c | c |c | c |} 
\hline
 \makecell{\textbf{Initial} \\ \textbf{Pose}} & \textbf{Push} & \textbf{Algorithm} 
& \textbf{Contact Sequence} & \makecell{\textbf{Maximum} \\ \textbf{Impulse}} \\ \hline \hline
&       & Unplanned & Torso ($0.90$) & $0.90$ \\ \cline{3-5}
Two-feet & $3$~N & DP        & Hands ($0.58$) & $0.58$ \\ \cline{3-5}
&       & Ours      & Knees ($0.53$), Hands ($0.20$) & $0.53$ \\ \hline \hline
                    
&       & Unplanned & Torso ($2.50$) & $2.50$ \\ \cline{3-5}
Unbalanced & $5$~N & DP        & Hands ($0.53$) & $0.53$ \\ \cline{3-5}
&       & Ours      & Hands ($0.45$) & $0.45$ \\ \hline \hline
          
&       & Unplanned & Torso ($2.10$) & $2.10$ \\ \cline{3-5}
One-foot& $6$~N & DP         & L-Heel ($0.20$), Hands ($0.45$) & $0.45$ \\ \cline{3-5}
&       & Ours        & L-Heel ($0.10$), Hands ($0.43$) & $0.43$ \\ \hline  
\end{tabular} \label{tab:simulation_results}
}
\end{table}

\begin{figure}
\centering
\includegraphics[width=0.5\columnwidth]{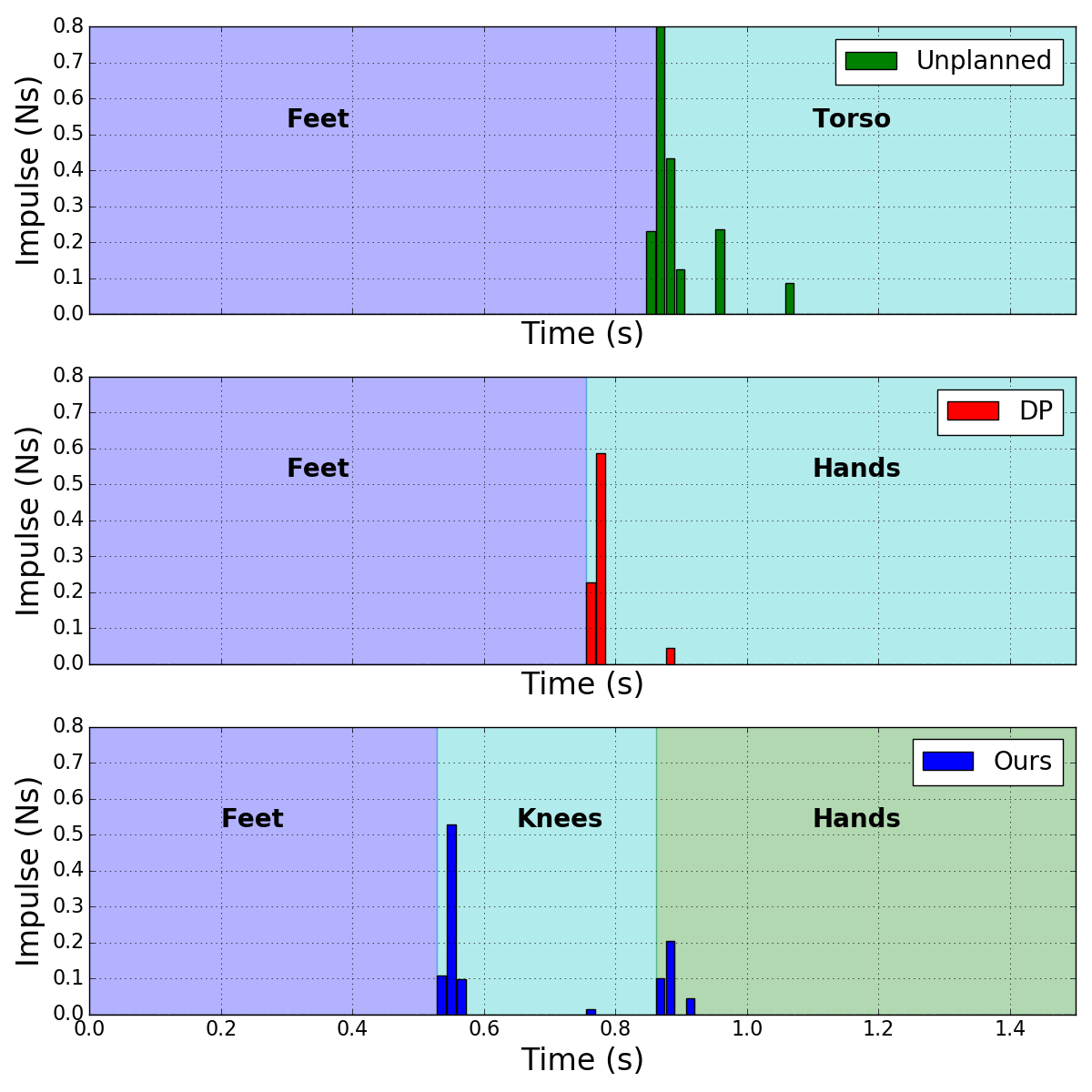}
\caption{Comparison of the impulse profiles among the unplanned motion, the motion planned by DP, and the motion planned by our policy.}
\label{fig:impulse_profile}
\end{figure}

With different initial conditions, various falling strategies emerge as the solution computed by our policy. \tabref{simulation_results} showcases three distinctive falling strategies from the test cases. The table shows the initial pose, the external push, the resulting contact sequence, the impulse due to each contact (the number in the parenthesis), as well as the maximal impulse for each test case. Starting with a two-feet stance, the robot uses knees and hands to stop a fall. If the initial state is leaning forward and unbalanced, the robot directly uses its hands to catch itself. If the robot starts with a one-foot stance, it is easer to use the swing foot followed by the hands to stop a fall. The robot motion sequences can be visualized in \ref{fig:motions} and the supplementary video. For each case, we compare our policy against DP and a naive controller which simply tracks the initial pose (referred as \emph{Unplanned}). Both our policy and DP significantly reduce the maximum impulse comparing to Unplanned. In the cases where our policy outperforms DP, the improvement can be achieved by different contact timing (One-foot case), better target poses (Unbalanced case), or different contact sequences (Two-feet case, \figref{impulse_profile}).

\subsubsection{Hardware Results}

\begin{figure}
\center
\setlength{\tabcolsep}{1pt}
\renewcommand{\arraystretch}{0.5}
\begin{tabular}{c | c}
\includegraphics[width=0.5\columnwidth]{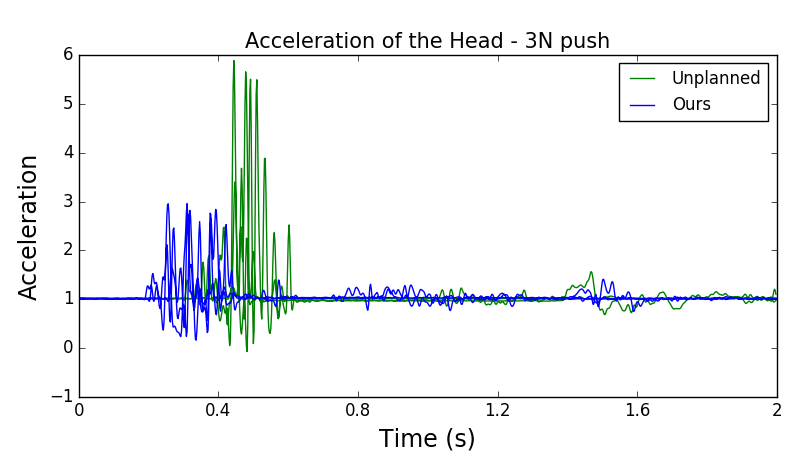} &
\includegraphics[width=0.5\columnwidth]{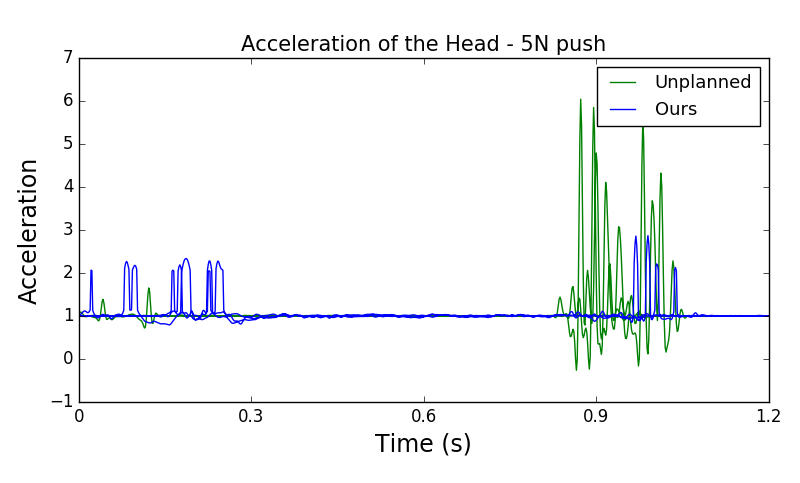} 
\end{tabular}
\caption{Comparison of measured acceleration between motion computed by our policy and unplanned motion. Three trials for each condition are plotted.
         \textbf{Left:} A fall from a two-feet stance due to a $3$~N push.  
         \textbf{Right:} A fall from an one-foot stance due to a $5$~N push. }
  \label{fig:hardware_accel}
\end{figure}

Finally, we compare the falling strategy generated by our policy against the unplanned motion on the hardware of BioloidGP. Due to the lack of on-board sensing capability, BioloidGP cannot take advantage of the feedback aspect of our policy. Nevertheless, we can still use this platform to demonstrate the falling strategy generated by our policy and compare it against an unplanned motion.

We first match the initial pose of the simulated BioloidGP with the real one and push the simulated BioloidGP from the back by $3$~N and $5$~N, assuming that the pushes we applied to the robot by hand are about the same. We then apply our policy on the simulated BioloidGP to obtain a sequence of target poses. In the hardware experiment, we program BioloidGP to track these poses once a fall is detected. During the falls, we measure the acceleration of the head using an external IMU. \ref{fig:hardware_accel} shows the results of two different falls. In the first case, the robot is pushed with a force of 3N and is initialized with both the feet on the ground and an upright position, the robot uses its knees first and then the hands to control the fall. The maximal acceleration from our policy is $2.9$~G while that from an unplanned motion is $5.7$~G, showing a $49$\% of improvement. In the second case, the robot is pushed with a force of 5N starting with one foot on the ground, the falling strategy for this includes using the left-heel first then the hands to control the fall. The maximal acceleration from our policy is $2.3$~G while that from an unplanned motion is $6.4$~G, showing a $64$\% of improvement.

\subsection{Conclusions}
We proposed a new policy optimization method to learn the appropriate actions for minimizing the damage of a humanoid fall. Unlike most optimal control problems, the action space of our problem consists of both discrete and continuous variables. To address this issue, our algorithm trains $n$ control policies (actors) and the corresponding value functions (critics) in an actor-critic network. Each actor-critic pair is associated with a candidate contacting body part. When the robot establishes a new contact with the ground, the policy corresponding to the highest value function will be executed while the associated body part will be the next contact. As a result of this mixture of actor-critic architecture, we cast the discrete contact planning into the problem of expert selection, while optimizing the policy in continuous space. We show that our algorithm reliably reduces the maximal impulse of a variety of falls. Comparing to the previous work \cite{Ha2015} that employs an expensive dynamic programming method during online execution, our policy can reach better reward and only takes $0.25\%$ to $2\%$ of computation time on average.

One limitation of this work is the assumption that humanoid falls primarily lie on the sagittal plane. This limitation is due to our choice of the simplified model, which reduces computation time but only models planar motions. This assumption can be easily challenged when considering real-world falling scenarios, such as those described in \cite{Yun2014, Goswami2014}. One possible solution to handling more general falls is to employ a more complex model similar to the inertia-loaded inverted pendulum proposed by \cite{Goswami2014}. Our algorithm also assumes that the robot is capable of motion that is fast enough to achieve the pose that the network outputs. Sensors to detect contacts are also important to use the trained policy as a feedback controller. 

Another possible future work direction is to learn control policies directly in the full-body joint space, bypassing the need of an abstract model and the restrictions come with it. This allows us to consider more detailed features of the robot during training, such as full body dynamics or precise collision shapes. Given the increasingly more powerful policy learning algorithms for deep reinforcement learning \cite{schulman2015trust,schulman2015high}, motor skill learning with a large number of variables, as is the case with falling, becomes a feasible option.

\section{Fall prevention for bipedal robots}

\subsection{Motivation}

In the previous section, we described our approach to generate falling motion using policy optimization method. However, in many situations a fall can be prevented by employing some effecting balance strategies that humans use. Similar to falling, balancing strategies are often not periodic in nature and require special control algorithms to achieve stability. 

\begin{figure}
\centering
\includegraphics[width=0.45\linewidth]{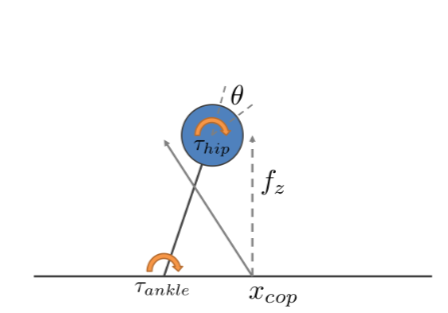}
\caption{An Linear inverted pendulum abstract model used in computing control signals for ankle and hip strategy}
\label{LIPMankle}
\end{figure}

In Kumar \etal \cite{AdaptSample}, we  propose a curriculum learning framework to learn push recovery control policies for humanoid robots. Our algorithm aimed to improve existing model-based balance controllers inspired by human balance recovery motion like hip and ankle strategies \cite{Stephens2007,Stephens,macchietto2009momentum} and stepping strategies \cite{aftab2012ankle,Perrin2013,Komura2005,pratt2006capture,Pratt2012}.
Typically, model-based control algorithms use an abstract dynamical model (like Linear-inverted pendulum LIPM or the 3D version 3D-LIPM) for computational tractability, however lower dimensional models fail to capture accurate dynamics of the system. Abstract models also neglect arm motions which can be important for balancing. Model-free reinforcement learning, on the other hand, has been shown to work well for high-dimensional systems at the cost of sample efficiency during training. To take advantage of both these methods, 
our policy outputs residual control signals in addition to control computed by model-based controllers thereby simplifying learning the control policy.

\begin{figure}
\centering
\includegraphics[width=0.45\linewidth]{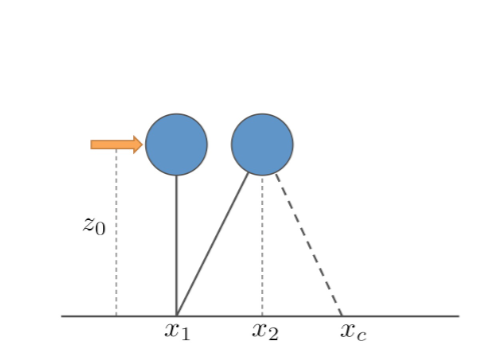}
\caption{An abstract model to compute stepping distance when a large external push is applied}
\label{LIPMstep}
\end{figure}

\ref{LIPMankle} and \ref{LIPMstep} illustrate the abstract dynamical models used by model-based controllers to compute the control signal.
The objective function to train the policy minimizes Center-of-Mass position and velocity changes when reacting to pushes to encourage balancing.
The training process also incorporates an adaptive sampling scheme that identifies the magnitude of push that needs to be applied to enable efficient learning. Our key insight here was based on observation that during training, applying pushes of large magnitudes could be detrimental to learning where as pushes of low magnitudes do not help the policy learn much. We maintain an estimate of the push magnitudes the policy is currently capable of handling , called \emph{Region of Attraction} illustrated in \ref{RoA}, and draw perturbations near its boundary. 
We show that the policy trained using our approach is capable of handling pushes of larger magnitudes compared to simple DRL baselines while also being able to generate appropriate arm-motions to aid in balance recovery.

\subsection{Adaptive sampling to simplify learning}

The control policy takes as input a partial observation vector $\vc{o} = [\vc{C}^T, \vc{\dot{C}}^T, \vc{q}^T ]^T$, where each term represents the COM position $\vc{C}$, the COM velocity $\vc{\dot{C}}$, and the joint positions $\vc{q}$.
The outputs are $12$ control signals $\vc{u} = [\boldsymbol{\Delta\tau}^T_{h}, \boldsymbol{\Delta\tau}^T_{a}, \vc{\Delta{q}}^T_{s}]^T$ where $\boldsymbol{\Delta\tau}_{h}$ and $\boldsymbol{\Delta\tau}_{a}$ are offsets to the hip and ankle torques and $\vc{\Delta{q}}_{s}$ is the offset to the shoulder target angles.
All terms have four entries for the pitch and roll axes in the left and right limbs.
The final torques are computed as:
\begin{multline} \label{eq:torque}
	\boldsymbol{\tau} = \vc{PD}(\vc{\bar{q}} + \mat{I}_{s}\vc{\Delta{\bar{q}}}_{s}, \vc{q}, \vc{\dot{q}})
    + \mat{I}_{h}(\boldsymbol{\bar{\tau}}_{h} + \boldsymbol{\Delta\tau}_{h}) 
    + \mat{I}_{a}(\boldsymbol{\bar{\tau}}_{a} + \boldsymbol{\Delta\tau}_{a})
\end{multline}
where $\vc{\bar{q}}$, $\boldsymbol{\bar{\tau}}_{h}$, $\boldsymbol{\bar{\tau}}_{a}$ are outputs from the model-based controllers and $\mat{I}_{h}$, $\mat{I}_{a}$, $\mat{I}_{s}$ matrices map the hip, ankle, and shoulder joints to the full-body joint space.
The PD controller $\vc{PD}$ provides the joint torques to track the modified target position $\vc{\bar{q}} + \mat{I}_{s}\vc{\Delta{\bar{q}}}_{s}$ for all joints except hips and ankles.

Another important component of RL is the reward function.
One of the possible reward functions is a binary success flag that is 1 if and only if the COM height is greater than the certain threshold $\bar{z}$.
However, training such a binary function is usually time-consuming and impractical.
Rather, we choose a simple continuous function that penalizes the COM position and velocity as follows:
\begin{equation} \label{eq:reward}
	r(\vc{q}, (\vc{\dot{q}})) = -w_p|\bar{\vc{C}} - \vc{C}(\vc{q})|^2 - w_d |\vc{\dot{C}(\vc{q})}|^2
\end{equation}
where the first term penalizes the positional error of the COM and the second term penalizes the velocity.
The terms $w_p$ and $w_d$ are the weights for adjusting the scales.

\begin{figure}
  \centering
  \includegraphics[width=0.5\columnwidth]{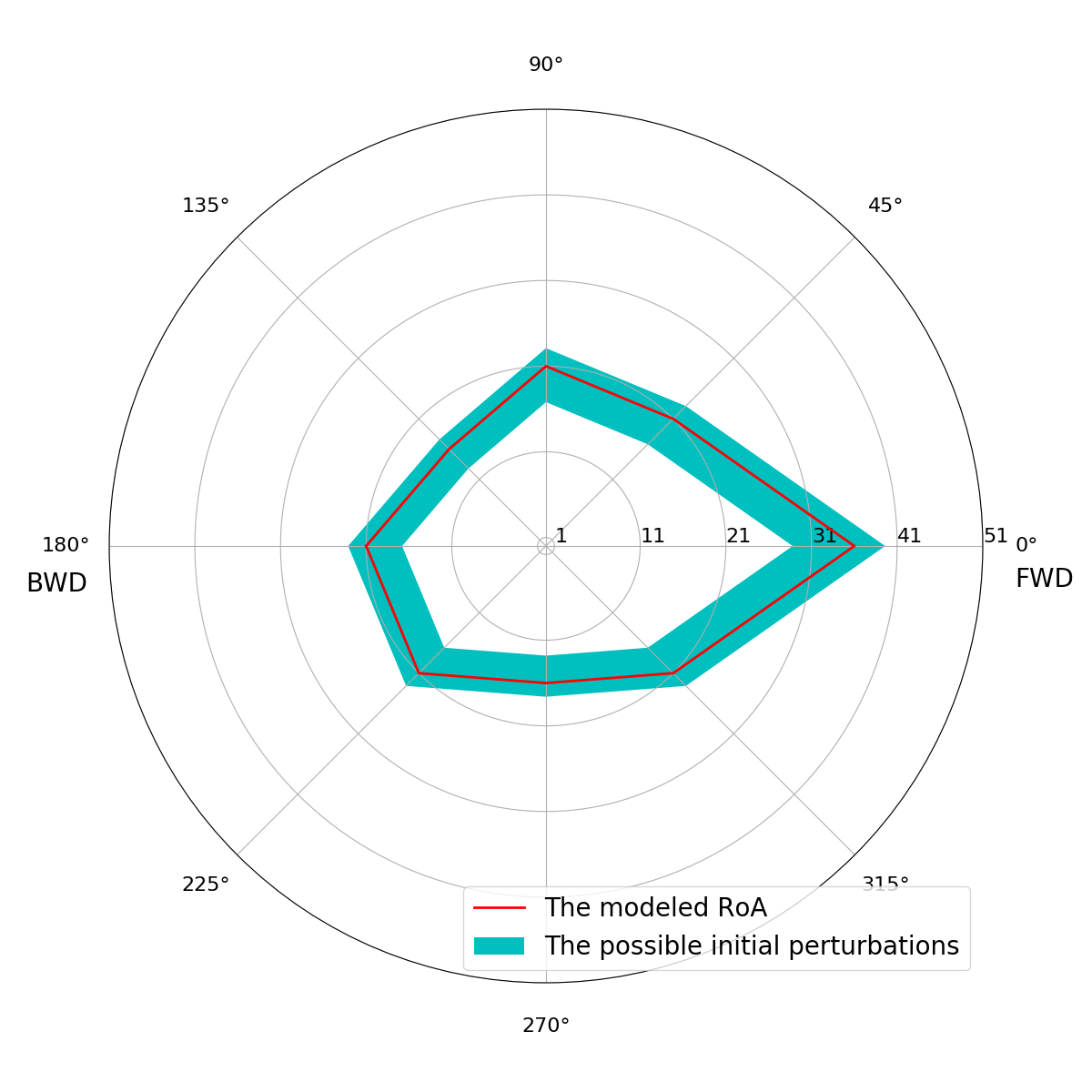}
  \vspace*{-5mm}
  \caption{Polygon representation of a RoA (red line) and the range of perturbations the controller can handle (cyan area). 
  }
  \label{fig:example_doa}
\end{figure}

\subsection{Adaptive Sampling of Perturbations} \label{sec:adaptive_sampling}
This section describes our scheme for adaptively sampling perturbations.
In our implementation, perturbations are parameterized by the directions and magnitudes of external forces.
The objective of adaptive sampling is to expedite the learning process by providing more informative trials.
For instance, trying to recover from an extremely strong perturbation will not be a useful data point because the robot will quickly fall even with the optimal control policy.
However, it is difficult to determine whether a randomly sampled perturbation is too weak or too strong because it depends on the performance of the current policy.

Our key idea is to maintain the RoA information during the learning process and sample perturbations for training using a probabilistic distribution around the boundary of the current RoA.
This process requires us to implement the following three functionalities:
\begin{itemize}
	\item Representing the RoA,
	\item Sampling perturbations from the given RoA, and
	\item Updating the RoA.
\end{itemize}

More precisely, we define the RoA $\mathcal{R}$ as the range of perturbations for which the controller shows success rates of above $90$~\%.
A trial is declared success if the COM height $C_z$ at the end of the simulation is greater than a user-defined threshold $\bar{z}$.

We represent the RoA as a polygon with $M$ sides where the vertices are defined in the polar coordinate by a set of magnitudes $ \mathcal{R} = \left\{\Omega_1, \Omega_2, \cdots, \Omega_M\right\}$ at predefined angles $\left\{\theta_1, \theta_2, \cdots, \theta_M\right\}$ chosen to uniformly cover $0$ to $2\pi$~radians (\figref{example_doa}).
The assumption underlying this representation is that a controller that works for a particular perturbation is likely to recover from a weaker perturbation in the same direction.
Although there may be counterexamples, this assumption allows us to use a simplified representation that works in practice.
In some cases, it is more convenient to represent the magnitude as a function of the angle, i.e. $\Omega = f_{\mathcal{R}}(\theta)$.

The adaptive sampling method is summarized in~\algref{adaptive_sampling}.
For each episode of learning, we first randomly sample $\theta \in \left\{\theta_1, \theta_2, \cdots, \theta_M\right\}$, and then sample $\Omega \in [k_l f_{\mathcal{R}}(\theta), k_u f_{\mathcal{R}}(\theta)]$
where $k_l\leq1$ and $1 \leq k_u$ are predefined constants.
An example of RoA representation and the range of permissible perturbations is shown in~\figref{example_doa}.

\begin{figure}
  \centering
  \includegraphics[width=0.7\columnwidth]{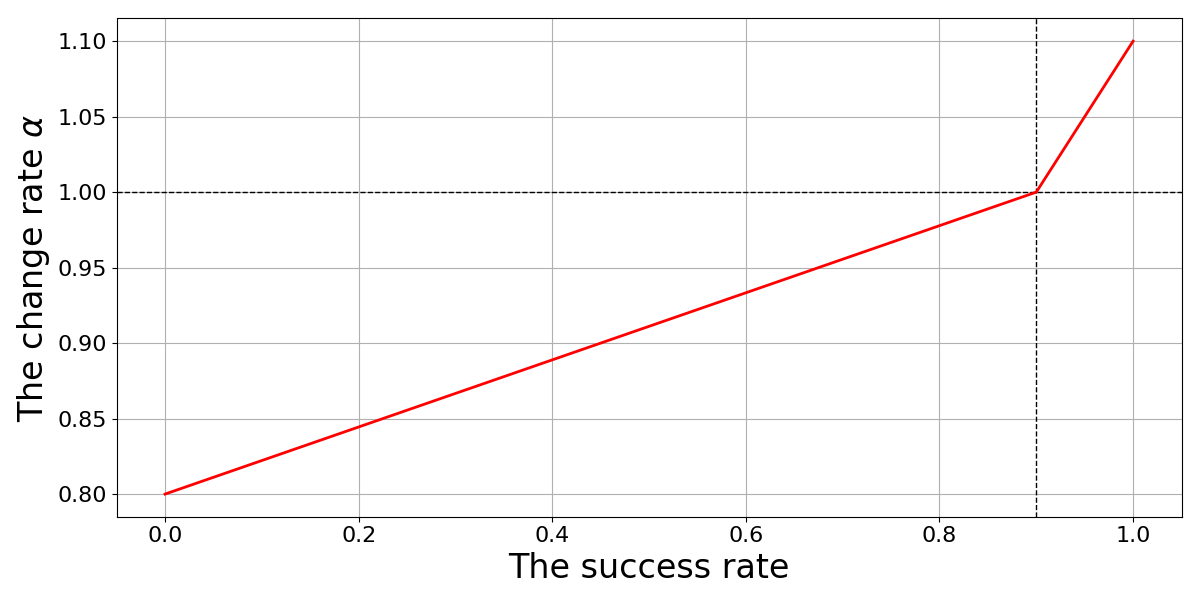}
  \caption{The function $f_{\alpha}$ that maps the success rate to the update rate $\alpha$.}
  \label{fig:rate_curve}
\end{figure}
During the training process, we update $\mathcal{R}$ for every $K$ iterations using a simple feedback rule based on the success rates at recent episodes.
At the beginning, we initialize $\mathcal{R}$ with a constant magnitude $\tilde{\Omega}$ in all directions, i.e. $\mathcal{R} = \left\{\tilde{\Omega}, \cdots, \tilde{\Omega}\right\}$. 
We collect the success rates in each direction by counting the number of good and bad trials $n^{good}_i$ and $n^{bad}_i$ where $i$ is the index of the direction closest to the direction of the sampled perturbation.
At every $K$ iterations, we increase $\omega_i$ if the success rate $n^{good}_i / (n^{good}_i + n^{bad}_i)$ is greater than $0.9$, and shrink it otherwise.
The update rate $\alpha$ is defined by a  function $f_{\alpha}$ of the success rate shown in~\figref{rate_curve}.

\begin{algorithm}
	Initialize $\mathcal{R}$ with a constant $\tilde{\Omega}$\;
    Initialize $n^{good}_1, n^{bad}_1, \cdots, n^{good}_M, n^{bad}_M$ to 0\;
    Initialize the policy $\Phi$\;
    \For {each iteration} {
    	\For{each trial} {
  			$\theta = random(0, 2\pi)$\;      	
            $\Omega = random(k_{l} f_{\mathcal{R}}(\theta), k_u f_{\mathcal{R}}(\theta))$\;
            run a simulation with perturbation $\theta$, $\Omega$\;
            $i$ = index of direction closest to $\theta$\;
            \If {$\vc{C}_z > \bar{z}$} {
            	$n^{good}_i$+=1\;
            } \Else {$n^{bad}_i$+=1\;}
        }
        update the policy $\Phi$\;
        \If {every $K$-th iteration} {
        	\For{each direction $i$} {
            	$\alpha_i = f_{\alpha}(n^{good}_i / (n^{good}_i + n^{bad}_i))$\;
                $\Omega_i = \alpha_i \Omega_i$\;
            }
            set $n^{good}_1, n^{bad}_1, \cdots, n^{good}_M, n^{bad}_M$ to 0\;
        }
    }
    \Return trained policy $\Phi$\;
    \caption{Learning with Adaptive Sampling}
    \label{alg:adaptive_sampling}
\end{algorithm}

\subsection{Results}

\begin{table}
  \begin{minipage}{\columnwidth}
    \center
    \caption{Problem and Learning Parameters}
    \setlength{\tabcolsep}{2pt}
    \scriptsize
    \begin{tabular}{| c | c | c | c |}\hline
      \textbf{Category}
      & \textbf{Name}
      & \makecell{\textbf{Postural}\\ \textbf{Controller}}
      & \makecell{\textbf{Stepping}\\ \textbf{Controller}}
      \\ \hline
      \multirow{ 4}{*}{\textbf{Problem}}
       & Perturbation Magnitudes & $(1.0, 51.0)$ & $(41.0, 81.0)$ \\ \cline{2-4}
       & Perturbation Angles & $(-\pi, \pi)$ & $(-\pi/4, \pi/4)$ \\ \cline{2-4}
       & Position Weight ($w_p$) & $1.0$ & $1.0$ \\ \cline{2-4}
       & Velocity Weight ($w_d$) & $0.1$ & $0.5$ \\ \hline
      \multirow{ 6}{*}{\textbf{TRPO}}
       & Neural Network Structure & $32,32$ & $32,32,32$  \\ \cline{2-4}
       & Activation Functions & $tanh$ & $tanh$  \\ \cline{2-4}
       & Initial Standard Deviation & $0.3$ & $0.3$  \\ \cline{2-4}
	   & Batch Size & $50000$ & $50000$\\ \cline{2-4}
       & Max Iteration & $1200$ & $1200$\\ \cline{2-4}
       & Step Size & 0.01 & 0.01 \\ \hline
	   \multirow{ 5}{*}{\textbf{Adaptive Sampling}}
       & \# of Sides ($M$) & $8$ & $3$ \\ \cline{2-4}
       & Initial Magnitude ($\tilde{\Omega}$) & $5.00$ & $41.0$ \\ \cline{2-4}
       & Update Frequency ($K$) & $4$ & $4$ \\ \cline{2-4}
       & Sampling Range Low ($k_l$) & $0.8$ & $0.7$ \\ \cline{2-4}
       & Sampling Range High ($k_l$) & $1.1$ & $1.1$ \\ \hline       
    \end{tabular}
  \label{tab:params1}
  \end{minipage}
\end{table}

We use the simulation model of the humanoid robot COMAN~\cite{tsagarakis2013compliant} to conduct our experiments.
The humanoid robot model is about $0.9$~meters tall and $31$~kg in weight.
It has $23$ joints (7 in each leg, 4 in each arm, and 3 in the torso) and $31$ DoFs including the six underactuated DoFs of the floating base, although we do not use the elbow joints in our controller.
The maximum torque of each joint is set to $35$~Nm, which is $70$~\% of the specification sheet values.
The proportional and derivative gains for the PD controller are set to $100.0$~Nm/rad and $1.0$~Nms/rad respectively.

The simulations are conducted with an open-source physics simulation engine PyDART~\cite{pydart}.
It handles contacts and collisions by formulating velocity-based linear-complementarity problems to guarantee non-penetration and approximated Coulomb friction cone conditions. 
The simulation and control time steps are set to $0.002$~s ($500$~Hz), which is enough to be executed on the actual COMAN hardware.
The computations are conducted on a single core of 3.40GHz Intel i7 processor.

For all cases, perturbations are applied to the torso of the robot for $0.2$ seconds.
The perturbation direction is defined as the direction of the external force.
For instance, a backward perturbation ($180^{\circ}$) is a push from the front direction that causes the robot to fall backward.

We use the TRPO~\cite{Schulman2015} implementation of Duan \etal~\cite{duan2016benchmarking} to train the control policy whose structures and activation functions are listed in~\tabref{params1}.
Learning of each control policy takes $12$~hours in average.
Also refer to \tabref{params1} for other hyper-parameters for problems and algorithms.

In the following sections, we will compare the following three learning methods:
\begin{itemize}
	\item Naive RL
	\item RL with a model-based controller and uniform sampling
	\item RL with a model-based controller and adaptive sampling
\end{itemize}
The baseline controller for naive RL is tracking with local PD position servos at each joint to maintain a given target pose $\bar{\vc{q}}$.
Uniform sampling draws the perturbations from all possible ranges defined in~\tabref{params1}.

\subsection{Postural Balance Controller} \label{sec:result_lqr}

\begin{table}
  \begin{minipage}{\columnwidth}
    \center
    \caption{Comparison of Various LQR Parameters}
    \setlength{\tabcolsep}{2pt}
    \scriptsize
    \begin{tabular}{| c | c | c | c |}\hline
      \makecell{\textbf{State}\\\textbf{Weight ($\mat{Q}$)}}
	  & \makecell{\textbf{Control}\\\textbf{Weight ($\mat{R}$)}}
      & \makecell{\textbf{Maximum}\\\textbf{Backward}\\\textbf{Perturb.}}
      & \makecell{\textbf{Maximum}\\\textbf{Forward}\\\textbf{Perturb.}} \\ \hline
      $diag(100, 100, 1, 1)$ & $diag(0.005, 0.005)$ & $-11$~N & $11$~N \\ \hline
      $diag(200, 200, 2, 2)$ & $diag(0.005, 0.005)$ & $-16$~N & $26$~N \\ \hline
      $diag(1000, 1000, 10, 10)$ & $diag(0.005, 0.005)$ & $-21$~N & $46$~N \\ \hline
      $diag(2000, 2000, 20, 20)$ & $diag(0.005, 0.005)$ & $-16$~N & $41$~N \\ \hline
  \end{tabular}
  \label{tab:lqrparams}
  \end{minipage}
\end{table}

First, we apply the proposed learning framework to the postural balance controller.

We select the LQR parameters by comparing various feedback matrices $\mat{K}$ from different state and input weights $\mat{Q}$ and $\mat{R}$ (\tabref{lqrparams}).
Based on the maximum permissible perturbation, we set $\mat{Q}$ to $diag(1000, 1000, 10, 10)$ and $\mat{R}$ to $diag(0.005, 0.005)$ that yields the following feedback gain:
\begin{equation}
	\mat{K} = 
    \begin{bmatrix}
        -448 & -1.30 & -45.2 & -0.14 \\
        -1.30& 447 & -0.13 & 47.3 \\
	\end{bmatrix}.
\end{equation}

\begin{figure}
  \centering
  \includegraphics[width=0.5\columnwidth]{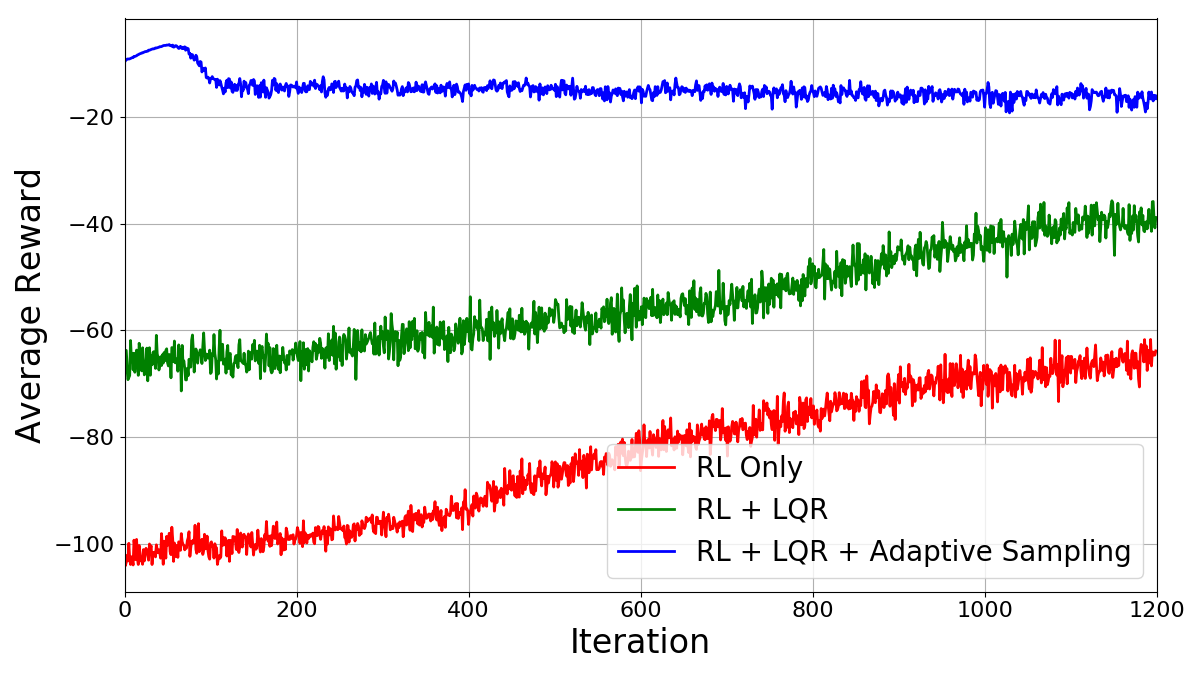}
  \caption{The learning curves for postural balance controllers.
  	Note that the learning curve with adaptive sampling (blue) is measured from a different set of perturbations.
  }
  \label{fig:learning1}
\end{figure}

First, we compare the learning curves of naive RL and RL with the LQR controller (\figref{learning1}). 
Within the same iterations, learning with the LQR achieves a reward of $-39.5$ while naive RL achieves $-64.8$.
Therefore, it is not surprising that the controller with the LQR has a larger RoA than that of naive RL, especially for forward perturbations that the LQR controller itself is already able to handle well (\figref{doa1}).


\begin{figure}
  \centering
  \includegraphics[width=0.5\columnwidth]{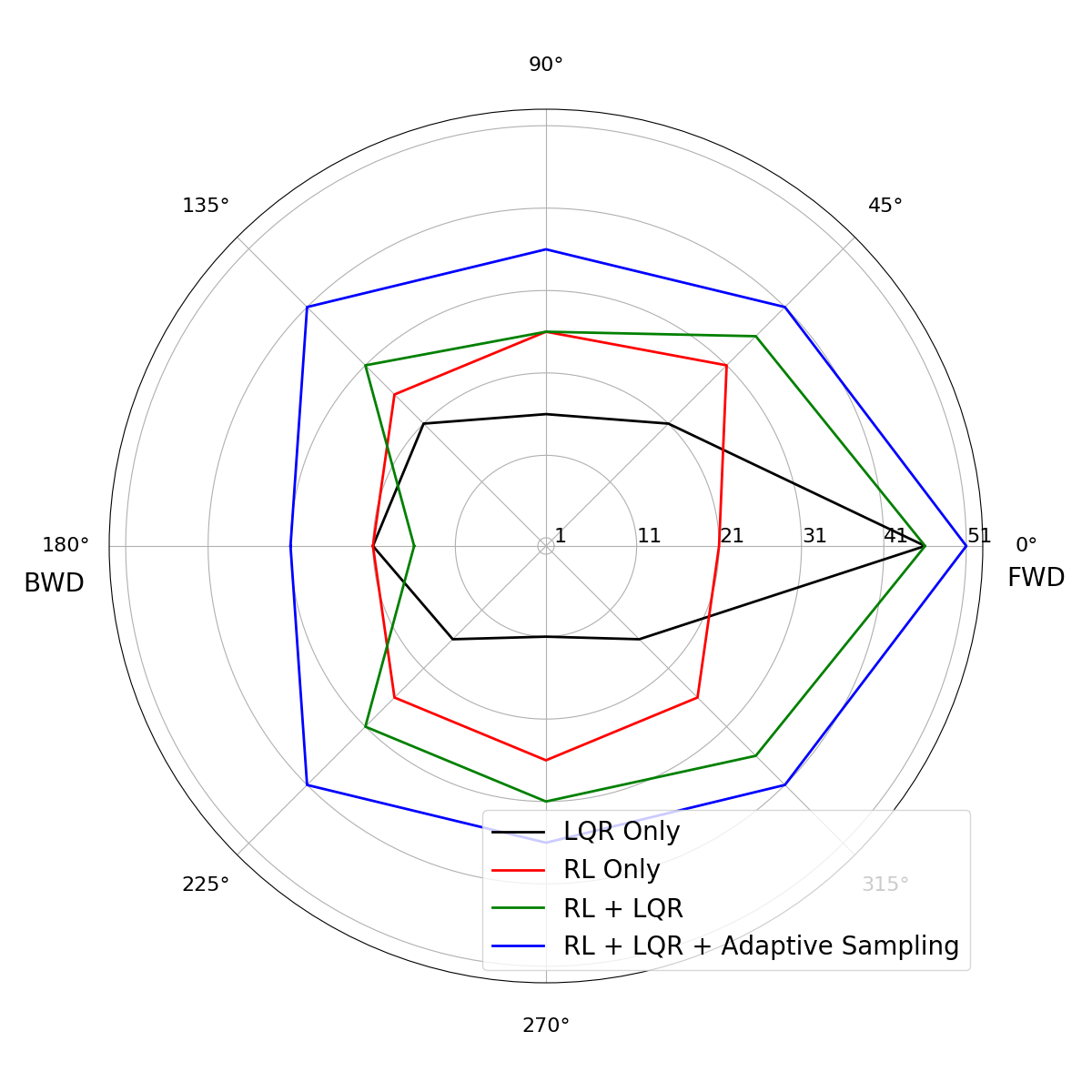}
  \vspace*{-5mm}
  \caption{RoAs for postural balance controllers.}
  \label{fig:doa1}
\end{figure}

Although learning with the LQR controller is successful, the resulting RoA is not as expanded as we expected.
For example, the final controller becomes more vulnerable to backward pushes than the initial controller, even though the total area of the RoA has been increased.
We suspect that the final controller tends to bend forward to increase the RoA in all other directions while sacrificing the ability to recover from backward perturbations.

\begin{figure*}[t]
\centering
\setlength{\tabcolsep}{1pt}
\renewcommand{\arraystretch}{0.7}
\begin{tabular}{c c c c}
  \includegraphics[width=0.24\textwidth]{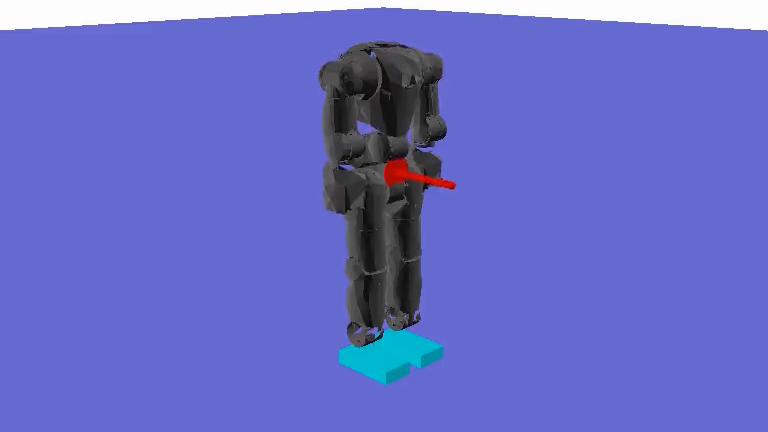}&
  \includegraphics[width=0.24\textwidth]{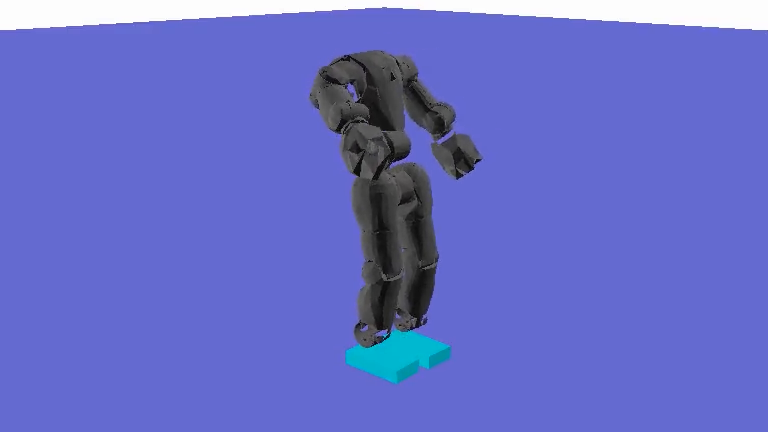}&
  \includegraphics[width=0.24\textwidth]{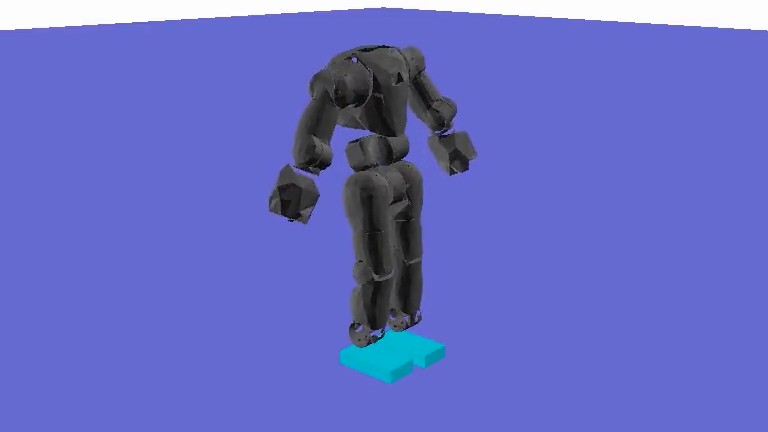}&
  \includegraphics[width=0.24\textwidth]{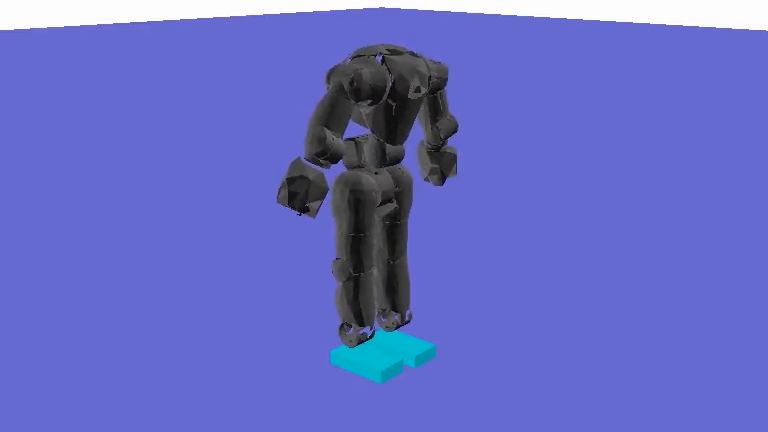} \\ 
  \includegraphics[width=0.24\textwidth]{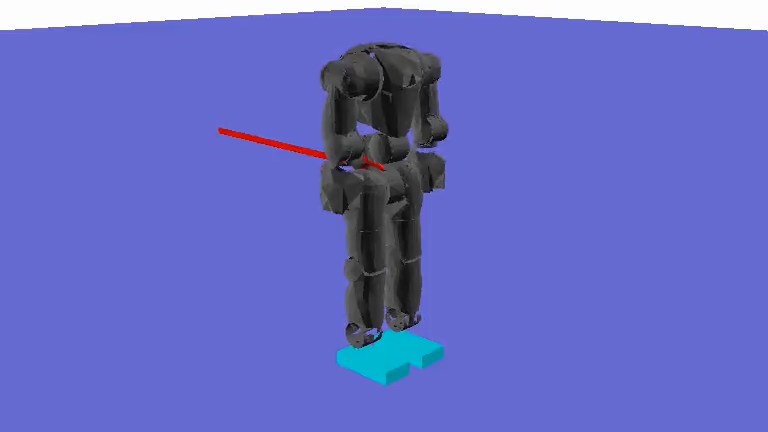}&
  \includegraphics[width=0.24\textwidth]{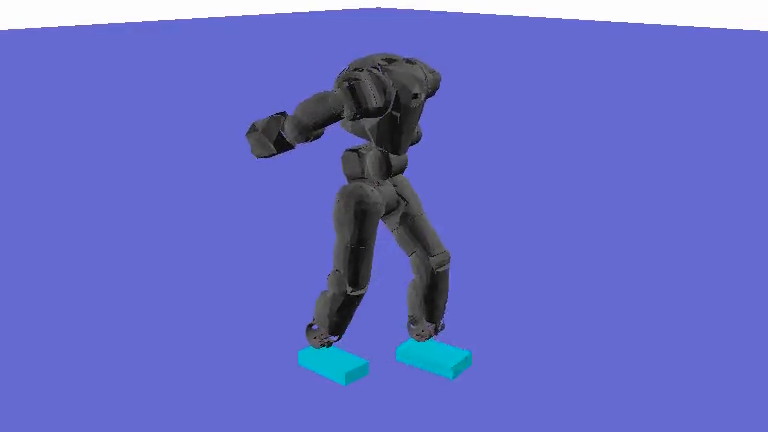}&
  \includegraphics[width=0.24\textwidth]{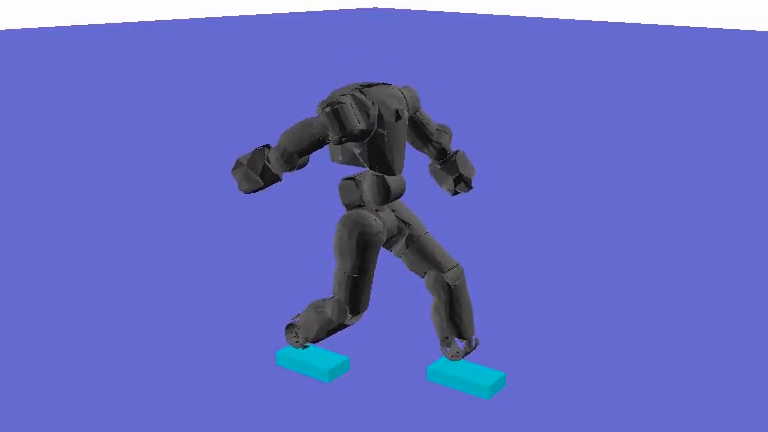}&
  \includegraphics[width=0.24\textwidth]{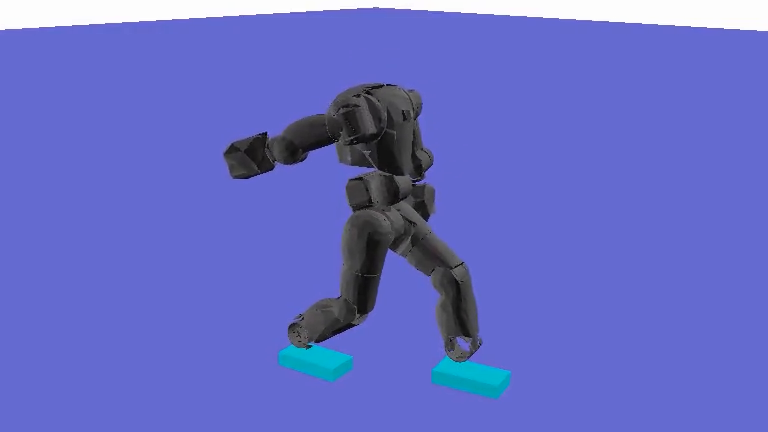} \\ 
\end{tabular}
\caption{The learned motions with adaptive sampling.
	\textbf{Top:} Postural balancing with the $26.0$~N backward perturbation.
    \textbf{Bottom:} Stepping with the $81.0$~N forward perturbation.}
\label{fig:motions}
\end{figure*}

By training with adaptive sampling, we are able to achieve a larger RoA than with uniform sampling.
The most improved directions of perturbations are backward ($180^{\circ}$) and left ($90^{\circ}$), where the controller can recover from $50$~\% and $38$~\% larger perturbations than other controllers, respectively.
When we take a look at the motions, we observe more proper arm reactions that are important to recover from side pushes.

Note that the learning curve of the adaptive sampling method cannot be compared directly to those of uniform sampling methods because the sample perturbations are drawn from different distributions.
For example, the initial average reward for adaptive sampling is much greater than uniform sampling at the first iteration due to the small $\tilde{\Omega}$ we chose for the initial RoA estimation.
Even within the adaptive sampling method, the sample distributions are different depending on the shape of the estimated RoA.
Therefore, the average reward of $-16.0$ at the $100$~th iteration and the $1200$~th iteration have different meanings.

The learning curve with adaptive sampling fluctuates a lot at the beginning of the learning process because the initial estimation of the RoA does not match well with the actual RoA.
Once they become similar to each other as learning proceeds, the reward is stabilized and shows slow changes.
The relatively flat learning curve is due to the fact that all the successful trials have similar rewards mostly between $-18.0$ and $-10.0$.
We believe this relationship between the reward and RoA expansion can be an interesting future work.

\subsection{Stepping Controller} \label{sec:result_step}

We also conduct experiments for expanding the RoA of the stepping controller~\secref{stepping}.
For this set of experiments, we only consider forward perturbations~\tabref{params1}.

\begin{figure}
  \centering
  \includegraphics[width=0.5\columnwidth]{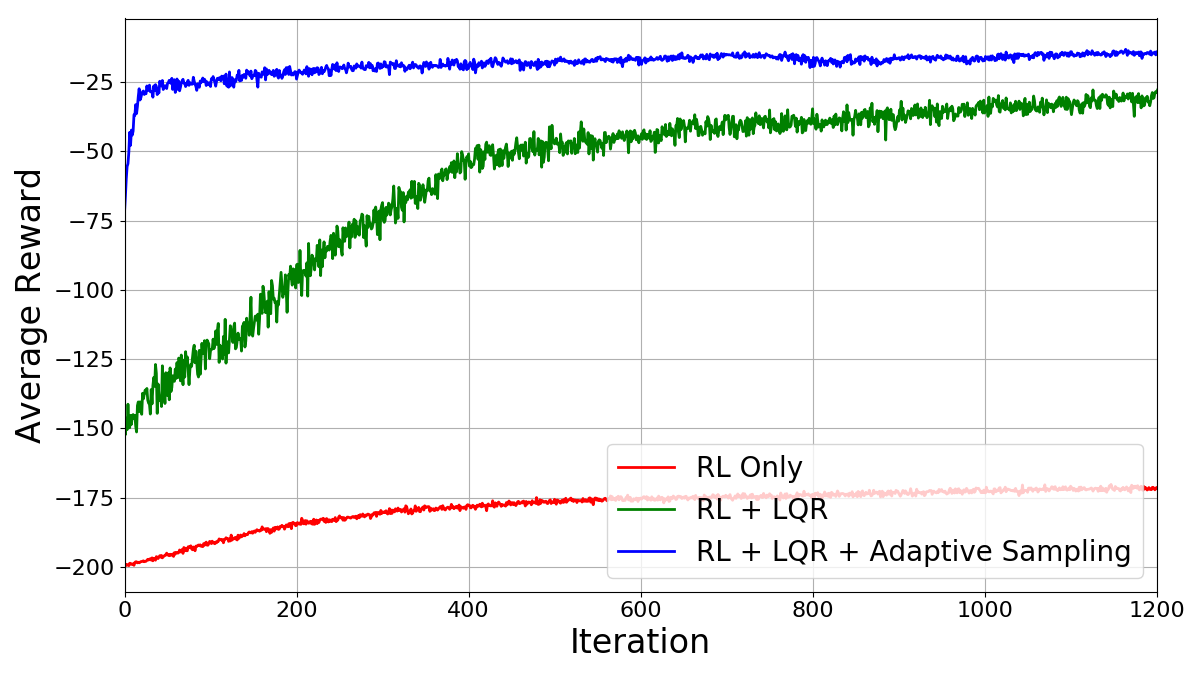}
  \caption{The learning curves for stepping controllers. 
  	Note that the learning curve with adaptive sampling (blue) is measured from a different set of perturbations.}
  \label{fig:learning2}
\end{figure}

We compare the learning curves in~\figref{learning2}.
Learning without a model-based controller cannot recover balance from most of perturbations, even though it shows improvement in terms of the reward function value.
We believe that the stepping behavior is difficult to automatically emerge without providing any prior knowledge, even after many iterations.
On the other hand, learning with a model-based controller is able to handle a wide range of perturbations.
The trained policy successfully adjusts stepping locations for completely stopping the robot by generating additional torques to the hip and ankle joints.
It also shows reactive upper body motions to compensate induced angular momentum.

\begin{figure}
  \centering
  \includegraphics[width=0.5\columnwidth]{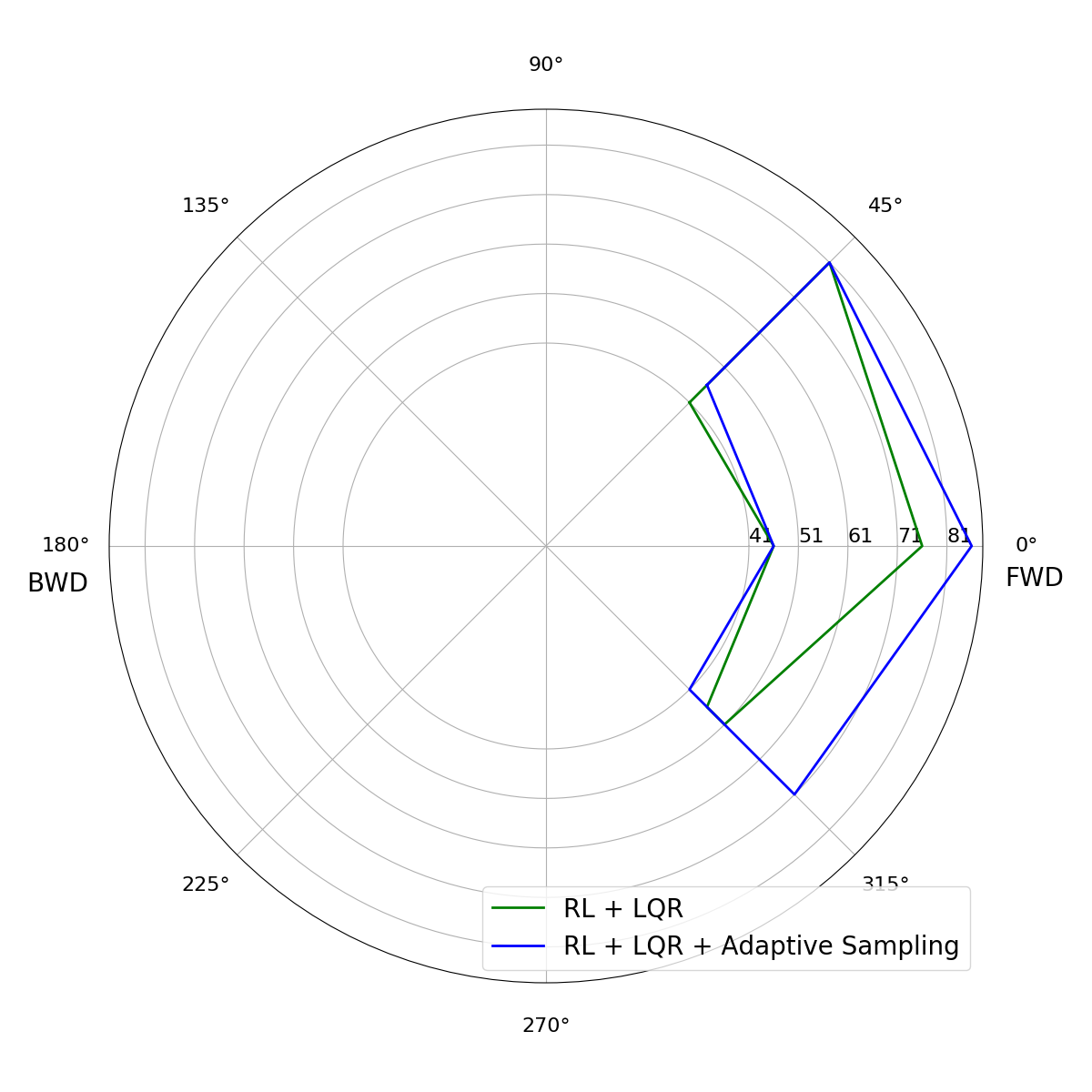}
  \vspace*{-5mm}
  \caption{RoAs for stepping controllers.}
  \label{fig:doa2}
\end{figure}

The adaptive sampling expands the RoA for the stepping controller more effectively.
Note that the learning curve of adaptive sampling (\figref{learning2}) does not have the initial fluctuation seen with the postural controller (\figref{learning1}).
This is because the initial RoA estimate is larger than the actual RoA, which was smaller in the case of postural control.
Interestingly, learning with adaptive sampling almost sequentially expands each direction of RoA: it learns first how to step right, and uses the experience to the front and left steps. 
It might be due to asymmetric upper body motions of the initial control policy that tends to move the left arm forward.
We believe that this can be another way of structuring tasks, which might be an interesting direction for future work.


\subsection{Correlation between RoA Size and Average Reward}
One interesting observation from our experiments is that the average reward is not necessarily correlated to the RoA size, which makes learning with uniform sampling more difficult if the objective is maximizing the RoA.
This is mainly because the rewards of failed simulation samples have larger variances than successful trials.
Therefore, the average reward can increase when the controller slightly improves failed samples while sacrificing the successful samples.
This issue can be even worse when there exists many impossible perturbations that cannot be handled by any controllers.

However, it is difficult to identify the exact RoA of the optimal controller before training many different controllers.
Therefore, learning with uniform sampling is likely to have many impossible samples that can skew the average reward.
On the other hand, adaptive sampling can reduce impossible trials by continuously updating the current RoA.

\subsection{Conclusion}

This paper presented a learning framework for enhancing the performance of humanoid balance controllers so that they can recover from a wider range of perturbations.
The key idea is to combine reinforcement learning with model-based controllers in order to expedite the learning process.
We also proposed a novel adaptive sampling scheme that maintains the RoA during the learning process and samples the perturbations for training around its boundary.

We conducted simulation experiments using two model-based controllers: an LQR-based postural balance controller and an LIP-based stepping controller.
We demonstrated that the proposed framework can improve the controller performance within the same number of iterations.
Furthermore, the controllers trained with adaptive sampling can accommodate larger perturbations than those with uniform sampling.

We tested the framework with different hyper-parameters for the proposed adaptive sampling method.
In our experience, the performance of the learning algorithm is not very sensitive to the hyper-parameters.
For instance, the algorithm is able to automatically adjust the estimated RoA even if we choose wrong initial magnitudes ($\tilde{\Omega}$).
The update frequency $K$ also did not significantly affect the convergence rate. 
However, we observed that a very narrow range such as $k_l$=$0.9$ and $k_u$=$1.0$ caused over-fitting for stepping controllers and cannot handle weaker perturbations around $41$~N.

Although we empirically demonstrated the effectiveness of our framework, future work could include more theoretical analysis.
For instance, the idea of drawing samples from the RoA boundary is based on intuition.
It would be more convincing if we can provide more statistical data that those boundary samples are more useful for learning than others.

Another possible future direction is to apply the proposed framework to other control problems such as locomotion or getting up.
In particular, the polygon (or polyhedron) representation of the RoA may not be sufficient for these tasks due to high-dimensional state space.
Furthermore, the discontinuity due to collisions may result in discontinuous RoA that cannot be expressed with the current representation.
However, we believe that the general idea of providing the learning process with tasks of appropriate difficulty levels can be effective in other problems.

\section{Expanding motor skills using relay networks}

\subsection{Motivation}
Often, challenging robotic tasks suffer from sub-optimal behaviour due to finding local optimum. Our insight into developing this algorithm is that, instead of learning one control policy, the task could by simplified by learning multiple policies which work towards a common goal of completing the task. Our approach begins by defining an initial set of states from which achieving the goal is easy. For example, in the classical control task of pendulum swing-up and balance, we start by learning a policy that achieves pole balancing. Then, our algorithm gradually expands by finding states where the first policy fails, and adds a new policy to the graph where the goal is to reach states in which the first policy is good at. The algorithm gradually expands the set of successful initial states by connecting new policies, one at a time, to the existing graph until the robot can achieve the original complex task. Each policy is trained for a new set of initial states with an objective function that encourages the policy to drive the new states to previously identified successful states.

In our next work \cite{expandmotor} , we introduced a new technique of building a graph of policies to simplify learning a challenging control task. Similar to hierarchical reinforcement learning (HRL) \cite{Dietterich:2000:HRL,Barto:2003:RAH} which decomposes a large-scale Markov Decision Process (MDP) into sub-tasks, the key idea of this work is to build a directed graph of local policies represented by neural networks, which we refer to as \emph{relay neural networks}, illustrated in \ref{Relay}. The nodes of the graph are state distributions and the edges are control policies that connect two state distributions.

\begin{wrapfigure}{r}{0.35\textwidth}\vspace{-2em}
\centering
\includegraphics[width=\linewidth]{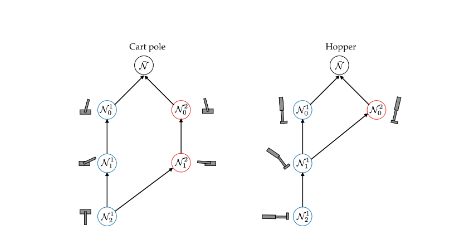}
\caption{Illustration of a graph of policies in a relay network. The policies in the graph as sequentially executed in order to complete a task. }\vspace{-2em}
\label{Relay}
\end{wrapfigure}

We apply our algorithm to a set of challenging control problems and show that the policy is capable of solving the tasks in a sample efficient manner compared to baselines such as a single policy or curriculum learning. In addition to classical control problems like Cartpole swingup, Hopper and 2D walker, we show that with a combination of policies, we can control a 3-Dimensional humanoid character with 23-Degrees of Freedom (DoF) with a goal of walking forward while also balancing.

\subsection{Method}

Our approach to a complex robotic task is to automatically decompose it to a sequence of subtasks, each of which aims to reach a state where the policy of its preceding subtask is able to handle. The original complex task is formulated as a MDP described by a tuple $\{\mathcal{S}, \mathcal{A}, r, \mathcal{T}, \rho, P\}$, where $\mathcal{S}$ is the state space, $\mathcal{A}$ is the action space, $r$ is the reward function, $\mathcal{T}$ is the set of termination conditions, $\rho=N(\boldsymbol{\mu}_\rho, \Sigma_\rho)$ is the initial state distribution, and $P$ is the transition probability. Instead of solving for a single policy for the MDP, our algorithm solves for a set of policies and value functions for a sequence of simpler MDP's. A policy, $\pi: \mathcal{S} \times \mathcal{A} \mapsto [0, 1]$, is represented as a Gaussian distribution of action $\vc{a} \in \mathcal{A}$ conditioned on a state $\vc{s} \in \mathcal{S}$. The mean of the distribution is represented by a fully connected neural network and the covariance is defined as part of the policy parameters to be optimized. A value function, $V: \mathcal{S} \mapsto R$, is also represented as a neural network whose parameters are optimized by the policy learning algorithm. 

We organize the MDP's and their solutions in a directed graph $\Gamma$. A node $\mathcal{N}_k$ in $\Gamma$ stores the initial state distribution $\rho_k$ of the $k^{th}$ MDP, while a directed edge connects an MDP to its parent MDP and stores the solved policy ($\pi_k$), the value function ($V_k$), and the threshold of value function ($\bar{V}_k$) (details in Section \ref{sec:threshold}). As the robot expands its skill set to accomplish the original task, a chain of MDP's and policies is developed in $\Gamma$. If desired, our algorithm can be extended to explore multiple solutions to achieve the original task. Section \ref{sec:multiple-chains} describes how multiple chains can be discovered and merged in $\Gamma$ to solve multi-modal problems.

\subsection{Learning Relay Networks}
\label{sec:newnodes}
The process of learning the relay networks (Algorithm \ref{alg:relay_networks}) begins with defining a new initial state distribution $\rho_0$ which reduces the difficulty of the original MDP (Line \ref{line:initial}). Although our algorithm requires the user to define $\rho_0$, it is typically intuitive to find a $\rho_0$ which leads to a simpler MDP. For example, we can define $\rho_0$ as a Gaussian whose mean, $\boldsymbol{\mu}_{\rho_0}$, is near the goal state of the problem.

Once $\rho_0$ is defined, we proceed to solve the first subtask $\{\mathcal{S}, \mathcal{A}, r, \mathcal{T}, \rho_0, P\}$, whose objective function is defined as the expected accumulated discounted rewards along a trajectory:
\begin{equation}
\label{eqn:rootObj}
J_0 = \mathbb{E}_{\vc{s}_{0:t_f}, \vc{a}_{0:t_f}}[\sum_{t=0}^{t_f} \gamma^t r(\vc{s}_t, \vc{a}_t)],
\end{equation}
where $\gamma$ is the discount factor, $\vc{s}_0$ is the initial state of the trajectory drawn from $\rho_0$, and $t_f$ is the terminal time step of the trajectory. We can solve the MDP using PPO/TRPO or A3C/DDPG to obtain the policy $\pi_0$, which drives the robot from a small set of initial states from $\rho_0$ to states where the original task is completed, as well as the value function $V_0(\vc{s})$, which evaluates the return by following $\pi_0$ from state $\vc{s}$ (Line \ref{line:solve_0}).

\begin{algorithm}[t]
\caption{LearnRelayNetworks}\label{alg:relay_networks}
\begin{algorithmic}[1]
\STATE \textbf{Input:} MDP $\{\mathcal{S}, \mathcal{A}, r, \mathcal{T}, \rho, P\}$
\STATE Add root node, $\bar{\mathcal{N}} = \{\emptyset\}$, to $\Gamma$
\STATE Define a simpler initial state distribution $\rho_0$ \label{line:initial} 
\STATE Define objective function $J_0$ according to Equation \ref{eqn:rootObj}
\STATE $[\pi_0, V_0] \leftarrow$ \textrm{PolicySearch}($\mathcal{S}, \mathcal{A}, \rho_0, J_0, \mathcal{T})$ \label{line:solve_0}
\STATE $\bar{V}_0 \leftarrow$
ComputeThreshold($\pi_0, V_0, \mathcal{T}, \rho_0, J_0$) \label{line:threshold_0}
\STATE Add node, $\mathcal{N}_0 = \{\rho_0\}$, to $\Gamma$ \label{line:addnode_0}
\STATE Add edge, $\mathcal{E} = \{\pi_0, V_0, \bar{V}_0, \}$, from $\mathcal{N}_0$ to $\bar{\mathcal{N}}$\label{line:addedge_0}
\STATE $k=0$
\WHILE{$\pi_k$ does not succeed from $\rho$} \label{line:while}
\STATE $\mathcal{T}_{k+1} = \mathcal{T} \cup (V_{k}(\vc{s}) > \bar{V}_{k})$ \label{line:terminal}
\STATE $\rho_{k+1} \leftarrow$ Compute new initial state distribution using Equation \ref{eqn:discoverNewNode} \label{line:next_initial}
\STATE Define objective function $J_{k+1}$ according to Equation \ref{eqn:relay_obj}
\STATE $[\pi_{k+1}, V_{k+1}] \leftarrow$ \textrm{PolicySearch}($\mathcal{S}, \mathcal{A}, \rho_{k+1}, J_{k+1}, \mathcal{T}_{k+1})$ \label{line:solve}
\STATE $\bar{V}_{k+1} \leftarrow$
ComputeThreshold($\pi_{k+1}, V_{k+1}, \mathcal{T}, \rho_{k+1}, J_{k+1}$) \label{line:threshold}
\STATE Add node, $\mathcal{N}^i_{k+1} = \{\rho_{k+1}\}$, to $\Gamma$ \label{line:addnode}
\STATE Add edge, $\mathcal{E} = \{\pi_{k+1}, V_{k+1}, \bar{V}_{k+1}\}$, from node $\mathcal{N}_{k+1}$ to $\mathcal{N}_k$ \label{line:addedge}
\STATE $k \leftarrow k+1$
\ENDWHILE
\RETURN{$\Gamma$}
\end{algorithmic}
\end{algorithm}

The policy for the subsequent MDP aims to drive the rollouts toward the states which $\pi_0$ can successfully handle, instead of solving for a policy that directly completes the original task. To determine whether $\pi_0$ can handle a given state $\vc{s}$, one can generate a rollout by following $\pi_0$ from $\vc{s}$ and calculate its return. However, this approach can be too slow for online applications. Fortunately, many of the modern policy gradient methods, such as PPO or A3C, produce a value function, which provides an approximated return from $\vc{s}$ without generating a rollout. Our goal is then to determine a threshold $\bar{V}_0$ for $V_0(\vc{s})$ above which $\vc{s}$ is deemed ``good'' (Line \ref{line:threshold_0}). The details on how to determine such a threshold are described in Section \ref{sec:threshold}. We can now create the first  node $\mathcal{N}_0 = \{\rho_0\}$ and add it to $\Gamma$, as well as an edge $\mathcal{E} = \{\pi_0, V_0, \bar{V}_0\}$ that connects $\mathcal{N}_0$ to the dummy root node $\bar{\mathcal{N}}$ (Line \ref{line:addnode_0}-\ref{line:addedge_0}).

Starting from $\mathcal{N}_0$, the main loop of the algorithm iteratively adds more nodes to $\Gamma$ by solving subsequent MDP's until the last policy $\pi_k$, via relaying to previous policies $\pi_{k-1}, \cdots \pi_0$, can generate successful rollouts from the states drawn from the original initial state distribution $\rho$ (Line \ref{line:while}). At each iteration $k$, we formulate the $(k+1)^{th}$ subsequent MDP by redefining $\mathcal{T}_{k+1}$, $\rho_{k+1}$, and the objective function $J_{k+1}$, while using the shared $\mathcal{S}$, $\mathcal{A}$, and $P$. First, we define the terminal conditions $\mathcal{T}_{k+1}$ as the original set of termination conditions ($\mathcal{T}$) augmented with another condition, $V_{k}(\vc{s}) > \bar{V}_{k}$ (Line \ref{line:terminal}). Next, unlike $\rho_0$, we define the initial state distribution $\rho_{k+1}$ through an optimization process. The goal of the optimization is to find the mean of the next initial state distribution, $\boldsymbol{\mu}_{\rho_{k+1}}$, that leads to unsuccessful rollouts under the current policy $\pi_k$, without making the next MDP too difficult to relay. To balance this trade-off, the optimization moves in a direction that reduces the value function of the most recently solved MDP, $V_k$, until it reaches the boundary of the ``good state zone'' defined by $V_k(\vc{s})\geq \bar{V}_k$. In addition, we would like $\boldsymbol{\mu}_{\rho_{k+1}}$ to be closer to the mean of the initial state distribution of the original MDP, $\boldsymbol{\mu}_\rho$. Specifically, we compute $\boldsymbol{\mu}_{\rho_{k+1}}$ by minimizing the following objective function subject to constraints: 

\begin{align}
\label{eqn:discoverNewNode}
\boldsymbol{\mu}_{\rho_{k}} = \operatorname*{argmin}_{\vc{s}} V_{k-1}(\vc{s}) + w \|\vc{s} - \boldsymbol{\mu}_{\rho}\|^2 \nonumber \\
\text{subject to}\;\;\; V_{k-1}(\vc{s}) \geq \bar{V}_{k-1} \nonumber \\
C(\vc{s}) \geq 0 
\end{align}

where $C(\vc{s})$ represents the environment constraints, such as the constraint that enforces collision-free states. Since the value function $V_k$ in Equation \ref{eqn:discoverNewNode} is differentiable through back-propagation, we can use any standard gradient-based algorithms to solve this optimization.

In addition, we define the objective function of the $(k+1)^{th}$ MDP as follows:
\begin{align}
\label{eqn:relay_obj}
J_{k+1} = \mathbb{E}_{\vc{s}_{0:t_f}, \vc{a}_{0:t_f}}[\sum_{t=0}^{t_f} \gamma^t r(\vc{s}_t, \vc{a}_t) + \textbf{$\alpha$} \gamma^{t_f} g(\vc{s}_{t_f})] ,\hspace{1.5cm} &  \\
\mathrm{where\;\;\;} g(\vc{s}_{t_f}) = \begin{cases} 
V_{k}(\vc{s}_{t_f}), & V_{k}(\vc{s}_{t_f}) > \bar{V}_{k}\\
0 & \mathrm{otherwise}.
\end{cases}. \nonumber
\end{align}

Besides the accumulated reward, this cost function has an additional term to encourage ``relaying''. That is, if the rollout is terminated because it enters the subset of $\mathcal{S}$ where the policy of the parent node is capable of handling (\ie $V_{k}(\vc{s}_{t_f}) > \bar{V}_{k}$), it will receive the accumulated reward by following the parent policy from $\vc{s}_{t_f}$. This quantity is approximated by $V_{k}(\vc{s}_{t_f})$ because it recursively adds the accumulated reward earned by each policy along the chain from $\mathcal{N}_k$ to $\mathcal{N}_0$. If a rollout terminates due to other terminal conditions (\eg falling on the ground for a locomotion task), it will receive no relay reward. Using this objective function, we can learn a policy $\pi_{k+1}$ that drives a rollout towards states deemed good by the parent's value function, as well as a value function $V_{k+1}$ that measures the long-term reward from the current state, following the policies along the chain (Line \ref{line:solve}). Finally, we compute the threshold of the value function (Line \ref{line:threshold}), add a new node, $\mathcal{N}_{k+1}$, to $\Gamma$ (Line \ref{line:addnode}), and add a new edge that connects $\mathcal{N}_{k+1}$ to $\mathcal{N}_k$ (Line \ref{line:addedge}).
 
The weighting parameter $\alpha$ determines the importance of ``relaying'' behavior. If $\alpha$ is set to zero, each MDP will attempt to solve the original task on its own without leveraging previously solved policies. The value of $\alpha$ in all our experiments is set to $30$ and we found that the results are not sensitive to $\alpha$ value, as long as it is sufficiently large (Section \ref{sec:alphaComparison}).

\subsection{Computing Threshold for Value Function}
\label{sec:threshold}

\begin{algorithm}[t]
\caption{ComputeThreshold}\label{alg:computeThreshold}
\begin{algorithmic}[1]
\STATE \textbf{Input:} $\pi, V, \mathcal{T}, \rho = N(\boldsymbol{\mu}_\rho, \Sigma_\rho), J$
\STATE Initialize buffer $\mathcal{D}$ for training data
\STATE $[\vc{s}_1, \cdots, \vc{s}_M] \leftarrow$ Sample states from $N(\boldsymbol{\mu}_\rho, 1.5\Sigma_\rho) $ \label{line:sample}
\STATE $[\tau_1, \cdots, \tau_M] \leftarrow$ Generate rollouts by following $\pi$ and $\mathcal{T}$ from $[\vc{s}_1, \cdots, \vc{s}_M]$ \label{line:generate}
\STATE Compute returns for rollouts: $R_i = J(\tau_i),\; i \in [1,M]$ \label{line:computeReturn}
\STATE $\bar{R} \leftarrow$ Compute average of returns for rollouts not terminated by $\mathcal{T}$ \label{line:averageReturn}
\FOR{$i = 1:M$} \label{line:trainingSetBegin}
\IF{$R_i > \bar{R}$}
\STATE Add $(V(\vc{s}_i), 1)$ in $\mathcal{D}$
\ELSE
\STATE Add $(V(\vc{s}_i), 0)$ in $\mathcal{D}$
\ENDIF 
\ENDFOR \label{line:trainingSetEnd}
\STATE $\bar{V} \leftarrow$ Classify($\mathcal{D}$) \label{line:classify}
\RETURN{$\bar{V}$}
\end{algorithmic}
\end{algorithm}

In practice, $V(\vc{s})$ provided by the learning algorithm is only an approximation of the true value function. We observe that the scores $V(\vc{s})$ assigns to the successful states are relatively higher than the unsuccessful ones, but they are not exactly the same as the true returns. As such, we can use $V(\vc{s})$ as a binary classifier to separate ``good'' states from ``bad'' ones, but not as a reliable predictor of the true returns.

To use $V$ as a binary classifier of the state space, we first need to select a threshold $\bar{V}$. For a given policy, separating successful states from unsuccessful ones can be done as follows. First, we compute the average of true return, $\bar{R}$, from a set of sampled rollouts that do not terminate due to failure conditions. Second, we compare the true return of a given state to $\bar{R}$ to determine whether it is a successful state (successful if the return is above $\bar{R}$). In practice, however, we can only obtain an approximated return of a state via $V(\vc{s})$ during policy learning and execution. Our goal is then to find the optimal $\bar{V}$ such that the separation of approximated returns by $\bar{V}$ is as close as possible to the separation of true returns by $\bar{R}$.


Algorithm \ref{alg:computeThreshold} summarizes the procedure to compute $\bar{V}$. Given a policy $\pi$, an approximated value function $V$, termination conditions $\mathcal{T}$, a Gaussian initial state distribution $\rho = N(\boldsymbol{\mu}_{\rho}, \Sigma_\rho)$, and the objective function of the MDP $J$, we first draw $M$ states from an expanded initial state, $N(\boldsymbol{\mu}_{\rho}, 1.5 \Sigma_\rho)$ (Line \ref{line:sample}), generate rollouts from these sampled states using $\pi$ (Line \ref{line:generate}), and compute the true return of each rollout using $J$ (Line \ref{line:computeReturn}). Because the initial states are drawn from an inflated distribution, we obtain a mixed set of successful and unsuccessful rollouts. We then compute the average return of successful rollouts $\bar{R}$ that do not terminate due to the terminal conditions $\mathcal{T}$ (Line \ref{line:averageReturn}). Next, we generate the training set where each data point is a pair of the predicted value $V(\vc{s}_i)$ and a binary classification label, ``good'' or ``bad'', according to $\bar{R}$ (\ie $R_i > \bar{R}$ means $\vc{s}_i$ is good) (Line \ref{line:trainingSetBegin}-\ref{line:trainingSetEnd} ). We then train a binary classifier represented as a decision tree to find to find $\bar{V}$ (Line \ref{line:classify}).

\subsection{Applying Relay Networks}


Once the graph of relay networks $\Gamma$ is trained, applying the polices is quite straightforward. For a given initial state $\vc{s}$, we select a node $c$ whose $V(\vc{s})$ has the highest value among all nodes. We execute the current policy $\pi_c$ until it reaches a state where the value of the parent node is greater than its threshold ($V_{p(c)}(\vc{s}) > \bar{V}_{p(c)}$), where $p(c)$ indicates the index of the parent node of c. At that point, the parent control policy takes over and the process repeats until we reach the root policy. Alternatively, instead of always switching to the parent policy, we can switch to another policy whose $V(\vc{s})$ has the highest value. 

\subsection{Extending to Multiple Strategies}
\label{sec:multiple-chains}
Optionally, our algorithm can be extended to discover multiple solutions to the original MDP. Take the problem of cartpole swing-up and balance as an example. In Algorithm \ref{alg:relay_networks}, if we choose the mean of $\rho_0$ to be slightly off to the left from the balanced goal position, we will learn a chain of relay networks that often swings to the left and approaches the balanced position from the left. If we run Algorithm \ref{alg:relay_networks} again with the mean of $\rho_0$ leaning toward right, we will end up learning a different chain of polices that tends to swing to the right. For a problem with multi-modal solutions, we extend Algorithm \ref{alg:relay_networks} to solve for a directed graph with multiple chains and describe an automatic method to merge the current chain into an existing one to improve sample efficiency. Specifically, after the current node $\mathcal{N}_k$ is added to $\Gamma$ and the next initial state distribution $\rho_{k+1}$ is proposed (Line \ref{line:next_initial} in Algorithm \ref{alg:relay_networks}), we compare $\rho_{k+1}$ against the initial state distribution stored in every node on the existing chains (excluding the current chain). If there exists a node $\tilde{\mathcal{N}}$ with a similar initial state distribution, we merge the current chain into $\tilde{\mathcal{N}}$ by learning a policy (and a value function) that relays to $\mathcal{N}_k$ from the initial state distribution of $\tilde{\mathcal{N}}$, essentially adding an edge from $\tilde{\mathcal{N}}$ to $\mathcal{N}_k$ and terminating the current chain. Since now $\tilde{\mathcal{N}}$ has two parents, it can choose which of the two policies to execute based on the value function or by chance. Either path will lead to completion of the task, but will do so using different strategies.

\subsection{Results}

We evaluate our algorithm on motor skill control problems in simulated environments. We use DART physics engine \cite{DART} to create five learning environments similar to Cartpole, Hopper, 2D Walker, and Humanoid environments in Open-AI Gym \cite{OpenAI}. To demonstrate the advantages of relay networks, our tasks are designed to be more difficult than those in Open-AI Gym. Implementation details can be found in the supplementary document. We compare our algorithm to three baselines:
\begin{itemize}
\item{A single policy (ONE)}: ONE is a policy trained to maximize the objective function of the original task from the  initial state distribution $\rho$. For fair comparison, we train ONE with the same number of samples used to train the entire relay networks graph. ONE also has the same number of neurons as the sum of neurons used by all relay policies. 
\item{No relay (NR)}: NR validates the importance of relaying which amounts to the second term of the objective function in Equation \ref{eqn:relay_obj} and the terminal condition, $V_k(\vc{s}) > \bar{V}_k$. NR removes these two treatments, but otherwise identical to relay networks. To ensure fairness, we use the same network architectures and the same amount of training samples as those used for relay networks. Due to the absence of the relay reward and the terminal condition $V_k(\vc{s} > \bar{V}_k$, each policy in NR attempts to achieve the original task on its own without relaying. During execution, we evaluate every value function at the current state and execute the policy that corresponds to the highest value.
\item{Curriculum learning (CL): We compare with curriculum learning which trains a single policy with increasingly more difficult curriculum. We use the initial state distributions computed by Algorithm \ref{alg:relay_networks} to define different curriculum. That is, we train a policy to solve a sequence of MDP's defined as $\{\mathcal{S}, \mathcal{A}, r, \mathcal{T}, \rho_k, P\}$, $k\in[0,K]$, where $K$ is the index of the last node on the chain. When training the next MDP, we use previously solved $\pi$ and $V$ to "warm-start" the learning.}
\end{itemize}

\subsection{Tasks}
We will briefly describe each task in this section. Please see Appendix B in the supplementary document for detailed description of the state space, action space, reward function, termination conditions, and initial state distribution for each problem.

\begin{itemize}
\item{\textbf{Cartpole:}}
Combining the classic cartpole balance problem with the pendulum swing-up problem, this example trains a cartpole to swing up and balance at the upright position. The mean of initial state distribution, $\boldsymbol{\mu}_{\rho}$, is a state in which the pole points straight down and the cart is stationary. Our algorithm learns three relay policies to solve the problem.



\item{\textbf{Hopper:}}
This example trains a 2D one-legged hopper to get up from a supine position and hop forward as fast as possible. We use the same hopper model described in Open AI Gym. The main difference is that $\boldsymbol{\mu}_{\rho}$ is a state in which the hopper lies flat on the ground with zero velocity, making the problem more challenging than the one described in OpenAI Gym. Our algorithm learns three relay policies to solve the problem.



\item{\textbf{2D walker with initial push:}}
The goal of this example is to train a 2D walker to overcome an initial backward push and walk forward as fast as possible. We use the same 2D walker environment from Open AI Gym, but modify $\boldsymbol{\mu}_{\rho}$ to have a negative horizontal velocity. Our algorithm learns two relay policies to solve the problem.

\item{\textbf{2D walker from supine position:}}
We train the same 2D walker to get up from a supine position and walk as fast as it can. $\boldsymbol{\mu}_{\rho}$ is a state in which the walker lies flat on the ground with zero velocity. Our algorithm learns three relay policies to solve the problem.

\item{\textbf{Humanoid walks:}} 
This example differs from the rest in that the subtasks are manually designed and the environment constraints are modified during training. We train a 3D humanoid to walk forward by first training the policy on the sagittal plane and then training in the full 3D space. As a result, the first policy is capable of walking forward while the second policy tries to brings the humanoid back to the sagittal plane when it starts to deviate in the lateral direction. For this example, we allow the policy to switch to non-parent node. This is necessary because while walking forward the humanoid deviates from the sagittal plane many times. 
\end{itemize}

\subsection{Baselines Comparisons}
We compare two versions of our algorithm to the three baselines mentioned above. The first version (AUTO) is exactly the one described in Algorithm \ref{alg:relay_networks}. The second version (MANUAL) requires the user to determine the subtasks and the initial state distributions associated with them. While AUTO presents a cleaner algorithm with less user intervention, MANUAL offers the flexibility to break down the problem in a specific way to incorporate domain knowledge about the problem.

Figure \ref{fig:TestingCurve} shows the \emph{testing curves} during task training. The learning curves are not informative for comparison because the initial state distributions and/or objective functions vary as the training progresses for AUTO, MANUAL, NR, and CL. The testing curves, on the other hand, always computes the average return on the original MDP. That is, the average objective value (Equation \ref{eqn:rootObj}) of rollouts drawn from the original initial state distribution $\rho$. Figure \ref{fig:TestingCurve} indicates that while both AUTO and MANUAL can reach higher returns than  the baselines, AUTO is in general more sample efficient than MANUAL. Further, training the policy to relay to the ``good states'' is important as demonstrated by the comparison between AUTO and NR. The results of CL vary task by task, indicating that relying learning from a warm-started policy is not necessarily helpful.

\begin{figure*}[t]
\centering
\setlength{\tabcolsep}{1pt}
\renewcommand{\arraystretch}{0.7}
\begin{tabular}{c c c c}
  \hline
  \includegraphics[width=0.24\textwidth,height=3cm]{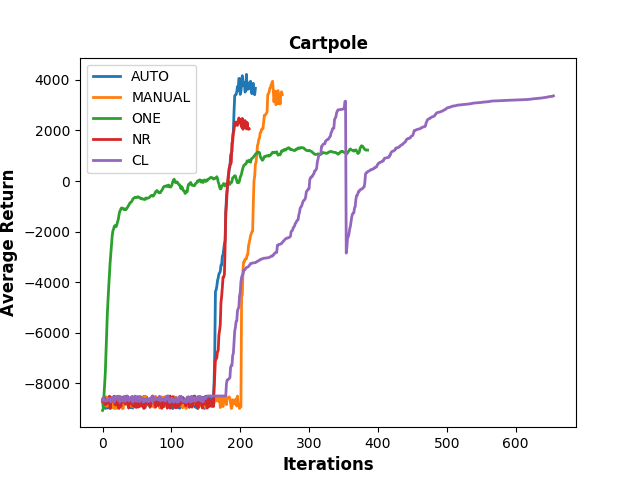}&
  \includegraphics[width=0.24\textwidth,height=3cm]
  {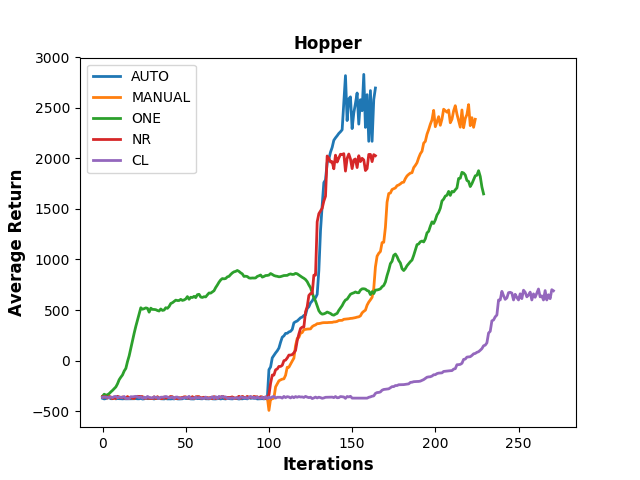}&
  \includegraphics[width=0.24\textwidth,height=3cm]{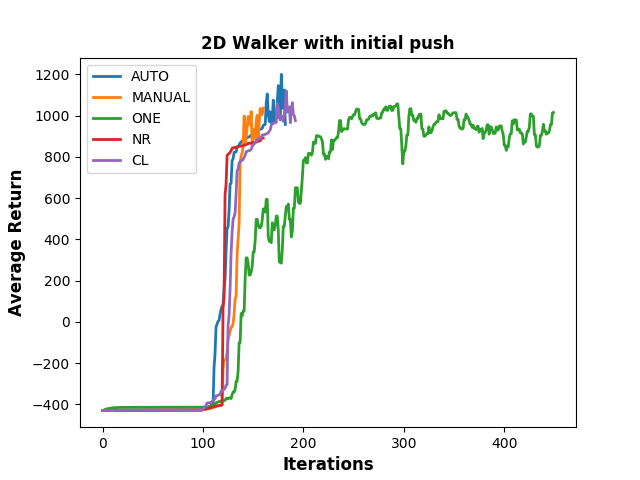}&
  \includegraphics[width=0.24\textwidth,height=3cm]{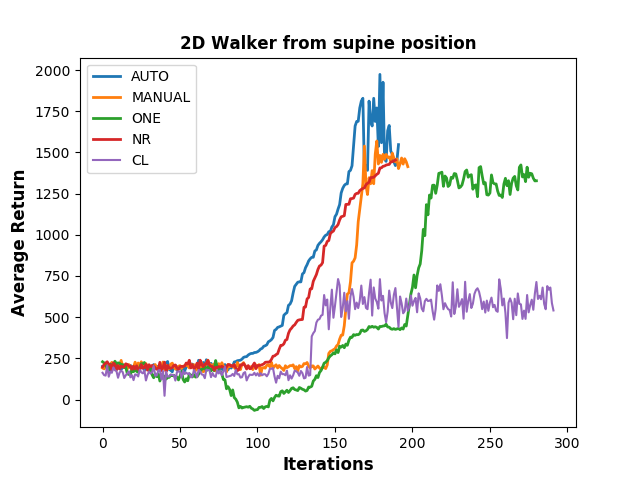} \\ 
  \hline  
\end{tabular}
\caption{ Testing curve comparisons.
    }
\label{fig:TestingCurve}
\end{figure*}

\subsection{Analyses}
\label{sec:alphaComparison}

\textbf{Relay reward:} One important parameter in our algorithm is the weight for relay reward, \ie $\alpha$ in Equation \ref{eqn:relay_obj}.  \ref{OtherFigures}(a) shows that the policy performance is not sensitive to $\alpha$ as long as it is sufficiently large.

\textbf{Accuracy of value function:} Our algorithm relies on the value function to make accurate binary prediction. To evaluate the accuracy of the value function, we generate $100$ rollouts using a learned policy and label them negative if they are terminated by the termination conditions $\mathcal{T}$. Otherwise, we label them positive. We then predict the rollouts positive if they satisfy the condition, $V(\vc{s}) > \bar{V}$. Otherwise, they are negative. \ref{OtherFigures}(c) shows the confusion matrix of the prediction. In practice, we run an additional regression on the value function after the policy training is completed to further improve the consistency between the value function and the final policy. This additional step can further improve the accuracy of the value function as a binary predictor (\ref{OtherFigures}(d)).

\begin{figure*}[t]
\centering
\setlength{\tabcolsep}{1pt}
\renewcommand{\arraystretch}{0.7}
\begin{tabular}{c c c c}
  \hline
  \includegraphics[width=0.26\textwidth,height=3cm]{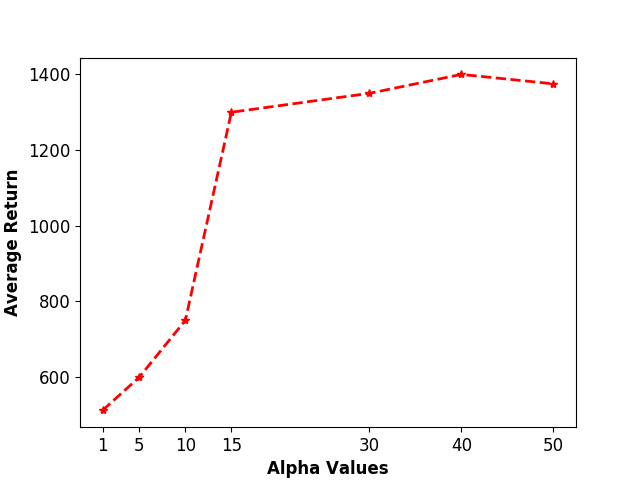}&
  \includegraphics[width=0.26\textwidth,height=3cm]{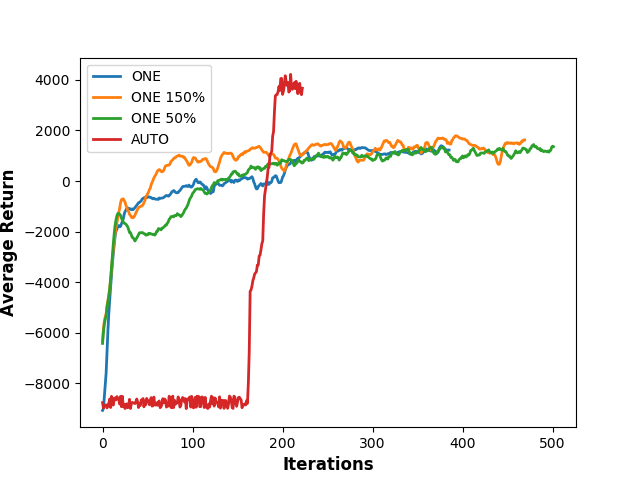}&
  \includegraphics[width=0.2\textwidth]{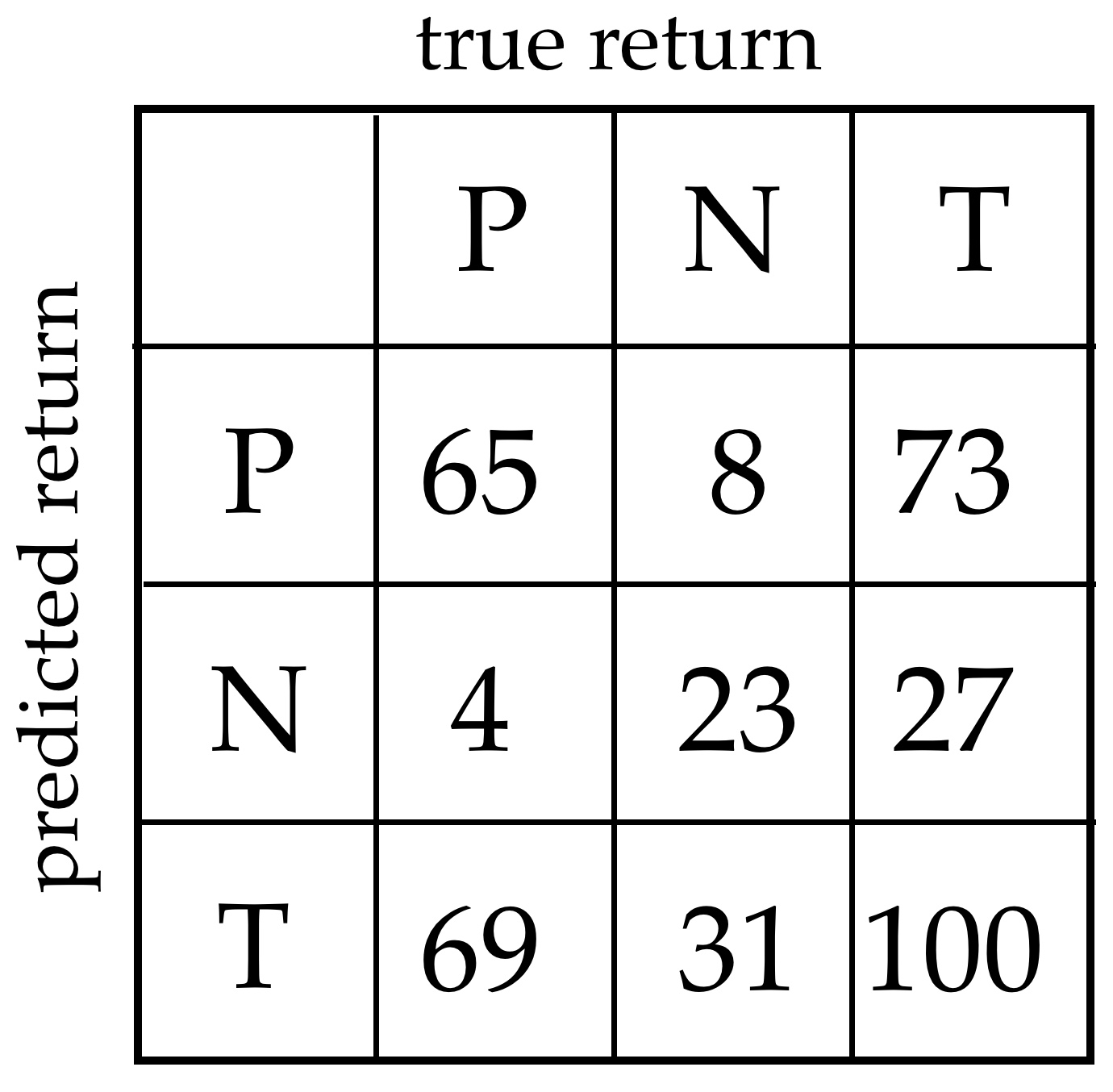}&
  \includegraphics[width=0.2\textwidth]{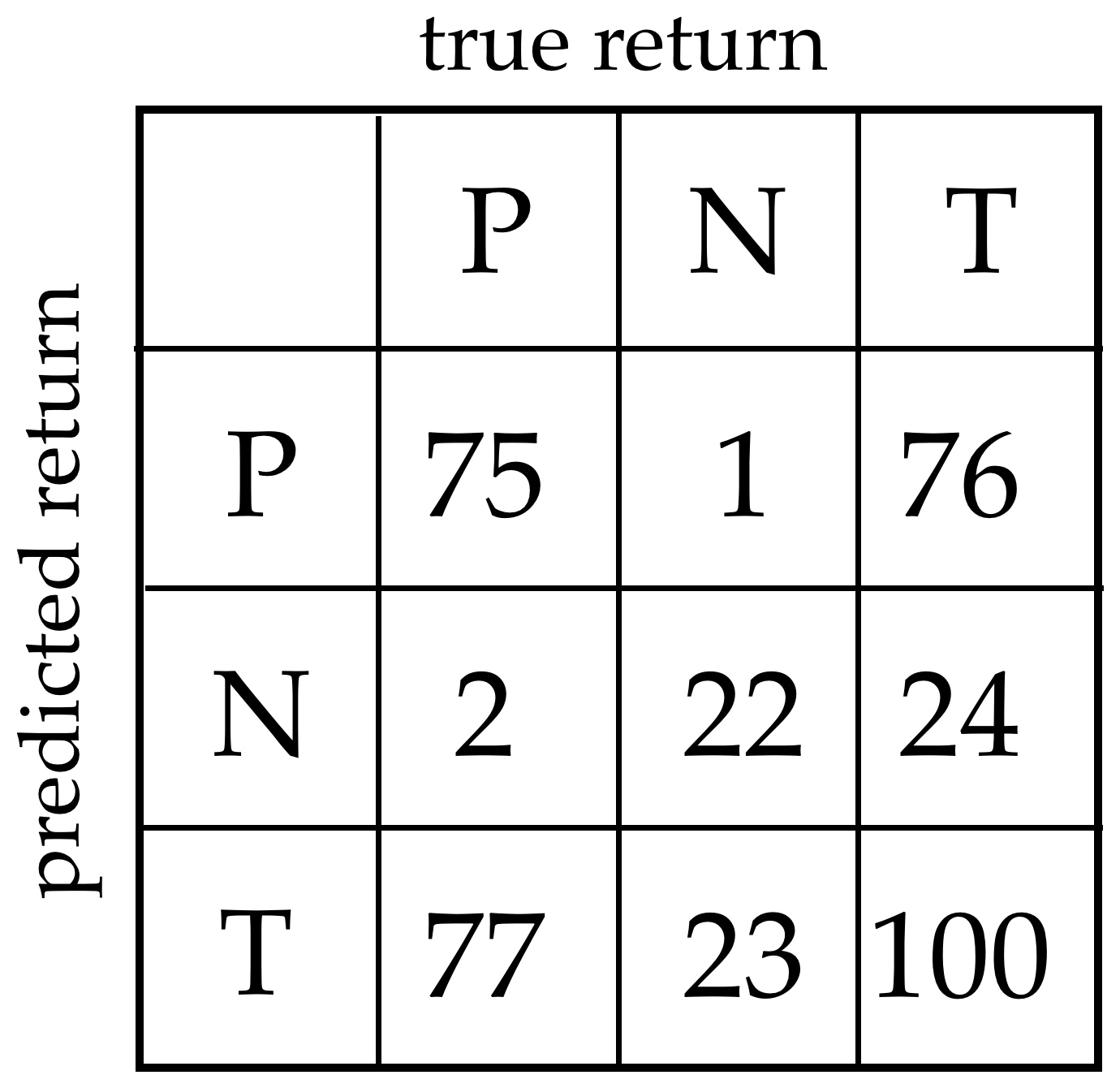} \\
  
 \hline  
\end{tabular}
\caption{ (a) The experiment with $\alpha$. (b) Comparison of ONE with different numbers of neurons. (c) Confusion matrix of value function binary classifier (d) Confusion matrix after additional regression.}
\label{OtherFigures}
\end{figure*}

\subsection{Conclusion}
We propose a technique to learn a robust policy capable of controlling a wide range of state space by breaking down a complex task to simpler subtasks. Our algorithm has a few limitations. The value function is approximated based on the visited states during training. For a state that is far away from the visited states, the value function can be very inaccurate. Thus, the initial state distribution of the child node cannot be too far from the parent's initial state distribution. In addition, as mentioned in the introduction, the relay networks are built on locally optimal policies, resulting globally suboptimal solutions to the original task. The theoretical bounds of the optimality of relay networks can be an interesting future direction.

\chapter{Fall prevention using Assistive Devices}

\section{Motivation}

More than three million older adults every year in the United States are treated for fall injuries. In 2015, the medical costs for falls amounted to more than \$50 billion. Compounding to the direct injuries, fall-related accidents have long-lasting impact because falling once doubles one's chances of falling again. Even with successful recovery, many older adults develop fear of falling, which may make them reduce their everyday activities. When a person is less active, their health condition plummets which increases their chances of falling again.

Designing a control policy to prevent falls on an existing wearable robotic device has multiple challenges. First, the control policy needs to run in real-time with limited sensing and actuation capabilities dictated by the walking device. Second, a large dataset of human falling motions is difficult to acquire and unavailable to public to date, which imposes fundamental obstacles to learning-based approaches. Lastly and perhaps most importantly, the development and evaluation of the fall-prevention policy depends on intimate interaction with human users. The challenge of modeling realistic human behaviors in simulation is daunting, but the risk of testing on real humans is even greater.

We tackle these issues by taking the approach of model-free reinforcement learning (RL) in simulation to train a fall-prevention policy that operates on the walking device in real-time, as well as to model the human locomotion under disturbances. The model-free RL is particularly appealing for learning a fall-prevention policy because the problem involves non-differentiable dynamics and lacks existing examples to imitate. In addition, demonstrated by recent work in learning policies for human motor skills \cite{peng2018deepmimic,YuSIGGRAPH2018}, the model-free RL provides a simple and automatic approach to solving under-actuated control problems with contacts, as is the case of human locomotion. To ensure the validity of these models, we compare the key characteristics of human gait under disturbances to those reported in the biomechanics literature \cite{Winter,Wang2014}. 

Specifically, we propose a framework to automate the process of developing a \textbf{fall predictor} and a \textbf{recovery policy} on an assistive walking device, by only utilizing the on-board sensors and actuators. When the fall predictor predicts that a fall is imminent based on the current state of the user, the recovery policy will be activated to prevent the fall and deactivated when the stable gait cycle is recovered. The core component of this work is a \textbf{human walking policy} that is robust to a moderate level of perturbations. We use this human walking policy to provide training data for the fall predictor, as well as to teach the recovery policy how to best modify the person's gait to prevent falling.  

Our evaluation shows that the human policy generates walking sequences similar to the real-world human walking data both with and without perturbation. We also show quantitative evaluation on the stability of the recovery policy against various perturbation. In addition, our method provides a quantitative way to evaluate the design choices of assistive walking device. We analyze and compare the performance of six different configurations of sensors and actuators, enabling the engineers to make informed design decisions which account for the control capability prior to manufacturing process.

\section{Method}
\label{sec:method}
We propose a framework to automate the process of augmenting an assistive walking device with the capability of fall prevention. Our method is built on three components: a human walking policy, a fall predictor, and a recovery policy. We formulate the problem of learning human walking and recovery policies as Markov Decision Processes (MDPs), $(\mathcal{S}, \mathcal{A}, \mathcal{T}, r, p_0, \gamma)$, where $\mathcal{S}$ is the state space, $\mathcal{A}$ is the action space, $\mathcal{T}$ is the transition function, $r$ is the reward function, $p_0$ is the initial state distribution and $\gamma$ is a discount factor. We take the approach of model-free reinforcement learning to find a policy $\pi$, such that it maximizes the accumulated reward:
\begin{equation}
    J(\pi) = \mathbb{E}_{\mathbf{s}_0, \mathbf{a}_0, \dots, \mathbf{s}_T} \sum_{t=0}^{T} \gamma^t r(\mathbf{s}_t, \mathbf{a}_t),\nonumber
\end{equation}
 where $\mathbf{s}_0 \sim p_0$, $\mathbf{a}_t \sim \pi(\mathbf{s}_t)$ and $\mathbf{s}_{t+1}=\mathcal{T}(\mathbf{s}_t, \mathbf{a}_t)$.

We denote the human walking policy as $\pi_{h}(\vc{a}_{h}|\vc{s}_{h})$ and the recovery policy as $\pi_{e}(\vc{a}_{e}|\vc{s}_{e})$, where $\vc{s}_h$, $\vc{a}_h$, $\vc{s}_e$, and $\vc{a}_e$,  represent the corresponding states and actions, respectively. Our method can be applied to assitive walking devices with any sensors or actuators, though we assume that the observable state $\vc{s}_{e}$ of the walking device is a subset of the full human state $\vc{s}_{h}$ due to sensor limitations. Since our method is intended to augment an assistive walking device, we also assume that the user who wears the device is capable of walking. Under such an assumption, our method only needs to model normal human gait instead of various pathological gaits. Our framework can be applied to any kind of external perturbation which causes a fall. We evaluate our algorithm with pushes to the pelvis, although falls in the real-world are typically not caused by pushes but rather by slips or trips, data for validating our human policy is more readily available for pushing, such as \cite{Wang2014} and \cite{Agarwal}. we consider this as a first-step towards a more general recovery policy.

\begin{figure}[!htb]
\centering
\setlength{\tabcolsep}{1pt}
\renewcommand{\arraystretch}{0.7}
\begin{tabular}{c c}
 
  \includegraphics[width=0.19\textwidth,height=3cm]{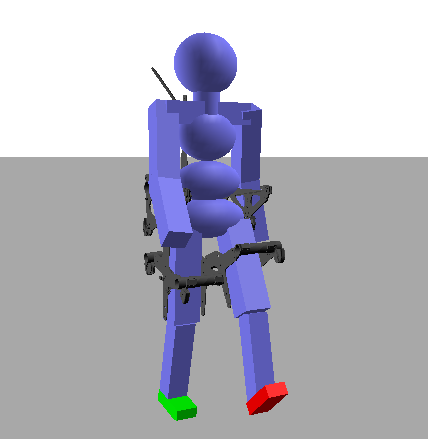}&
  \includegraphics[width=0.19\textwidth,height=3cm]{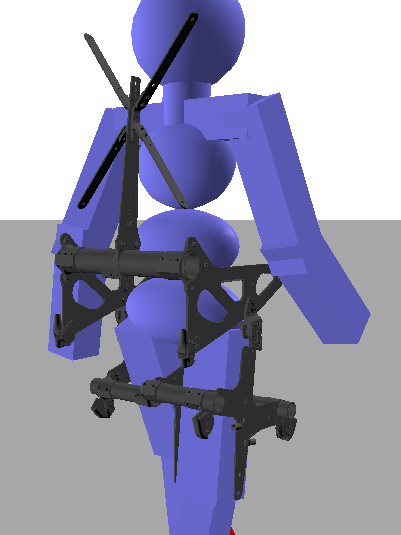} \\ 
  
\end{tabular}
\caption{\textbf{Left :} We model a 29-Degree of Freedom(DoF) humanoid and the 2-DoF exoskeleton in PyDart. \textbf{Right :} Assistive device design used in our experiments. }
\label{fig:Dart}
\end{figure}






\subsection{Human Walking Policy}
\label{sec:human_policy}
We take the model-free reinforcement learning approach to develop a human locomotion policy $\pi_{h}(\vc{a}_{h}|\vc{s}_{h})$. To achieve natural walking behaviors, we train a policy that imitates the human walking reference motion similar to Peng \etal \cite{peng2018deepmimic}. The human 3D model (agent) consists of $23$ actuated joints with a floating base as shown in \figref{Dart}. This gives rise to a $53$ dimensional state space $\vc{s}_{h} = [\vc{q},\vc{\dot{q}},\vc{v}_{com}, \boldsymbol{\omega}_{com},\phi]$, including joint positions, joint velocities, linear and angular velocities of the center of mass (COM), and a phase variable that indicates the target frame in the motion clip. We model the intrinsic sensing delay of a human musculoskeletal system by adding a latency of $40$ milliseconds to the state vector before it is fed into the policy. The action determines the target joint angles $\vc{q}_t^{target}$ of the proportional-derivative (PD) controllers, deviating from the joint angles in the reference motion: 
\begin{equation}
    \label{eq:action}
    \vc{q}^{target}_{t} = \hat{\vc{q}}_{t}(\phi) + \vc{a}_{t},
\end{equation}
where $\hat{\vc{q}}_{t}(\phi)$ is the corresponding joint position in the reference motion at the given phase $\phi$. Our reward function is designed to imitate the reference motion:
\begin{multline}
    \label{eqn:reward}
    r_{h}(\vc{s}_{h},\vc{a}_{h}) =
    w_{q}(\vc{q} - \hat{\vc{q}}(\phi))
    + w_{c}(\vc{c} - \hat{\vc{c}}(\phi)) + w_{e}(\vc{e} - \hat{\vc{e}}(\phi)) - w_{\tau}||\boldsymbol{\tau}||^{2},
\end{multline}
where $\hat{\vc{q}}$, $\hat{\vc{c}}$, and $\hat{\vc{e}}$ are the desired joint positions, COM positions, and end-effector positions from the reference motion data, respectively. The reward function also penalizes the magnitude of torque $\boldsymbol{\tau}$. We use the same weight $w_{q} = 5.0$, $w_{c} = 2.0$, $w_{e} = 0.5$, and $w_{\tau} = 0.005$ for all experiments. We also use early termination of the rollouts, if the agent's pelvis drops below a certain height or if the base rotates about any axis beyond a threshold, we end the rollout and re-initialize the state.

Although the above formulation can produce control policies that reject small disturbances near the target trajectory, they often fail to recover from perturbations with larger magnitude, such as those encountered during locomotion. It is critical to ensure that our human walking policy can withstand the same level of perturbation as a capable real person, so that we can establish a fair baseline to measure the increased stability due to our recovery policy. 

Therefore, we exert random forces to the agent during policy training. Each random force has a magnitude uniformly sampled from $[0,800]\ N$ and a direction uniformly sampled from [-$\pi/2$,$\pi/2$], applied for $50$ milliseconds on the agent's pelvis in parallel to the ground. The maximum force magnitude induces a velocity change of roughly $0.6$m/sec. This magnitude of change in velocity is comparable to experiments found in literature such as \cite{Wang2014},\cite{Agarwal} and \cite{hof2010balance}. We also randomize the time when the force is applied within a gait cycle. Training in such a stochastic environment is crucial for reproducing the human ability to recover from a larger disturbance during locomotion. We represent a human policy as a multi-layered perceptron (MLP) neural network with two hidden layers of $128$ neurons each. The formulated MDP is trained with Proximal Policy Optimization (PPO) [PPO].

\subsection{Fall Predictor}

Being able to predict a fall before it happens gives the recovery policy critical time to alter the outcome in the near future. We take a data-driven approach to train a classifier capable of predicting the probability of the fall in the next $40$ milliseconds. Collecting the training data from the real world is challenging because induced human falls can be unrealistic and dangerous/tedious to instrument. As such, we propose to train such a classifier using only simulated human motion. Our key idea is to automatically label the future outcome of a state by leveraging the trained human policy $\pi_h$. We randomly sample a set of states $\vc{s}_h$ and add random perturbations to them. By following the policy $\pi_h$ from each of the sampled states, we simulate a rollout to determine whether the state leads to successful recovery or falling. We then label the corresponding state observed by the walking device, $(\vc{s}_e, 1)$ if succeeds, or $(\vc{s}_e, 0)$ if fails. We collect about $50000$ training samples. Note that the input of the training data corresponds to the state of the walking device, not the full state of human, as the classifier will only have access to the information available to the onboard sensors.

We train a support vector machine (SVM) classifier with radial basis function kernel to predict if a state $\vc{s}_{e}$ has a high chance of leading to a fall or not. We perform a six-fold validation test on the dataset and the classifier achieves an accuracy above $94$\%.

\subsection{Recovery Policy}
The recovery policy aims to utilize the onboard actuators of the assistive walking device to stabilize the gait such that the agent can continue to walk uninterruptedly. The recovery policy $\pi_e$ is trained to provide optimal assistance to the human walking policy when a fall is detected. The state of $\pi_e$ is defined as $\vc{s}_{e}=[\dot{\boldsymbol{\omega}},\boldsymbol{\omega},\vc{q}_{hip},\dot{\vc{q}}_{hip}]$, which comprises of angular acceleration, angular velocity, and hip joint angle position and velocity, amounting to $10$ dimensional state space. We envision to measure these quantities through an Inertial Measurement Unit(IMU) and a motor encoder at the hip. IMU outputs angular velocity and we would compute angular acceleration using finite difference method. Although, IMU provides us with linear acceleration as well, in practice we observed it has very noisy readings, hence we decided to omit it in our state. The action space consists of torques at two hip joints. The reward function maximizes the quality of the gait while minimizing the impact of disturbance:
\begin{equation}
    \label{eq:exoReward}
    r_{e}(\vc{s}_h,\vc{a}_e) = r_{walk}(\vc{s}_h) - w_{1}\|\vc{v}_{com}\| - w_{2}\|\boldsymbol{\omega}_{com}\| - w_3 \|\vc{a}_e\|,
\end{equation}
where $r_{walk}$ evaluates walking performance using Equation \ref{eqn:reward} except for the last term, and $\vc{v}_{com}$ and $\boldsymbol{\omega}_{com}$ are the global linear and angular velocities of the pelvis. We use the same weight $w_{1} = 2.0$, $w_{2} = 1.2$ and $w_{3} = 0.001$ for all our experiments.  Note that the input to the reward function includes the full human state $\vc{s}_h$. While the input to the recovery policy $\pi_e$ should be restricted by the onboard sensing capability of the assistive walking device, the input to the reward function can take advantage of the full state of the simulated world, since the reward function is only needed at training time. The policy is represented as a MLP neural network with two hidden layers of 64 neurons each and trained with PPO.

\subsection{Results}

We validate the proposed framework using the open-source physics engine DART \cite{lee2018dart}. Our human agent is modeled as an articulated rigid body system with $29$ degrees of freedom (dofs) including the six dofs for the floating base. The body segments and the mass distribution are determined based on a $50$th percentile adult male in North America. We select the prototype of our assistive walking device as the testbed.Similar prototypes are described in \cite{caputo,ExoDesign,witte}. It has two cable-driven actuators at hip joints, which can exert about $200$ Nm at maximum. However, we limit the torque capacity to $30$, beyond this value, the torque saturates. Sensors, such as Inertial Measurement Units (IMU) and hip joint motor encoders, are added to the device. We also introduce a sensing delay of $40$ to $50$ ms. We modeled the interaction between the device and human by adding positional constraints on the thigh and anchor points.  For all experiments, the simulation time step is set to $0.002$s. 

We design experiments to systematically validate the learned human behaviors and effectiveness of the recovery policy. Particularly, our goal is to answer the following questions: 
\begin{enumerate}
    \item How does the motion generated by the learned human policy compare to data in the biomechanics literature? 
    \item Does the recovery policy improve the robustness of the gaits to external pushes?
    \item How does the effectiveness of the assistive walking device change with design choices?
\end{enumerate}

\subsection{Comparison of Policy and Human Recovery Behaviors}

\begin{figure}
\centering
\includegraphics[width=0.5\linewidth]{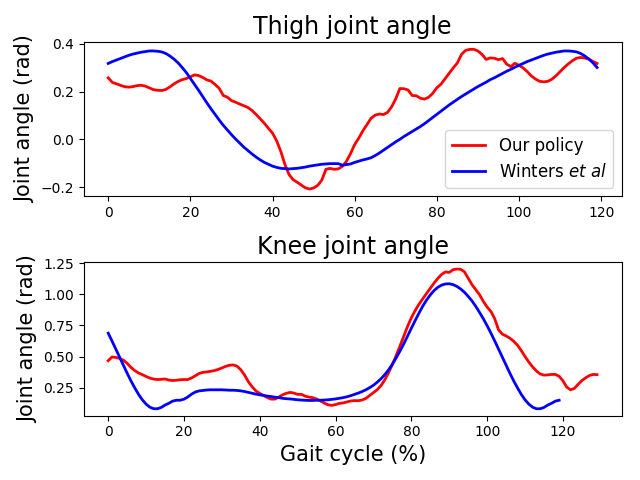}
\caption{Comparison between hip and knee joint angles during walking generated by the policy and human data~\cite{Winter}.}
\label{fig:Jangles}
\end{figure}

\begin{figure}
\center
\setlength{\tabcolsep}{1pt}
\renewcommand{\arraystretch}{0.5}
\begin{tabular}{c | c}
\includegraphics[width=0.5\columnwidth]{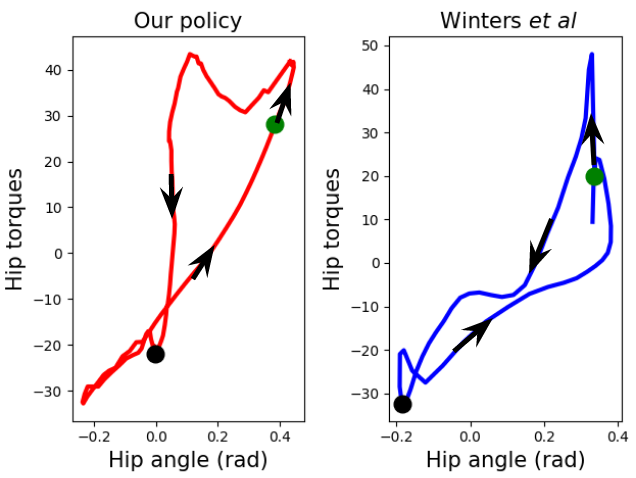} &
\includegraphics[width=0.5\columnwidth]{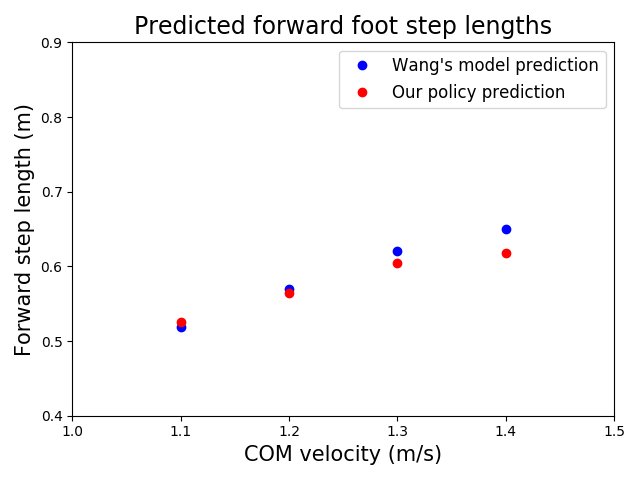} 
\end{tabular}
\caption{(a) Comparison of torque loops of a typical trajectory generated by our policy and human data reported by \cite{Winter} at the hip of stance leg during a gait cycle. The green dots indicate the start and the black dots indicate 50\% of the gait cycle. The arrows show the progression of the gait from 0\% to 100\%. (b) Comparison of the forward foot step locations predicted by the policy and by the model reported by Wang \etal \cite{Wang2014} as a function of the COM velocity.  }
  \label{fig:footplacement}
\end{figure}


We first validate the steady walking behavior of the human policy by comparing it to the data collected from human-subject experiments. \figref{Jangles} shows that the hip and knee joint angles generated by the walking policy well match the data reported in Winter \etal \cite{Winter}. We also compare the ``torque loop'' between the gait generated by our learned policy and the gait recorded from the real world~\cite{Winter}. A torque loop is a plot that shows the relation between the joint degree of freedom and the torque it exerts, frequently used in the biomechanics literature as a metric to quantify human gait. Although the torque loops in \figref{Tauloops} are not identical, both trajectories form loops during a single gait cycle indicating energy being added and removed during the cycle. We also notice that the torque range and the joint angle range are similar. 

In addition, we compare adjusted footstep locations due to external perturbations with the studies reported by Wang \etal \cite{Wang2014}. 
Their findings strongly indicate that the COM dynamics is crucial in predicting the step placement after disturbance that leads to a balanced state. They introduced a model to predict the changes in location of the foot placement of a normal gait as a function of the COM velocity. \figref{footplacement} illustrates the foot placements of our model and the model of Wang \etal against four pushes with different magnitudes in the sagittal plane. For all scenarios, the displacement error is below $4$~cm.


\subsection{Effectiveness of Recovery Policy}

\begin{figure}[!ht]
\centering
\setlength{\tabcolsep}{3pt}
\begin{tabular}{c c c c}
 
  \includegraphics[width=0.11\textwidth,height=2.1cm]{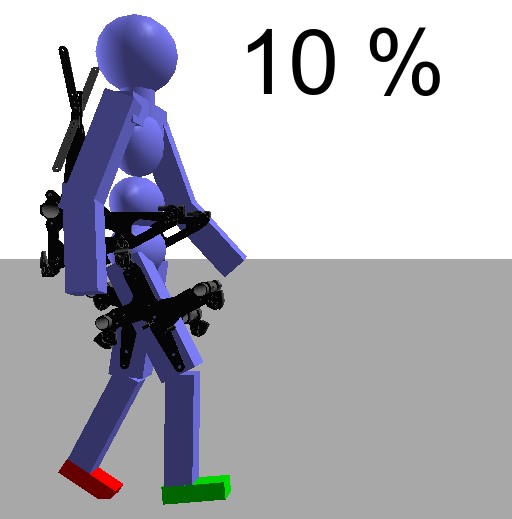}&
  \includegraphics[width=0.11\textwidth,height=2.1cm]{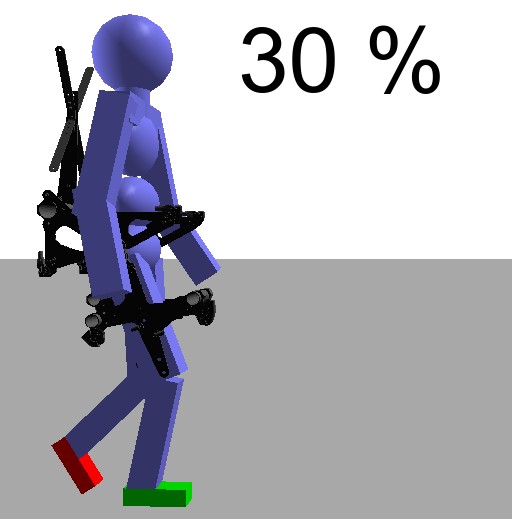}&
  \includegraphics[width=0.11\textwidth,height=2.1cm]{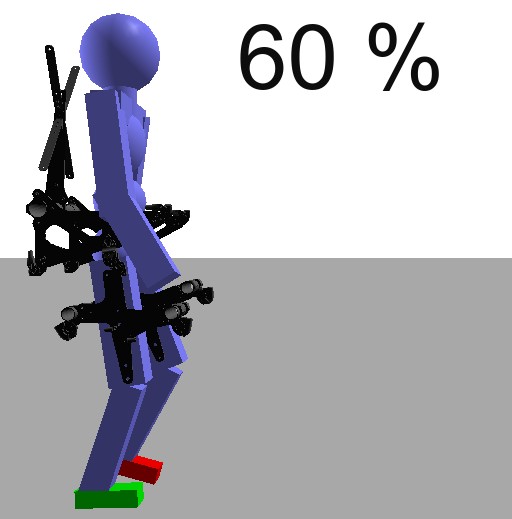}&
  \includegraphics[width=0.11\textwidth,height=2.1cm]{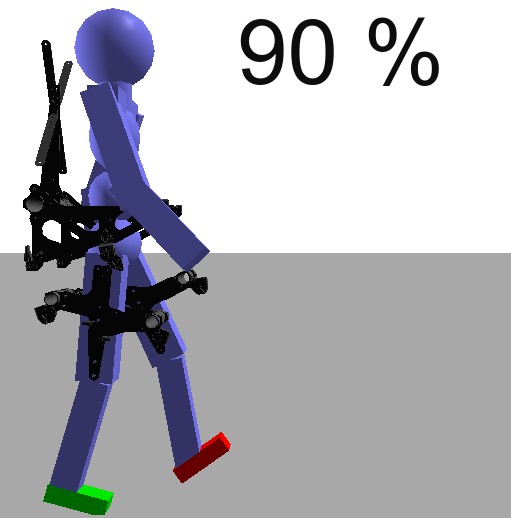} \\ 
  
\end{tabular}
\caption{ Four different timing of the left leg swing phase during which we test the performance of the assistive device. First is at 10\% of the phase and then subsequently 30\%, 60\% and 90\% of the left swing leg.}
\label{fig:phaseDefinition}
\end{figure}

We test the performance of the learned recovery policy in the simulated environment with external pushes. As a performance criterion, we define the \emph{stability region} as a range of external pushes from which the policy can return to the steady gait without falling. For better 2D visualization, we fix the pushes to be parallel to the plane, applied on the same location with the same timing and duration (50 milliseconds). All the experiments in this section use the default sensors and actuators provided by the prototype of the walking device: an IMU, hip joint motor encoders, and hip actuators that control the flexion and extension of the hip.

\begin{figure}
\centering
 \begin{subfigure}{0.5\linewidth}
   \centering
  \includegraphics[width=0.8\linewidth]{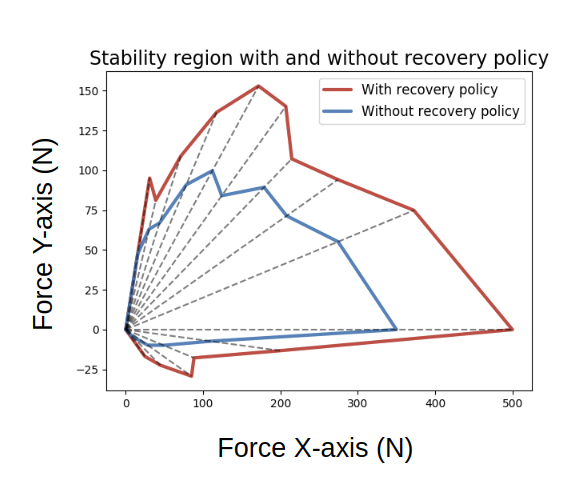}
   \caption{}
 \end{subfigure}

\caption{Stability region with and without the use of a recovery policy. A larger area shows increased robustness to an external push in both magnitude and direction. }
\label{fig:exohelp}
\end{figure}

\figref{exohelp} compares the stability region with and without the learned recovery policy. The area of stability region is expanded by 35\% when the recovery policy is used. Note that the stability region has very small coverage on the negative side of y-axis which corresponds to the rightward forces. This is because we push the agent when the swing leg is the left one, making it difficult to counteract the rightward pushes. Figure \ref{fig:s} shows one example of recovery motion.

\begin{figure}
\center
\setlength{\tabcolsep}{1pt}
\renewcommand{\arraystretch}{0.5}
\begin{tabular}{c | c}
\includegraphics[width=0.5\columnwidth]{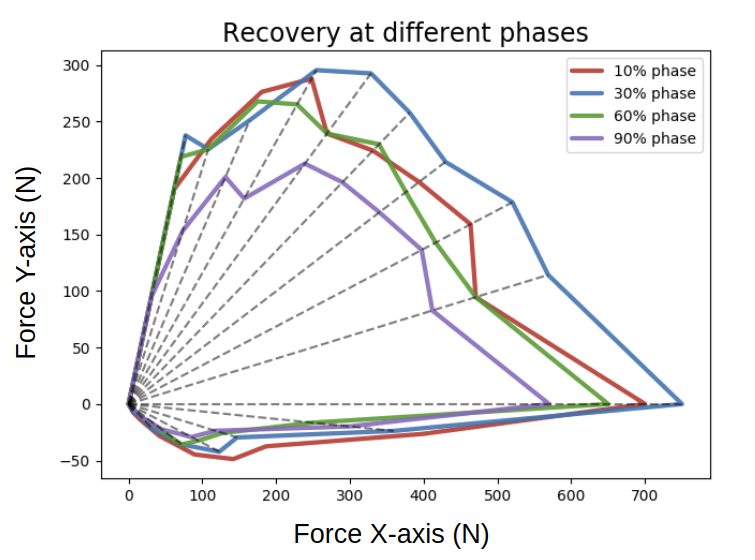} &
\includegraphics[width=0.5\columnwidth]{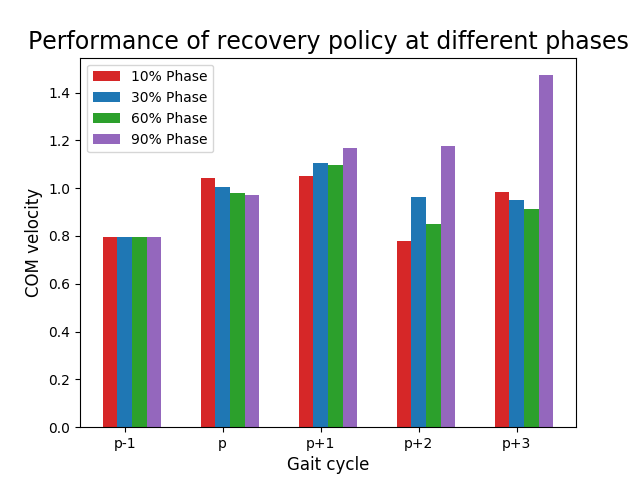} 
\end{tabular}
\caption{Comparison of recovery performance when perturbation is applied at four different phases. \textbf{Top:} Comparison of stability region. \textbf{Bottom:} Comparison of COM velocity across five gait cycles. Perturbation is applied during the gait cycle 'p'. The increasing velocity after perturbation indicates that our policy is least effective at recovering when the perturbation occurs later in the swing phase. }
  \label{fig:PhaseEffect}
\end{figure}


\begin{figure*}
\centering
\setlength{\tabcolsep}{1pt}
\renewcommand{\arraystretch}{0.7}
\begin{tabular}{c c c c c}
 
  \includegraphics[width=0.19\textwidth,height=2.3cm]{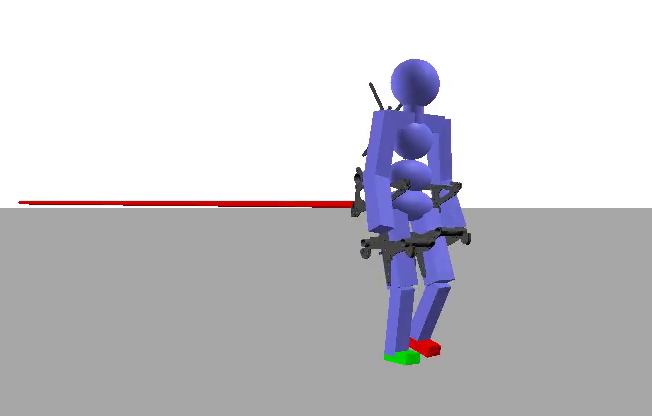}&
  \includegraphics[width=0.19\textwidth,height=2.3cm]
  {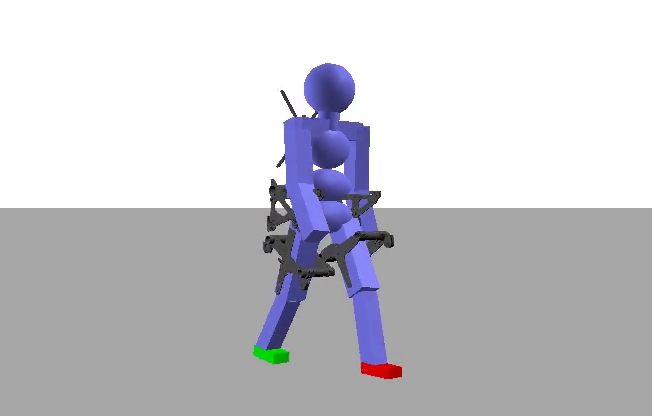}&
  \includegraphics[width=0.19\textwidth,height=2.3cm]{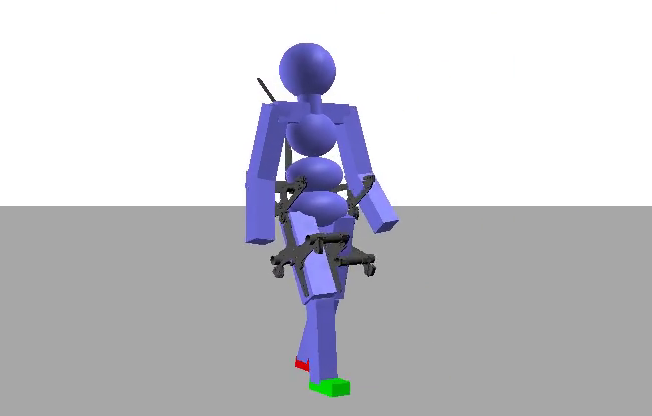}&
  \includegraphics[width=0.19\textwidth,height=2.3cm]{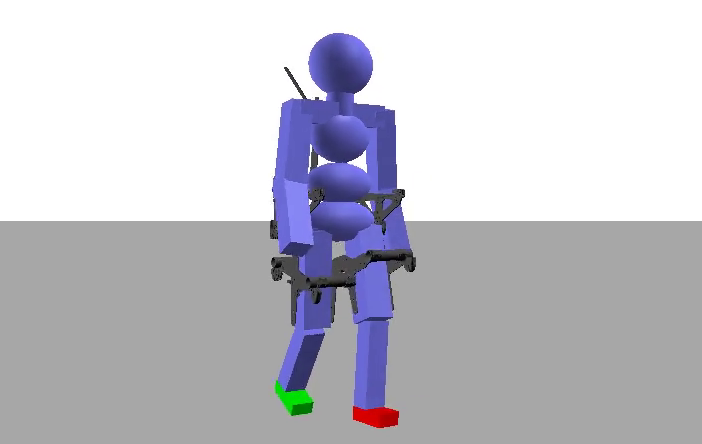}&
  \includegraphics[width=0.19\textwidth,height=2.3cm]{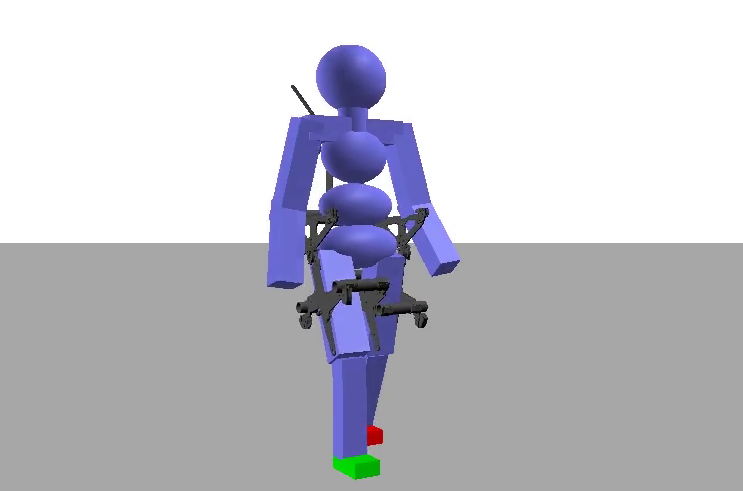} \\ 
  
  \includegraphics[width=0.19\textwidth,height=2.3cm]{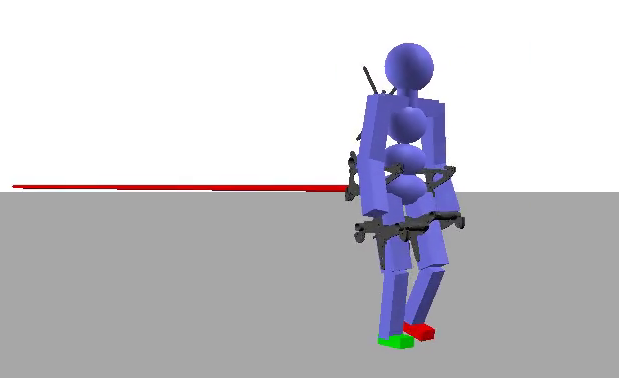}&
  \includegraphics[width=0.19\textwidth,height=2.3cm]
  {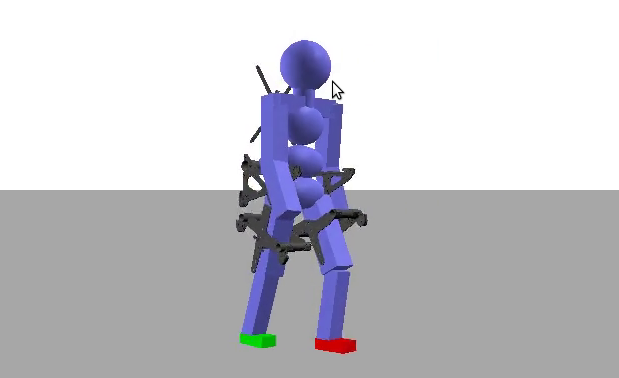}&
  \includegraphics[width=0.19\textwidth,height=2.3cm]{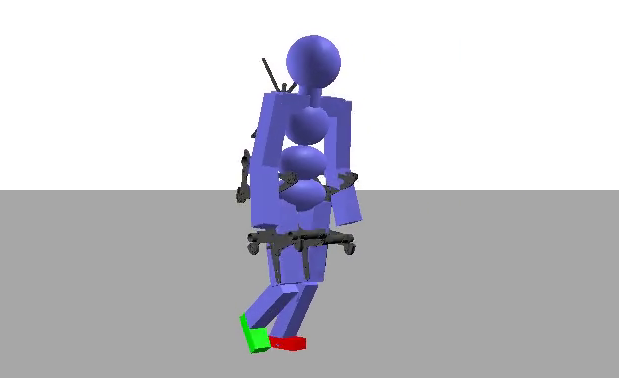}&
  \includegraphics[width=0.19\textwidth,height=2.3cm]{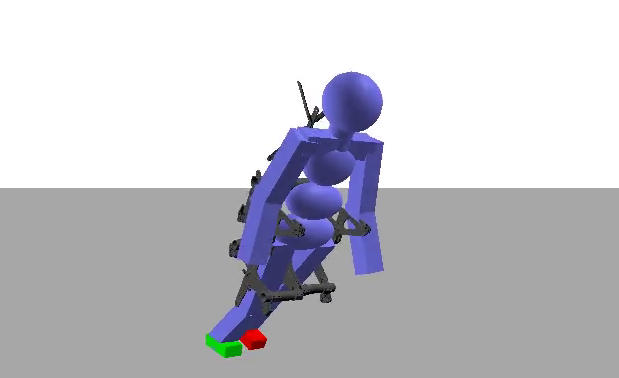}&
  \includegraphics[width=0.19\textwidth,height=2.3cm]{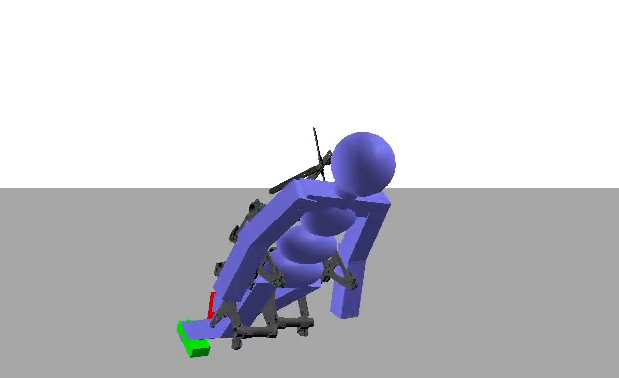} \\
\end{tabular}
\caption{ \textbf{Top:} Successful gait with an assistive device.
\textbf{Bottom:} Unsuccessful gait without an assistive device. Torques are set to zero.}
\label{fig:s}
\end{figure*}

The timing of the push in a gait cycle has a great impact on fall prevention. We test our recovery policy with perturbation applied at four different phases during the swing phase (\figref{phaseDefinition}). We found that the stability region is the largest when the push is applied at 30\% of the swing phase and the smallest at 90\% (\figref{PhaseEffect}, Top). This indicates that the perturbation occurs right before heel strike is more difficult to recover than the one occurs in early swing phase possibly due to the lack of time to adjust the foot location. The difference in the stability region is approximately $28$\%. The bottom of \figref{PhaseEffect} shows the impact of the perturbation timing on COM velocity over four gait cycles. The results echo the previous finding as it shows that the agent fails to return to the steady state when the perturbation occurs later in the swing phase.

\begin{figure}
\centering
\includegraphics[width=0.5\linewidth]{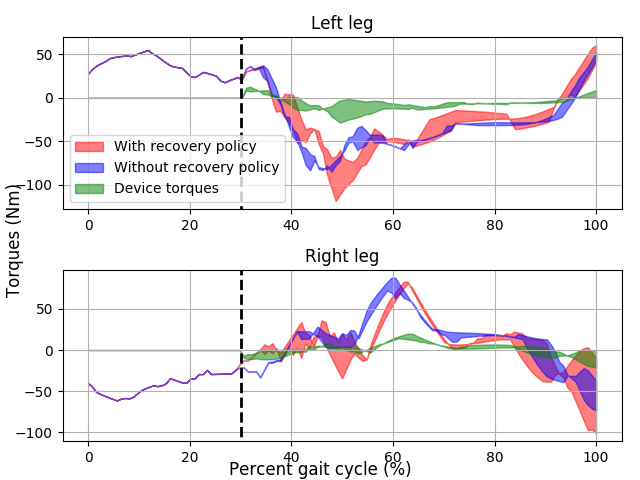}
\vspace{2mm}

\caption{Average torques at the hip joints from $50$ trials with various perturbations. The shaded regions represent the 3-sigma bounds. \textbf{Red:} Joint torques exerted by the human and the recovery policy. \textbf{Blue:} Human joint torques without a recovery policy. \textbf{Green:} Torques produced by a recovery policy.}
\label{fig:Power}
\end{figure}

We also compare the generated torques with and without the recovery policy when perturbation is applied. \figref{Power} shows the torques at the hip joint over the entire gait cycle (not just swing phase). We collect $50$ trajectory for each scenario by applying random forces ranging from $200$N to $800$N at the fixed timing of $30$\% of the gait cycle. The results show that hip torques exerted by the human together with the recovery policy do not change the overall torque profile significantly, suggesting that the recovery policy makes minor modification to the torque profile across the remaining gait cycle, instead of generating a large impulse to stop the fall. We also show that the torque exerted by the recovery policy never exceeds the actuation limits of the device.


\begin{figure}
\centering
\includegraphics[width=0.5\linewidth]{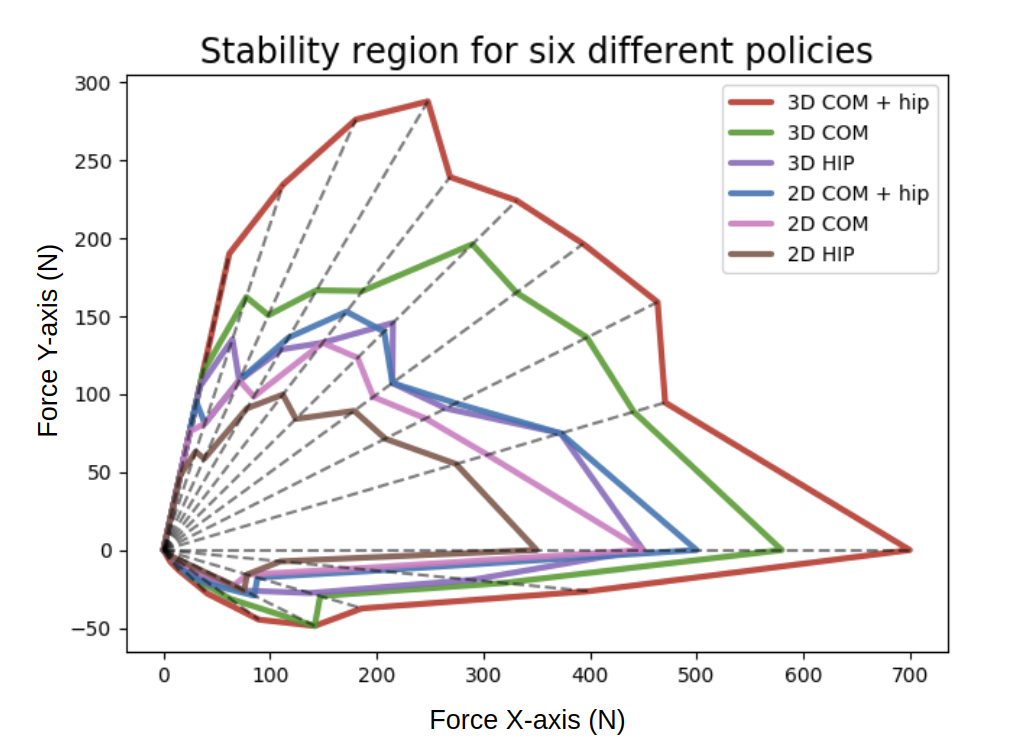}

\caption{Stability region for six policies trained with three sensor configurations and two actuator configurations. }

\label{fig:SensorEffect}
\end{figure}

\subsection{Evaluation of Different Design Choices}




Our method can be used to inform the selection of sensors and actuators when designing a walking device with the capability of fall prevention. We test two versions of actuators: the $2$D hip device can actuate the hip joints only in the sagittal plane while the $3$D device also allows actuation in the frontal plane. We also consider three different configurations of sensors: an inertial measurement unit (IMU) that provides the COM velocity and acceleration, a motor encoder that measures hip joint angles, and the combination of IMU and motor encoder. In total, we train six different recovery policies with three sensory inputs and two different actuation capabilities. For each sensor configuration, we train a fall predictor using only sensors available to that configuration.

\figref{SensorEffect} shows the stability region for each of the six design configurations. The results indicate that 3D actuation expands the stability region in all directions significantly comparing to 2D actuation, even when the external force lies on the sagittal plane. We also found that the IMU sensor plays a more important role than the motor encoder, which suggests that COM information is more critical than the hip joint angle in informing the action for recovery. The recovery policy performs the best when combining the IMU and the joint encoder, as expected.

\subsection{Conclusion}

We presented an approach to automate the process of augmenting an assistive walking device with ability to prevent falls. Our method has three key components : A human walking policy, fall predictor and a recovery policy. In a simulated environment we showed that an assistive device can indeed help recover balance from a wider range of external perturbations. We introduced \emph{stability region} as a quantitative metric to show the benefit of using a recovery policy. In addition to this, \emph{stability region} can also be used to analyze different design choices for an assistive device. We evaluated six different sensor and actuator configurations. 

In this work, we only evaluated the effectiveness of using a recovery policy for an external push. It would be interesting to extend our work to other kinds of disturbances such as tripping and slipping. Another future direction we would like to take is deploying our recovery policy on the real-world assistive device. This would need additional efforts to make sure that our recovery policy also can adjust for the differences in body structure of users.

\chapter{Error-aware policy learning}

\section{Motivation}
Humans exhibit remarkably large variations in gait characteristics during steady-state walking as well as push recovery. Due to this, assistive device controllers tuned for one individual often do not work well when tested on another person. Typically, fine tuning a controller for each person has been the go-to approach thus far, but this process can be a very time consuming ,tedious and unsafe for the user.  

In this work, we introduce a novel approach to tackle sim-to-real problems in which the environment dynamics has high variance and is highly unobservable. While our approach is motivated by physical assistive robotic applications, the method can be applied to other tasks in which many dynamic parameters are challenging to model.  We propose to train a policy explicitly aware of the effect of unobservable factors during training, called an Error-Aware policy (EAP). Akin to the high-level idea of meta learning, we divide the dynamical environments into training and validation sets and "emulate" a reality gap in simulation. Instead of estimating the model parameters that give rise to the emulated reality gap, we train a function that predicts the deviation (i.e. error) of future states due to the emulated reality gaps. Conditioned on the error predictions, the error-aware policies (EAPs) can learn to overcome the reality gap, in addition to mastering the task. 

The main application in this work is to learn an error-aware policy for assistive device control, such as a hip-exoskeleton that helps the user to recover balance during locomotion. From biomechanical data of human gait, we model multiple virtual human walking agents, each varying in physical characteristics as well as parameters that affect the dynamics such as joint damping, torque limits, ground friction, and sensing and actuation delay. We then train a single policy on this group of human agents and show that that the learned EAP works effectively when tested on a different human agent without needing additional data. We extend the prior work, \cite{kumar2019learning}, that trained a control policy for push-recovery assistive device for just one simulated human agent, and develop an algorithm that enables the learned policy to transfer to other human agents with unseen biomechanical characteristics.

We evaluate our approach on assistive wearable device by quantifying the stability and gait characteristics generated by an unseen human agent wearing the device with the trained EAP. We present a comprehensive study of the benefits of our approach over prior zero-shot methods such as universal policy (UP) and domain randomization (DR). We also provide results on some standard RL environments, such as Cartpole, Hopper, 2D walker and a quadrupedal robot.

\section{Method}

We present a method to achieve the zero-shot transfer of control policies in partially observable dynamical environments. We consider robotic systems and environments with unobservable or unmeasurable model parameters, which make building accurate simulation models difficult. 

We present a novel policy architecture, an Error-Aware Policy (EAP), that is explicitly aware of errors induced by unobservable dynamics parameters and self-corrects its actions according to the errors. An EAP takes the current state, observable dynamic parameters, and predicted errors as inputs and generates corrected actions. We learn an additional error-prediction function that outputs the expected error. Both the error-aware policy and the error-prediction function, are iteratively learned using model-free reinforcement learning and supervised learning.

\subsection{Problem Formulation}
We formulate the problem as Partially Observable Markov Decision Processes (PoMDPs), $(S, O, A, P, R, \rho_0, \gamma)$, where $S$ is the state space, $O$ is the observation space, $A$ is the action space, $P$ is the transition function, $R$ is the reward function, $\rho_0$ is the initial state distribution and $\gamma$ is a discount factor. 
In our formulation, we make a clear distinction between observable model parameters $\boldsymbol{\mu}$ and unobservable parameters $\boldsymbol{\nu}$ of the agent and environment. Observable quantities are parameters that can be easily measured such as masses or link lengths, whereas unobserved quantities are challenging to estimate, such as circuit dynamics or backlash.
Therefore, both $\boldsymbol{\mu}$ and $\boldsymbol{\nu}$ affect the transition function $P(\mathbf{s'}|\mathbf{a},\mathbf{s},\boldsymbol{\mu},\boldsymbol{\nu})$.
Since we can configure our simulator with both $\boldsymbol{\mu}$ and $\boldsymbol{\nu}$, we can randomly sample $\boldsymbol{\mu}$ and $\boldsymbol{\nu}$ and create a list of  $K$ different environments $\vc{D}=\{(\boldsymbol{\mu}_0, \boldsymbol{\nu}_0), (\boldsymbol{\mu}_1, \boldsymbol{\nu}_1), \cdots, (\boldsymbol{\mu}_K, \boldsymbol{\nu}_K)\}$, but it is hard to obtain $\boldsymbol{\nu}$ at testing time.
In this case, the transition function will be abbreviated as $P(\mathbf{s'}|\mathbf{s},\mathbf{a},\boldsymbol{\mu})$.

Instead of estimating the values of unobserved quantities, we capture the effect of these parameters by defining a metric called a \emph{state-error}. When transferring from one environment to another, the action $\mathbf{a}$ applied at a given state $\mathbf{s}$  will produce different next states due to the differences in both $\boldsymbol{\mu}$ and $\boldsymbol{\nu}$, in other words, a state-error.


\begin{figure}
\centering
\setlength{\tabcolsep}{1pt}
\renewcommand{\arraystretch}{0.7}
\vspace{5mm}
\includegraphics[width=0.48\textwidth]{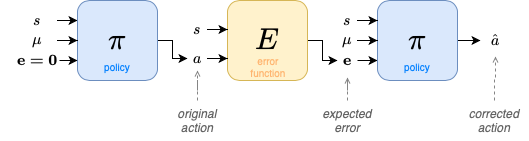}
\vspace{5mm}

\caption{Overview of An Error-aware Policy (EAP). An EAP takes the ``expected'' future state error as an additional input. The expected error is predicted based on the current state $\vc{s}$, observable parameters $\bm{\mu}$, and an uncorrected action $\vc{a}$ that assumes zero error. }
\label{fig:StateError}
\end{figure}

We hypothesize that a policy which is explicitly aware of the state-error would be able to make better decisions by self-correcting its action. We call this an error-aware policy $\pi(\mathbf{a}|\mathbf{s},\boldsymbol{\mu},\mathbf{e})$ (EAP),  which takes in observable parameters $\boldsymbol{\mu}$ as well as the ``expected'' future state error in a new environment $\mathbf{e}$ as input (Figure \ref{fig:StateError}).

We present a novel training methodology using model-free reinforcement learning that involves learning two functions: an error-aware policy and an error prediction function. First, we learn an error-aware policy that takes the output of error prediction function $E$ as an input and has the ability to generalize to novel environments in a zero-shot manner. Simultaneously, we learn an error-prediction function, which takes as inputs the state $\mathbf{s}$, an uncorrected action $\mathbf{a}$ and observable parameters $\boldsymbol{\mu}$, and outputs the expected state error $\mathbf{e}$ when a policy trained in one environment is deployed to a different one $E: (\mathbf{s}, \mathbf{a}, \boldsymbol{\mu}) \mapsto \mathbb{R}^n$.
We will discuss more details of training in the following sections.

\begin{algorithm}
\caption{Train an Error Aware Policy.}
\label{alg:eap}
\begin{algorithmic}[1]
\STATE \textbf{Input:} Environments $\vc{D}=\{(\boldsymbol{\mu}_0, \boldsymbol{\nu}_0), \cdots, (\boldsymbol{\mu}_K, \boldsymbol{\nu}_K)\}$

\STATE Pre-train $\pi(\mathbf{a}|\mathbf{s},\boldsymbol{\mu}_0,\mathbf{e}=0)$ for $P(\vc{s'}|\vc{s},\vc{a},\boldsymbol{{\mu}}_0)$ reference environment with $\vc{e}=0$
\WHILE {not done}
\STATE Sample an environment with $(\boldsymbol{\mu}, \boldsymbol{\nu})$ from $\vc{D}$
\FOR{ each policy update iteration}
\STATE Initialize buffer $\vc{B}=\{\}$ 
\STATE Update an error function $E$ using Algorithm~\ref{alg:errorfunc}
\STATE $\vc{B}$ = Generate rollouts using Algorithm~\ref{alg:rollout} 
\STATE Update policy $\pi$ using $\vc{B}$ with PPO.

\ENDFOR \label{line:trainingSetEnd}
\ENDWHILE
\RETURN{ $\pi(\mathbf{a}, |\mathbf{s}, \boldsymbol{\mu}, \mathbf{e})$}
\end{algorithmic}
\end{algorithm}
\subsection{Training an Error-aware Policy}
\noindent\textbf{Training Procedure.} The training process of an error-aware policy is summarized in Algorithm~\ref{alg:eap}.
Assume that we have an oracle error function $E(\mathbf{s}, \mathbf{a}, \boldsymbol{\mu})$ that outputs the expected state error in a novel environment, which will be explained in the following section. First, the policy is pre-trained to achieve the desired behavior only in the reference environment $(\boldsymbol{\mu}_0, \boldsymbol{\nu}_0)$ assuming there is no state error, $\pi(\mathbf{a}|\mathbf{s},\boldsymbol{\mu}_0,\mathbf{e}=0)$. Once the policy is trained in the reference environment, we sample dynamics parameters $\boldsymbol{\mu}_i$ and $\boldsymbol{\nu}_i$ ($i>0$) uniformly from the data set $\vc{D}$  and evaluate the EAP in this new environment. The policy parameters are updated using a model-free reinforcement learning algorithm, Proximal Policy Optimization~\cite{schulman2017proximal}.
Sampling new testing environments and updating policy parameters are repeated until the convergence.

\begin{algorithm}
\caption{Generate Rollouts}
\label{alg:rollout}
\begin{algorithmic}[1]
\STATE \textbf{Input:} Observable dynamics parameters $\bm{\mu}$, Transition function $P$, Current policy $\pi$ and error function $E$, Replay buffer $\vc{B}$ 
\STATE Sample state $\vc{s}$ from initial state distribution $\rho_0$ 
\WHILE{not done}
\STATE $\vc{a} \sim \pi(\vc{a}|\vc{s},\bm{{\mu}},\vc{e}=0)$ \tcp{original action}
\STATE $\vc{e} = E(\vc{s}, \vc{a}, \bm{{\mu}})$ \tcp{predicted error}
\STATE $\hat{\vc{a}} \sim \pi(\vc{a}|\vc{s},\bm{{\mu}},\vc{e})$ \tcp{error-aware action}
\STATE $\vc{s}'\sim  P(\vc{s}'|\vc{s}, \hat{\vc{a}}, \bm{\mu})$ 
\STATE $r = R(\vc{s},\hat{\vc{a}})$ 
\STATE $B = B \cup \{(\vc{s},\hat{\vc{a}},r,\vc{s}',\bm{\mu})\}$
\STATE $\vc{s}=\vc{s}'$
 
\ENDWHILE
\RETURN $B$
\end{algorithmic}

\end{algorithm}

\noindent\textbf{Rollout Generation.}
A roll-out generation procedure is described in Algorithm~\ref{alg:rollout}.
Given a state $\mathbf{s}$ in this environment $(\boldsymbol{\mu}, \boldsymbol{\nu})$, we query an action from policy $\pi$ as if the policy is being deployed in the reference environment with $\mathbf{e} = 0$. This action $\vc{a}$  is fed into the error function $E$ which predicts the expected state error in this environment, then the state error is  passed into the error-aware policy to query a corrected action $\hat{\vc{a}}$ which will be applied to the actual system. The task reward $R(\vc{s},\vc{a})$ guides the policy optimization to find the best ``corrected'' action that maximizes the reward.

\begin{algorithm}
\caption{Train an Error Prediction Function.}
\label{alg:errorfunc}
\begin{algorithmic}[1]
\STATE \textbf{Input:} Reference environment with $\bm{\mu_0}$
\STATE \textbf{Input:} Target environment with $\bm{\mu}$
\STATE \textbf{Input:} Replay Buffer $\vc{B}$
\STATE \textbf{Input:} Dataset $\vc{Z}$
\STATE \textbf{Input:} Error Horizon $T$

\WHILE{not done}
\STATE Sample the initial state $\vc{s}_0^0$ from $\vc{B}$
\STATE $\vc{s}^0 = \vc{s}_0^0$

\FOR{$t = 0:T-1$}
\STATE \text{}\tcp{Simulation in Reference Env}
\STATE $\vc{a}^{t}_0 \sim \pi(\vc{a}|\vc{s}_0^t,\bm{\mu}_0, \vc{e}=0)$ 
\STATE $\vc{s}^{t+1}_0 \sim P(\vc{s}^t_0, \vc{a}^{t}_0 ,\bm{\mu}_0)$


\STATE \text{}\tcp{Simulation in Validation Env}
\STATE $\vc{a}^{t} \sim \pi(\vc{a}|\vc{s}^t,\bm{\mu}, \vc{e}=0)$ 
\STATE $\vc{s}^{t+1} \sim P(\vc{s}^t, \vc{a}^{t}, \bm{\mu})$

\ENDFOR
\STATE $\vc{Z} = \vc{Z} \cup \{(\vc{s^0},\vc{a^0},\vc{s}^T, \vc{s}_0^T,\bm{\mu})\}$
\ENDWHILE
\STATE minimize the $L(\phi)$ in Eq.~\ref{eq:error_prediction_loss} using $\vc{Z}$.

\RETURN $\phi$
\end{algorithmic}

\end{algorithm}

\subsection{Training an Error Function} \label{sec:error_func}
In reality, we do not have an oracle error function that can predict the next state due to the lack of unobservable parameters $\boldsymbol{\nu}$. To this end, we will learn this function simultaneously with EAP, by splitting the dataset $\vc{D}$ into the training and validation sets. Similar to training methodology followed in meta-learning algorithms, we repeatedly apply the trained policy into sampled environments from the validation set. Because our nominal behavior is pre-trained in the reference environment $(\boldsymbol{\mu}_0, \boldsymbol{\nu}_0)$, we compute the errors by measuring the differences in the reference environment $(\boldsymbol{\mu}_0, \boldsymbol{\nu}_0)$ and the validation environment $(\boldsymbol{\mu}, \boldsymbol{\nu})$: $\bm{e} = (\vc{\bar{s}}' - \vc{s}') \in \mathbb{R}^n$, generated by two dynamic models ${P}(\vc{s}'|\vc{s}, \vc{a}, \bm{{\mu}}_0)$ and $\bar{P}(\vc{s}'|\vc{s},\vc{a}, \bm{\mu})$. 

\noindent \textbf{Horizon of Error Prediction.} In practice, we found that the error accumulated during one step is often not sufficient to provide useful information to the EAP. To overcome this challenge, we take the state in the collected trajectory and further simulate it for $T$ steps in both the reference environment ${P}(\vc{s}'|\vc{s}, \vc{a}, \bm{{\mu}}_0)$ and the validation environment $\bar{P}(\vc{s}'|\vc{s},\vc{a}, \bm{\mu})$. 
We provide analysis on the effect of horizon length from $T=1$ to $T=8$ in the Section \ref{sec:experiment}. 

\noindent \textbf{Loss Function.} Since the differences between the two dynamical environments reflects the reality gap caused by unobservable parameters, the error prediction function $E$ enables us to learn the effect of the unobserved parameters captured through the state error. We train our error prediction function $E$ to learn this ``emulated'' sim-to-real gap by minimizing the following loss:
\begin{equation} \label{eq:error_prediction_loss}
    L(\bm{\phi}) = \sum_{(\vc{s^0},\vc{a^0},\vc{s}^T, \vc{s}_0^T,\bm{\mu})\in \vc{Z}}  ||E(\vc{s}^0, \vc{a}^0,\bm{\mu})  - (\vc{s}_0^T - \vc{s}^T)||^2,
\end{equation}
where $\vc{Z}$ is the collected dataset and $\bm{\phi}$ is the parameters or the neural net representing $E$.
Algorithm~\ref{alg:errorfunc} summarizes the training procedure.

\begin{figure}
\centering
\setlength{\tabcolsep}{1pt}
\renewcommand{\arraystretch}{0.7}
  \includegraphics[width=0.48\textwidth]{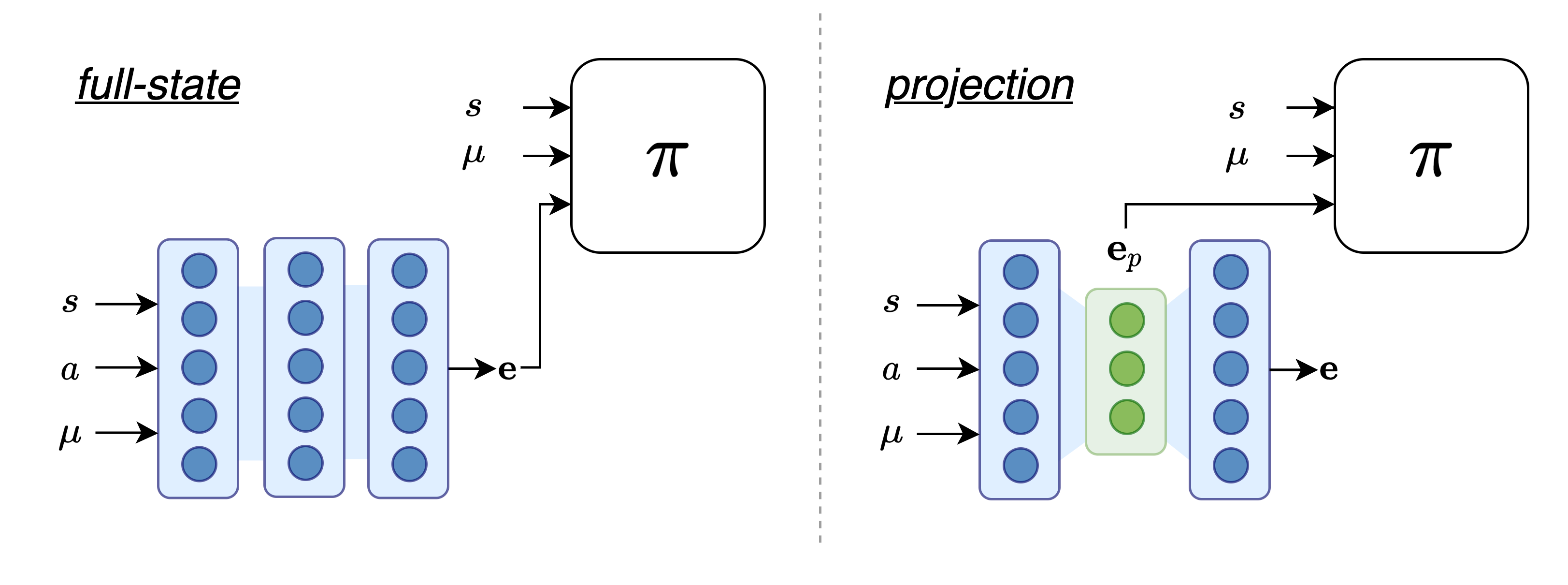}
\caption{\small{\textbf{Left :} A full state error representation input into the policy vs \textbf{Right :} Projected error representation as an input to the policy }}
\label{fig:ErrorRep}
\end{figure}

\noindent \textbf{Reduced Representations.} We experiment with two different representations of the error input to the policy. First, we input the full state error $\vc{e}=\vc{s}_0^T - \vc{s}^T$(with the same dimension as the state) approximated by a MLP neural network, into the policy. Second, we use a network architecture with an information bottle neck, as illustrated in Figure \ref{fig:ErrorRep}, and input the latent representation $\vc{e}_p$ 
into the policy. The same loss function $L$ is used to train both the functions.
 

\section{Results}

We design experiments to validate the performance of error-aware policies. We aim to answer the following research questions.
\begin{enumerate}
    \item Does an EAP show better zero-shot transfer on unseen environments compared to the baseline algorithms?
    \item How does the choice of hyperparameters affect the performance of an EAP? 
\end{enumerate}


\subsection{Baseline Algorithms}
We compare our method with two baselines commonly used for sim-to-real policy transfer, Domain Randomization (DR)\cite{openai2018learning,Pengdr} and Universal Policy (UP) \cite{yu2017preparing}.
DR aims to learn a more robust policy for zero-shot transfer, by training with randomly sampled dynamics parameters (in our case, both $\bm{\mu}$ and $\bm{\nu}$). UP extends DR by taking dynamics parameters as additional input. UP often transfer to target environments better than DR, but it explicitly requires to know dynamics parameters, where $\bm{\nu}$ is assumed to be unobservable in our scenario.
We did not compare EAPs against meta-learning algorithms \cite{belkhale2020modelbased,finn2017modelagnostic,ignasi}, which require additional samples from the validation environment.


\subsection{Tasks}

\begin{table*}
\centering
\caption{Tasks and Network Architectures 
}
\resizebox{\columnwidth}{!}{%
\begin{tabular}{|l|l|l|l|l|}
\hline
Task & Observable Params. $\bm\mu$ & Unobservable Params. $\bm\nu$  & Net. Arch. & Err. Dim. $|\vc{e}_p|$         \\ \hline

Assitive Walking  & mass, height, leg length, and foot length & joint damping, max torques, PD gains and delay &  (64, 32) & 6         \\ \hline
Aliengo & PD gains, link masses & sensor delay, joint damping, ground friction &  (64, 32) & 6         \\ \hline
Cartpole & pole length, pole mass, cart mass  & joint damping, joint friction &   (32, 16) & 2     \\ \hline
Hopper & thight mass, foot mass, shin length & joint damping, ground friction &   (32, 16) & 4       \\ \hline
Walker 2D & link masses, shin length & sensing delay, joint damping, ground friction & (64, 32) &   5        \\ \hline

\end{tabular}%
}

\label{tab:params}

\end{table*}

\begin{table}[]
\begin{center}
    
\caption{Ranges of variation for observable parameters during training and testing in the assistive walking task.}
\begin{tabular}{|l|l|l|}
\hline
\multicolumn{3}{|l|}{          Observable Params. $\bm\mu$} \\ \hline
  Parameter     &  Training range     & Testing range  \\ \hline
    Mass   &  [45,76] kg     &  [55,95] kg      \\ \hline
    Height   & [143,182] cm      & [155,197] cm      \\ \hline
    Leg-length   & [70,88] cm      &  [80,95] cm     \\ \hline
    foot length  & [21,24] cm      &    [24,26] cm    \\ \hline
\end{tabular}

\label{tab:obs_ranges}
\end{center}
\end{table}

\begin{table}[]
\begin{center}
\caption{Ranges of variation for unobservable parameters during training and testing in the assistive walking task.}
\begin{tabular}{|l|l|l|}
\hline
\multicolumn{3}{|l|}{Unobservable Params. $\bm\nu$} \\ \hline
  Parameter     &  Training range      & Testing range  \\ \hline
    Joint damping   &  [0.3,0.6]     &  [0.5,0.8]     \\ \hline
    Max torques   &  [120,180]      &   [155,200]    \\ \hline
    PD gains (P,D)   &    [(500,25),(750,50)]   & [(650,30),(800,50)]      \\ \hline
    Delay  &    [30,60] ms   &  [45,70] ms      \\ \hline
\end{tabular}

\label{tab:unobs_ranges}
\end{center}

\end{table}

We evaluate the performance of error-aware policies on five different tasks. The first task is about push-recovery of an assistive walking device for simulated humans, inspired by the work of Kumar et al~\cite{kumar2019learning}.
The second task is locomotion of a quadrupedal robot, Aliengo Explorer\cite{UnitreeAliengo}.
The rest three tasks are CartPole, Hopper, and Walker2D, which are from the OpenAI benchmark suite \cite{GymAI}.

\subsubsection{Assistive walking device for push recovery}
\begin{figure}
\centering
\setlength{\tabcolsep}{1pt}
\renewcommand{\arraystretch}{0.7}
\includegraphics[width=0.4\textwidth,height=4cm]{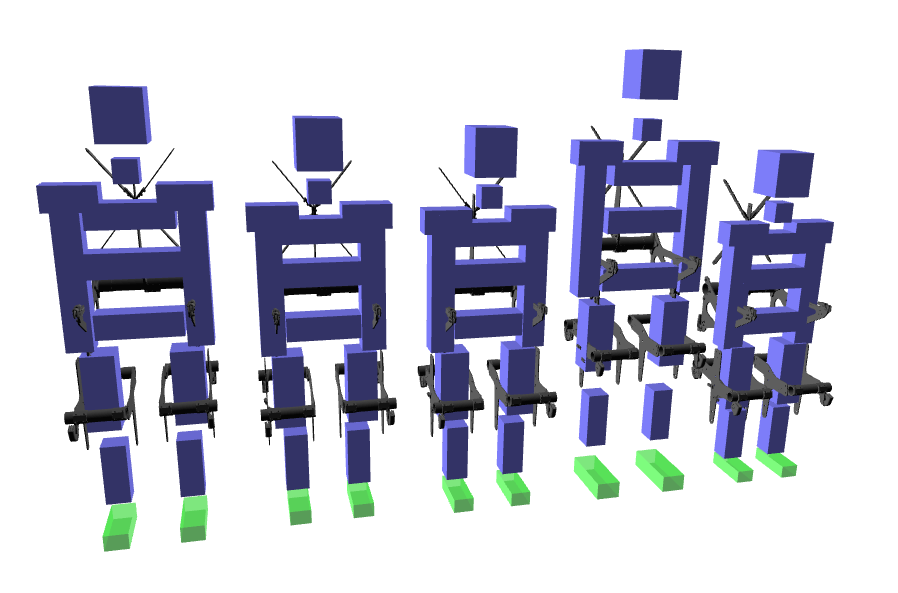}
\caption{Five different test subjects for the assistive walking experiment with varying height, mass, leg length and foot length from the biomechanical gait dataset \cite{dataset2018}.}
\label{fig:TS}
\end{figure}

In this task, the goal is to learn a policy for an assistive wearable device (i.e. exoskeleton) to help a human recover balance after an external push is applied.
(inset figure).
We use a hip exoskeleton that applies torques in 2-degrees of freedom at each hip joint. Our algorithm begins by training $15$ human agents using public biomechanical gait data \cite{dataset2018} to walk in a steady-state gait cycle, similar to the approach presented in \cite{peng2018deepmimic}. The $15$ agents vary in mass, height, leg length, and foot length according to the biomechanical data used to train their corresponding policies, which formulate the four-dimensional observable parameters $\bm{\mu}$ (Figure~\ref{fig:TS}). We also vary each human agent's joint damping, maximum joint torques, PD gains, and sensory delay as the four dimensional unobservable parameters $\bm{\nu}$. We split the $15$ human agents into $10$ for the training set and $5$ as the testing set.

\noindent \textbf{Human Behavior Modeling.}
First, we capture the human behavior by training a human-only walking policy $\pi_h$ that mimics the reference motion which frames are denoted as $\bar{\vc{q}}$.
Each human model has $23$ actuated joints along with a floating base. The state space has $53$ dimensions, $\vc{s}_{h} = [\vc{q},\vc{\dot{q}},\vc{v}_{com}, \boldsymbol{\omega}_{com},\psi]$, which represent joint positions, joint velocities, linear and angular velocities of the center of mass, and a phase variable $\psi$ that indicates the target frame in the reference biomechanical gait cycle. 
The action $\vc{a}$ is defined as the offset to the reference biomechanical joint trajectory $\bar{\vc{q}}(\psi)$, which results in the target angles: $\vc{q}^{target} = \bar{\vc{q}} + \vc{a}$.
The reward function encourages mimicking the reference motions from public biomechanical data: 
\begin{multline}
    \label{eqn:reward}
    R_{human}(\vc{s}_{h},\vc{a}_{h}) =
    w_{q}(\vc{q} - \bar{\vc{q}}) + w_{v}(\vc{\dot{q}} - \vc{\bar{\dot{q}}}) 
    + w_{c}(\vc{c} - \bar{\vc{c}}) + w_{p}(\vc{p} - \bar{\vc{p}}) - w_{\tau}||\boldsymbol{\tau}||^{2},
\end{multline}
where the terms include the reference joint positions $\vc{\bar{q}}$, joint velocities $\vc{\bar{\dot{q}}}$, end-effector locations $\vc{\bar{p}}$, contact flags $\vc{c}$, and the joint torques $\bm{\tau}$. During training,  we exert random forces to the agent during policy training. Each random force has a magnitude uniformly sampled from $[0,800]\ N$ and a direction uniformly sampled from [-$\pi/2$,$\pi/2$], applied for $50$ milliseconds on the agent's pelvis in parallel to the ground. The maximum force magnitude induces a velocity change of roughly $0.6 - 0.8$ m/sec. This magnitude of change in velocity is comparable to experiments found in biomechanics literature such as \cite{Wang2014},\cite{Agarwal} and \cite{hof2010balance}. We also randomize the time when the force is applied within a gait cycle. The forces are applied once at a randomly chosen time in each trajectory rollout. Similar to \cite{kumar2019learning}, we enforce joint torque constraints and introduce sensing delays during training to prevent the human agent to adapt to external disturbance really well.

\noindent \textbf{MDP Formulation.}
Once the human agents are trained, we begin learning the push-recovery EAP for the assistive device.
The objective is to stabilize the human gait from external perturbations.
The $17$ dimensional state of robot is defined as $\vc{s}_{e}=[{\bm{\omega}},\bm{\alpha},\vc{\Ddot{x}},\vc{q}_{hip},\dot{\vc{q}}_{hip}]$, which comprises angular velocity, orientation, linear acceleration, hip joint positions, and hip joint velocity.
The four dimensional action $\vc{a}_e$ consists of torques at two hip joints.
The reward function maximizes the quality of the gait while minimizing the impact of an external push.
\begin{multline}
    \label{eq:exoReward}
    R_{exo}(\vc{s}_h, \vc{s}_e, \vc{a}_e) = R_{human}(\vc{s}_h) - w_{1}\|\vc{v}_{com}\| - w_{2}\|\boldsymbol{\omega}_{com}\| - w_3 \|\vc{a}_e\|,
\end{multline}
where $R_{human}$ is defined in equation \ref{eqn:reward} , and $\vc{v}_{com}$ and $\boldsymbol{\omega}_{com}$ are the global linear and angular velocities of the pelvis. The last term penalizes the torque usage. We use the same weight $w_{1} = 2.0$, $w_{2} = 1.2$ and $w_{3} = 0.001$ for all our experiments. 





\subsubsection{Quadrupedal Locomotion}
In our second task, we learn a control policy that generates a walking motion for a quadrupedal robot, Aliengo Explorer \cite{UnitreeAliengo}. For this task, the $17$ observable parameters ($\bm{\mu}$) are PD gains of the joints, link and root masses and the $10$ unobservable parameters $\bm{\nu}$ include sensing delay, joint damping of thigh and knee joints and ground friction. The 39-dimensional state space consists of torso position and orientation and corresponding velocities, joint position and velocities, foot contact variable that indicates when each foot should be in contact with the ground, while the $12$-dimensional action space consists of joint velocity targets which is fed into a PD controller that outputs torques to each joint.

The reward function is designed to track the target motion that walks at $0.8$~m/s:
\begin{multline}
    r(\vc{s},\vc{a}) = w_1 e^{-k_{1}*(\vc{q} - \vc{\bar{q}})} + w_2 e^{-k_{2}*(\vc{\dot{q}} - \vc{\bar{\dot{q}}})} +  w_3 \min(\dot{x},0.8) + \sum_{i=1}^4 ||c_i - \bar{c}_i||^2.
\label{eqn:quad}
\end{multline}
In this equation, the first term encourages to track the desired joint positions, the second term is to track the desired joint velocities, the third term is for matching the forward velocity $\dot{x}$ to a target velocity of $0.8$ m/s. and the four term tracks the predefined contact flags.
We use the same weight $k_1=35$,$w_1=0.75$, $k_2=2$,$w_2=0.20$, and $w_3=1.0$ for all experiments.

\subsubsection{OpenAI Environments}
We test our method on three OpenAI environments: CartPole, Hopper and Walker2D.
While using the same state spaces, action spaces, and the reward functions described in the benchmark \cite{GymAI}, we additionally define observable and unobservable dynamics parameters as follows: 


\begin{enumerate}
\item \noindent\textbf{Cartpole.} Observable parameters $\bm{\mu} \in R^3$ includes the length of the pole, the mass of the pole, and the mass of cart. Unobservable parameters $\bm{\nu} \in R^3$ include the damping at the rotational joint, the friction at the rotational joint, and the friction at the translational joint. 


\item \noindent\textbf{Hopper.} Observable parameters $\bm{\mu} \in R^3$ include the mass of the thigh and foot and the length of the shin bodynode. Unobservable parameters $\bm{\nu} \in R^3$ include joint damping of shin and foot joints and ground friction.

\item \noindent\textbf{Walker 2D.} Observable parameters $\bm{\mu} \in R^6$ include the masses of thigh and foot for both legs, the mass of pelvis, and the length of shin. Unobservable parameters $\bm{\nu} \in R^4$ include joint damping of foot joints, the delay in observation, and ground friction.
\end{enumerate}

\subsection{Zero-shot Transfer with EAPs}

\begin{figure}
\centering
\begin{subfigure}[b]{0.35\columnwidth}
    \centering
    \includegraphics[width=\columnwidth]{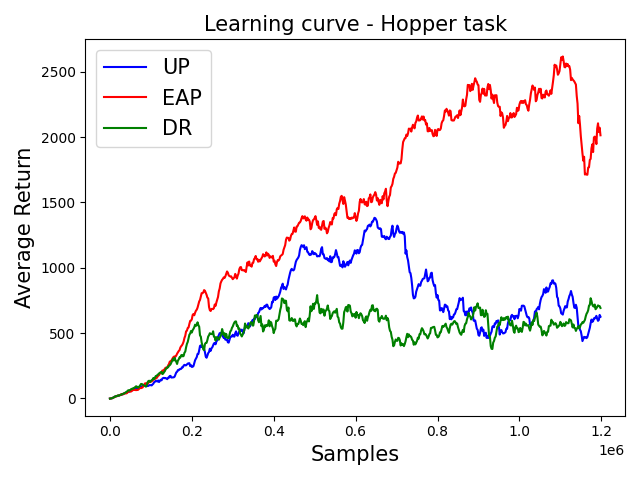}
    \caption[Network2]%
    {{\small Hopper}}    
    \label{fig:mean and std of net14}
\end{subfigure}
\hfill
\begin{subfigure}[b]{0.35\columnwidth}  
    \centering 
    \includegraphics[width=\columnwidth]{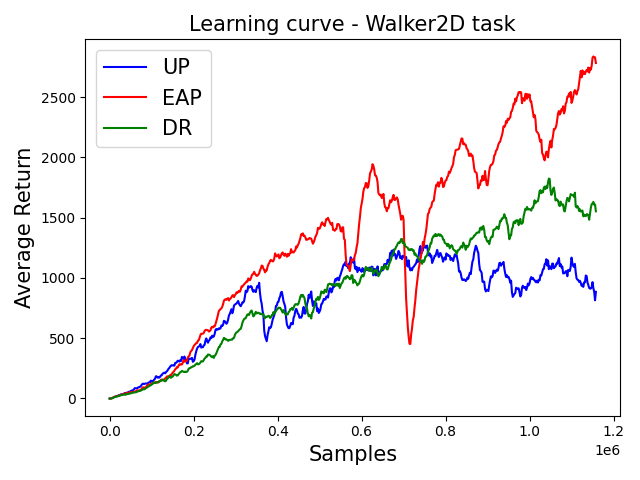}
    \caption[]%
    {{\small Walker 2D}}    
    \label{fig:mean and std of net24}
\end{subfigure}
\vskip\baselineskip
\begin{subfigure}[b]{0.35\columnwidth}   
    \centering 
    \includegraphics[width=\columnwidth]{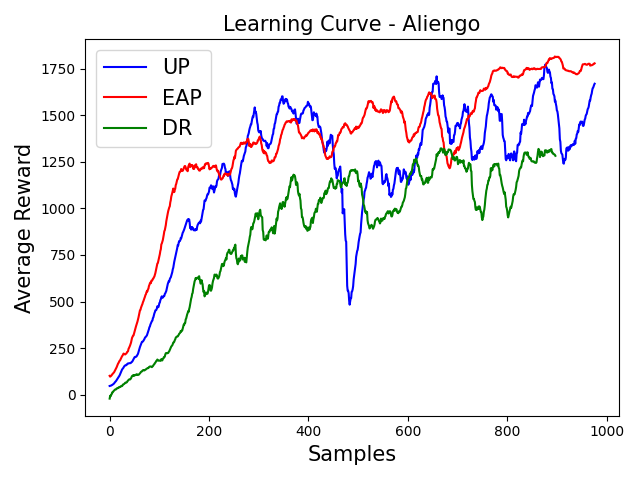}
    \caption[]%
    {{\small Quadrupedal Locomotion}}    
    \label{fig:mean and std of net34}
\end{subfigure}
\hfill
\begin{subfigure}[b]{0.35\columnwidth}   
    \centering 
    \includegraphics[width=\columnwidth]{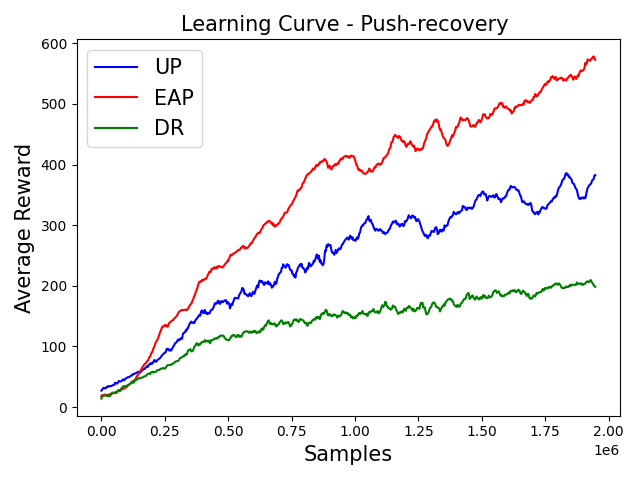}
    \caption[]%
    {{\small {Assistive Walking}}}    
    \label{fig:mean and std of net44}
\end{subfigure}
\caption{\small{ Learning curves for four tasks. The number of samples for EAP include the ones generated for training an error function.}} 
\label{fig:learning_curves}
\end{figure}

In this section, we compare the zero-shot transfer of error-aware policies against two other baseline algorithms, Domain Randomization (DR) and Universal Policies (UP).

\noindent \textbf{Learning Curves.} 
First, we compare the learning curves of the EAP, DR, and UP approaches on four selected tasks in Figure~\ref{fig:learning_curves}.
We set the same ranges of the observable and unobservable parameters for all three algorithms.
In our experience, EAPs learn faster than DR and UP for three tasks, the Hopper, Walker2D, and assistive walking tasks, while showing comparable performance for the quadrupedal locomotion task.
Note that, to make the comparison fair to baselines, we also include the samples for training error functions (Algorithm~\ref{alg:errorfunc}) when we evaluate the performance of EAPs. We do not include the experiment on the CartPole environment for brevity but the EAP outperforms the baselines as well 

\noindent \textbf{Zero-shot Transfer.} 
\begin{figure}
\centering
\setlength{\tabcolsep}{1pt}
\renewcommand{\arraystretch}{0.7}
\includegraphics[width=0.35\textwidth,height=4cm]{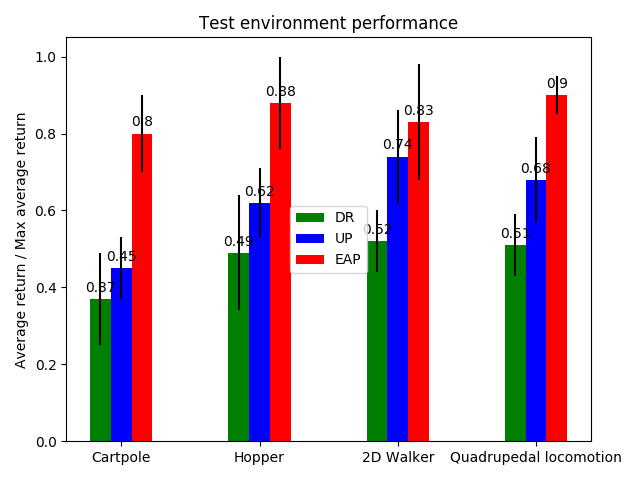}
\caption{\small{Comparison of EAP and baselines DR and UP. The error bars represent the variation in the average return of the policy in the target environment when trained with 4 different seeds.}}
\label{fig:TestPerf}
\end{figure}
Then we evaluate the learned policies on unseen validation environments, where their dynamics parameters $\bm\mu$ and $\bm\nu$ are sampled from the outside of the training range.
We conduct the experiments for the CartPole, Hopper, Walker2D and quadrupedal locomotion tasks and compare the \emph{normalized} average returns (the average return divided by the maximum return).
The results are plotted in Figure~\ref{fig:TestPerf}, which indicate that EAP outperforms DR by $60$\% to $116$\% and UP by $12$\% to $77$\%.
Note that UP may perform well for the real-world transfer due to the lack of the unobservable parameters.
We also observe that UP is consistently better than DR by being aware of the dynamics parameters, $\bm\mu$ and $\bm\nu$, which meets our expectation.

\begin{figure}
\centering
\setlength{\tabcolsep}{1pt}
\renewcommand{\arraystretch}{0.7}
\includegraphics[width=0.35\textwidth,height=4cm]{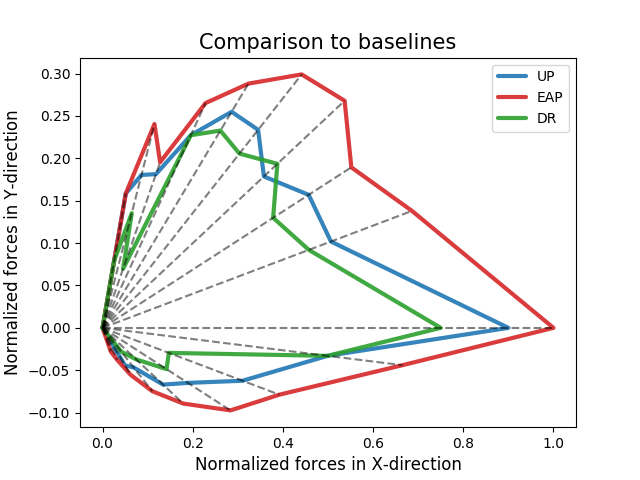}
\caption{Average stability region in five test subjects. The results indicate the better zero-shot transfer of EAP over DR and UP. }
\label{fig:SR_test}
\end{figure}

For evaluating the zero-shot transfer for the assistive walking task, we define an additional metric ``stability region'', which depicts the ranges of maximum perturbations in all directions that can be handled by the human with the EAP-controlled exoskeleton.
We train policies for 10 training human subjects and test the learned policies for 5 new human subjects.
Figure \ref{fig:SR_test} compares the average performance of EAP with DR and UP.
The larger area of stability region indicates that EAP significantly outperforms two baselines. The ranges of variation for observable and unobservable parameters during training and testing phases are included in tables \ref{tab:obs_ranges} and \ref{tab:unobs_ranges}.

\subsection{Ablation study}
We further analyze the performance of EAPs by conducting a set of ablation studies.
We studied four categories of parameters: choices of observable parameters, reference dynamics, error prediction horizons, and error representations.

\begin{figure}
\centering
\setlength{\tabcolsep}{1pt}
\renewcommand{\arraystretch}{0.7}
\includegraphics[width=0.4\textwidth,height=5cm]{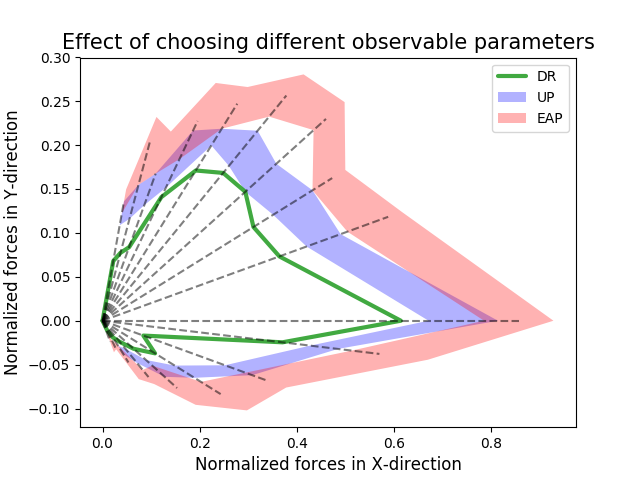}
\caption{\small{Ablation study with choosing different observable parameters as $\bm\mu$. The result indicates that our approach (EAP) shows more reliable zero-shot transfers for all different scenarios.}}
\label{fig:AblationMuo}
\end{figure}

\noindent\textbf{Choice of Observable and Unobservable Parameters.} We check the robustness of EAPs by testing with different choices of observable and unobservable parameters. We randomly split the parameters into $\bm{\mu}$ and $\bm{\nu}$ and test three different splits. 
Figure \ref{fig:AblationMuo} shows the stability regions for all three algorithms for three different scenarios.
In all cases, EAPs are more robust than the baseline algorithms.

\begin{figure}
\centering
\setlength{\tabcolsep}{1pt}
\renewcommand{\arraystretch}{0.7}
 
  \includegraphics[width=0.4\textwidth,height=5cm]{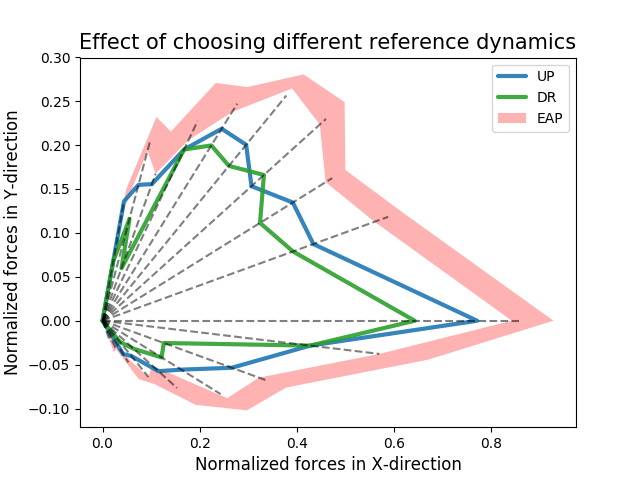}

\caption{\small{Ablation study with different reference dynamics. The results indicate that our algorithm is robust against the choice of different references.}}
\label{fig:ablationRefDyn}
\end{figure}

\noindent\textbf{Choice of Reference Dynamics.} In this study, we analyze the effect of choosing three different reference dynamics $P(\vc{s}' | \vc{a}, \vc{s}, \bm{\mu}_0, \bm{\nu}_0)$ on the performance of EAP. We randomly choose three different human agents as the reference dynamics and follow the learning procedure of EAPs to train three different policies. These policies are then deployed on the same test subjects along with UP and DR policies. Figure \ref{fig:ablationRefDyn} shows that all the EAPs outperforming the baselines by having larger stability regions, although EAPs have slightly larger variances.

\begin{figure}
\centering
\setlength{\tabcolsep}{1pt}
\renewcommand{\arraystretch}{0.7}
\includegraphics[width=0.4\textwidth,height=5cm]{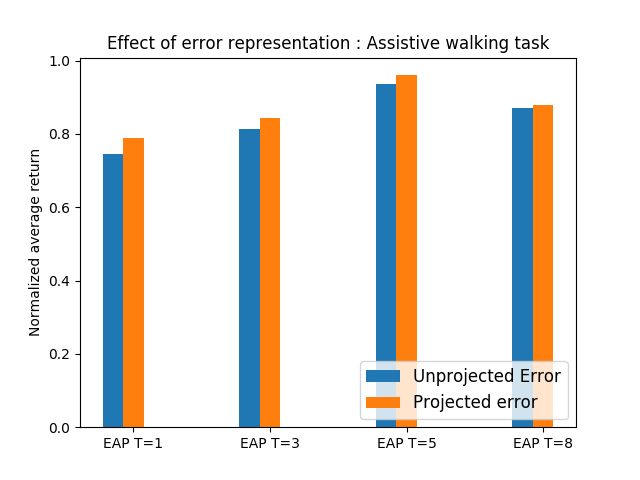}
\caption{\small{Ablation study with different parameter setting for EAP training.}}
\label{fig:ablation}
\end{figure}

\noindent\textbf{Horizon of Error Prediction.} 
As we motivated in Section~\ref{sec:error_func}, one step error might be too subtle to inform the learning of EAPs and we may need $T$ step expansion to enlarge them.
We studied the effect of the error prediction horizon $T$ in Algorithm~\ref{alg:errorfunc} by varying its value from $T=1$ to $T=8$ for the assistive walking task.
Figure \ref{fig:ablation} shows the normalized average return over T gradually changes over the different values of $T$ and peaks at $T=5$. 
Therefore, we set $T=5$ for all the experiments.

\noindent\textbf{The error representation.} 
We also compare the effect of the error representation.
Figure \ref{fig:ablation} also plots the normalized average returns of the unprojected errors (blue) and projected errors (orange), where projected errors show slightly better performance for all the different $T$ values.

\section{Hardware experiments}


\begin{figure}
\centering
\setlength{\tabcolsep}{1pt}
\renewcommand{\arraystretch}{0.7}
\includegraphics[width=0.4\textwidth,height=5cm]{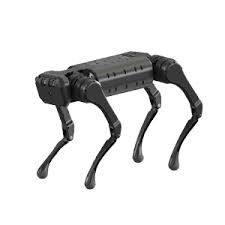}
\caption{\small{Unitree's A1 quardupedal robot.}}
\label{fig:A1Robot}
\end{figure}

Although the primary inspiration for developing this algorithm was to enable transfer of control policies for assistive devices, the proposed zero-shot transfer approach is applicable to any robot whose dynamics are partial observable due to challenges in measuring certain parameters. To validate the effectiveness of EAP algorithm on a real robotic system, we use the popular quadrupedal robot A1 from Unitree \cite{UnitreeA1} robotics, shown in \ref{fig:A1Robot}, as our testbed. A common, well established, approach while working with real robotic systems is to first perform a thorough system identification of the robot's kinematic and dynamic properties by collecting real world data. However, standard system identification methods fail when a robot is tested in an environment from which no prior data was available. For example, when data is collected on regular ground but the robot is tested on a ground with different contact parameters such as a soft foam mat, the resulting variation in dynamics is not captured by the identified parameters. As a result, control policies trained in simulation fail when tested on a novel environment in the real world. Using EAP, we demonstrate that a policy trained using our approach is able to overcome differences in the environments it is being tested in. We take the following approach to illustrate this :

\begin{enumerate}
    \item  We first perform a system identification of A1 robot parameters using data collected over carpeted ground.
    \item Then we define two tasks for the quadrupedal robot - In-place walking and walking forward. For each of these tasks, we define a set of observable and unobservable parameters and train EAP along with policies using baseline algorithms - UP and DR.
    \item We compare the performance of each of these policies on three novel environments from which no data for system identification was collected. First environment uses a soft foam mat as ground. Second, we add an additional unknown mass to the robot's torso. Third, we test the robot with additional mass using soft foam mat as ground.  
\end{enumerate}

\subsection{Data collection, simulation environment and system identification}

\subsubsection{Data collection}
The A1 robot is controlled using a position command which is converted into torques using a simple PD-controller.
It is equipped with joint angle sensors, Inertial-Measurement Unit (IMU) and a foot contact sensor. Using the joint angle sensors we compute both joint angle and joint velocities. We can also compute the robots torso's orientation and rate of change of orientation using the IMU sensor. 

The data collection process on the real A1 robot involves generating two kinds of data:

\begin{enumerate}
    \item We first suspend the robot in air, thus removing ground contact forces being applied to the robot, and apply a variety of sinusoidal and step commands to the robot. During this process, we make sure that the robots torso remains fixed as much as possible. We collect joint position and velocity responses from the robot.
    
    \item Next, we script simple sinusoidal commands with different frequencies and amplitudes that the robot legs can follow while being in contact with the ground (Note, we stick to simple commands because it is challenging to script commands for the robot in contact with the ground while ensuring safety). We collect joint position, velocities , the torso's orientation , angular velocity (computed using simple finite-difference method).
\end{enumerate}

We collect 60 trajectories on the robot each of length 10 seconds. The sensors are sampled at 30 Hz, which gives us a total of $60,000$ data points. These form dataset $\boldsymbol{D}$ , each trajectory $\boldsymbol{\tau}$ is consists of the command sent to the robot $\boldsymbol{a}$ and the state of the robot $\boldsymbol{s}$ (which includes the joint positions,velocities, torso position and orientation) for length of time $\boldsymbol{T}$.

\begin{align*}
    \boldsymbol{D = \{ (\tau_1,), (\tau_2) .. (\tau_n) \}} \\
    \boldsymbol{\tau_i = \{ (s_1,a_1),(s_2,a_2) ... (s_T,a_T) \}}
\end{align*}

\subsubsection{Simulation environment and system identification}

We use DART physics engines \cite{DART} to simulate the robot. In total, the robot has $18$ degrees of freedom, $12$ controllable joints of the legs and $6$ free degrees of freedom of the torso. The simulation runs at 500 Hz, however we use a control frequency of 20 Hz. In simulation, we enforce a torque limit of 30 $N/m$ based on the torque capabilities of the real robot. In our experiments with the real robot, we measured a sensor delay of 20-30 $ms$, we incorporate this sensor delay into our simulation framework as well. 
The feet of A1 robot are made from deformable rubber material which acts like a spring-damper when there is contact, to model this interaction we use hunt-crossley contact model \cite{hunt1975coefficient,seth2018opensim}. Hunt-crossley model comprises of parameters such as stiffness and friction which allows us to control the softness of the contact. To identify a simulation model from real world data,
we choose $8$ dynamical parameters : The proportional and derivative gains of the low-level controller $\boldsymbol{K_p,K_d}$ on the robot. DART also allows us to write custom low-level controllers (such as PD controllers), which enables users to design these controllers to compute the torques. Since, the PD controller gains affect the dynamic response of the robot joint legs, we include these parameters as optimization variables. We also include joint damping $\boldsymbol{\beta}$, contact stiffness $\boldsymbol{\kappa}$, ground friction $\boldsymbol{\gamma}$, mass of torso $\boldsymbol{m_{torso}}$, mass of thigh link $\boldsymbol{m_{thigh}}$ and mass of the calf link $\boldsymbol{m_{calf}}$. This makes the vector containing these variables $\boldsymbol{\eta \in R^{8}}$.

\begin{equation}
    \boldsymbol{\eta = \{ K_p, K_d, \beta,\kappa,\gamma,m_{torso},m_{thigh},m_{calf} \}}
\end{equation} 

For optimization, we use Covariance Matrix Adaptation (CMA-ES) \cite{Hansen16a} method to find the best set of parameter values in simulation that matches real world trajectories by minimizing the cost function described in \ref{eqn:cost}.

\begin{equation}
\label{eqn:cost}
    \boldsymbol{ J = \frac{1}{N}\min_{\eta} \sum_{n=1}^{N} \sum_{t=0}^{T} || s'_t - s_t||^2 }
\end{equation}

Where $s_t$ is the states collected in real world and $s'_t$ are the states observed in simulation.
The cost function minimizes the average error generated by $N$ state trajectories each of $T$ time steps in simulation when compared to real world data when the same actions are applied to both.  
Once the optimized parameters are obtained, we use this as the base environment in simulation.



\begin{table*}
\centering
\caption{Tasks and Network Architectures on the real robot
}
\resizebox{\columnwidth}{!}{%
\begin{tabular}{|l|l|l|l|l|}
\hline
Task & Observable Params. $\bm\mu$ & Unobservable Params. $\bm\nu$  & Net. Arch. & Err. Dim. $|\vc{e}_p|$         \\ \hline

In-Place Walking  & PD-gains, link masses & joint damping, contact params , delay &  (64,64,64) & 6         \\ \hline
Walking forward & PD gains, link masses & sensor delay, joint damping, contact params &  (64,64,64) & 6         \\ \hline
Walking forward w unknown mass & PD-gains  & joint damping, delay, torso mass &   (64,64,64) & 6     \\ \hline

\end{tabular}%
}

\label{tab:params_real}

\end{table*}

\section{In-place walking}

We define an in-place walking task for a quadrupedal robot in which the center of mass of the robot remains in the same place while the legs follow a walking motion. Inspired by the lower level control framework presented in \cite{Leeeabc5986}, we first define a end-effector reference trajectory for each leg and our action space is defined as delta end-effector cartesian positions from the pre-defined reference trajectory. We then use analytical inverse kinematics to compute the corresponding joint angles for each leg. For in-place walking, since there is no movement in the x-y plane of the robot, the reference trajectory only consists of motion in the z-axis which is perpendicular to the ground. The reference trajectory is defined in the equation below, where $A$ is the amplitude, $f$ is the frequency and $t$ is the time. When the cartesian $z$ is below zero or the ground plane the value just remains zero. 

$$
z = \begin{cases}
			A sin(2 \pi f t), & \text{if z $\geq$ 0}\\
            0, & \text{otherwise}
		 \end{cases}
$$

The action space is 12-dimensional ($R^{12}$) which comprises of residual cartesian position for each leg end-effector ($R^3$). So, the output of the neural network policy is added to the reference trajectory to compute the target leg positions. The state space consists of the angular velocity of torso $R^3$ computed using the gyroscope. The angular position $R^3$ of torso, joint angles $R^{12}$ and binary foot contact information for each leg which indicates if the foot is in contact or not $R^4$. An additional phase variable $\phi(t)$ is included as part of the state to indicate the phase of the gait cycle the robot is in. In total, the state vector is 23 dimensional $R^{23}$. The reward function , described in \ref{eqn:inplace} encourages the policy to generate actions that penalizes center of mass velocity of the robot while following the reference trajectory $\bar{q}$ of the end-effector of each leg. 

\begin{equation}
    r(\vc{s},\vc{a}) = w_1 e^{-k_{1}*(\vc{q} - \vc{\bar{q}})} + w_2 e^{-k_{2}*(\vc{\dot{c}})} - w_3 || a ||^{2}
\label{eqn:inplace}
\end{equation}

Where $w_1 = 0.5$, $w_2 = 0.5$ , $k_1 = 20$, $k_2 = 10$ and $w_3= 1e^{-3}$
During training, we consider the PD-gains , the mass of the robot links as observable parameters $\vc{\mu}$. In our preliminary experiments, we found that for simple tasks, these quantities can be identified by standard system identification procedures. However, quantities like joint damping, sensor delays and ground contact parameters are considered unobservable parameters $\vc{\nu}$.  We train EAP, UP and DR with these parameters as the set of observable and unobservable quantities.

The neural network policy is represented as a multi-layered perceptron (MLP) with three hidden layers of 64 neurons each. The error function is trained with a horizon length of five $T = 5$ and the projected error dimension is six $R^6$.


\section{Walking forward}

In this task, we train a policy to walk forward at a certain velocity while maintaining the yaw orientation. The framework for training the policy is similar to in-place walking task. A pre-defined reference trajectory is defined for each leg end-effector and a neural network policy outputs delta cartesian positions for each leg. The reference trajectories are defined as follows :

$$
z = \begin{cases}
			A sin(2 \pi f t - \frac{\pi}{2}), & \text{if z $\geq$ 0}\\
            0, & \text{otherwise}
		 \end{cases}
$$

$$
x = A sin(2 \pi f t)
,y = 0
$$

The reward function encourage forward walking at a velocity of $v=0.6 m/s$ while ensuring that the end-effectors follow the pre-defined reference trajectory. The reward function also encourages the policy to walk in a straight path by penalizing the yaw orientation $\psi$.

\begin{equation}
    r(\vc{s},\vc{a}) = w_1 e^{-k_{1}*(\vc{q} - \vc{\bar{q}})} + w_2 e^{-k_2 * \psi} +\min(\dot{x},0.6) - w_3 || a ||^{2}
\label{eqn:walkforward}
\end{equation}

Where $w_1 = 0.5$,$k_1 = 20$, $w_2=0.6$, $k_2=10$ and $w_3 = 1e^{-3}$. 
The state space, action space, observed, unobserved parameters and test environment remains the same as in-place walking task. The parameters of the policy and the error function remains the same as the previous task as well.

\section{Evaluation on real robot} 

We test policies on five environments : A training environment and four test environments. 
\begin{enumerate}
    \item Base training environment :  The same environment from which data for system identification was collected (robot moving on carpeted ground).
    \item Foam mat environment (T1): The robot is deployed on a soft foam mat with different contact parameters than over ground.
    \item Additional mass environment (T2):  Additional unknown mass is added to the robot's torso. 
    \item Foam mat with additional mass environment (T3):  Robot is tested with additional mass and deployed over a foam mat. 
\end{enumerate}

Owing to the difficulty in computing the rewards in the real world. We use a surrogate metric to evaluate the policies in the real world. For in-place walking task, we collect the joint angles when the policy is being deployed and compare how well the joint angles track the reference trajectory. This quantity is computed using the term $w_1 e^{-k_{1}*(\vc{q} - \vc{\bar{q}})}$ in equation \ref{eqn:inplace}. For walking forward task, we just use the straight line distance travelled by the robot as the surrogate metric, this is measured using a simple measuring tape.

\subsubsection{Performance in base training environment}

In the base training environment we notice that for both in-place walking as well as walking forward task all methods (EAP,DR and UP) show similar performance. The results are illustrated in \ref{fig:realworld_inplace} and \ref{fig:realworldwalk}  where the task performance for the base environment is similar. This corroborates with findings of prior work \cite{xie2020dynamics} which suggested that for simple tasks, carefully randomizing parameters is sufficient for sim-2-real transfer for quadrupedal robots. Since, the policies are being tested in the environment from which data for system identification was collected, the policies are able to transfer well. 

\subsubsection{Performance in test environments}

\begin{figure}
  \centering
  \includegraphics[width=0.5\columnwidth]{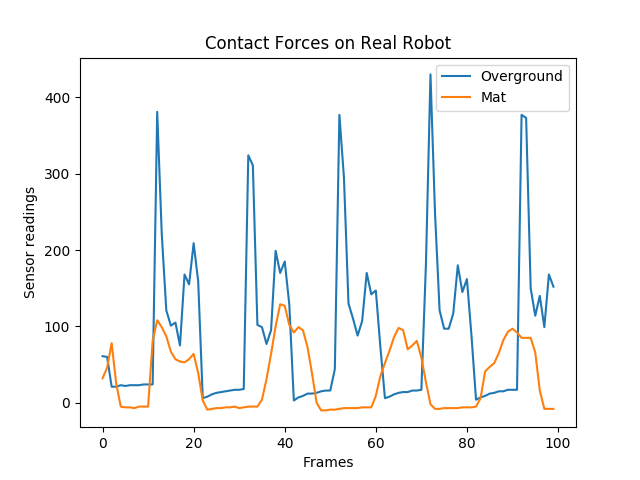}
  \vspace*{-5mm}
  \caption{Comparison of contact profile generated by ground and soft foam mat}
  \label{fig:groundvsmat}
\end{figure}

In our first test environment (T1), we change the surface on which the robot performs the task, with the aim of changing the contact parameter values. The differences in the contact profile generated by the two surfaces is illustrated in figure \ref{fig:groundvsmat}.
We notice that for in-place walking task, all methods perform similarly \ref{fig:realworld_inplace}. However, in the walking forward task \ref{fig:realworldwalk}, EAP beats both the baselines DR and UP by walking forward for a longer straight-line distance on the mat with an average of $2.3$m whereas both DR and UP move less than $2$m. We notice that both DR and UP tend to change the yaw-orientation of the robot, which is not a desired behaviour.

In test environment 2 (T2), we attach a mass of 7.5 $lbs$ (unknown to the policy parameters) to the robot's torso. In this environment, the ability for EAP to adapt is best illustrated, beating both baselines DR and UP in straight line distance travelled on carpeted ground. With EAP the robot moves an average distance of $1.7$m , UP and DR manage close $1$m. Results illustrated in \ref{fig:realworldwalk}.

The performance of the policies in test environment 3 (T3) are similar to that of T2 with EAP managing to travel the furthest distance compared to baselines DR and UP.

\begin{figure*}[t!]
    \centering
    \begin{subfigure}[t]{0.5\textwidth}
        \centering
        \includegraphics[height=2.2in]{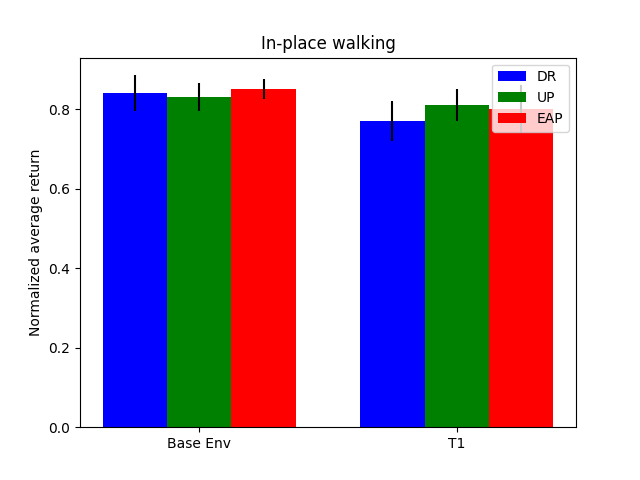}
        \caption{In-place walking task}
        \label{fig:realworld_inplace}
    \end{subfigure}%
    ~ 
    \begin{subfigure}[t]{0.5\textwidth}
        \centering
        \includegraphics[height=2.2in]{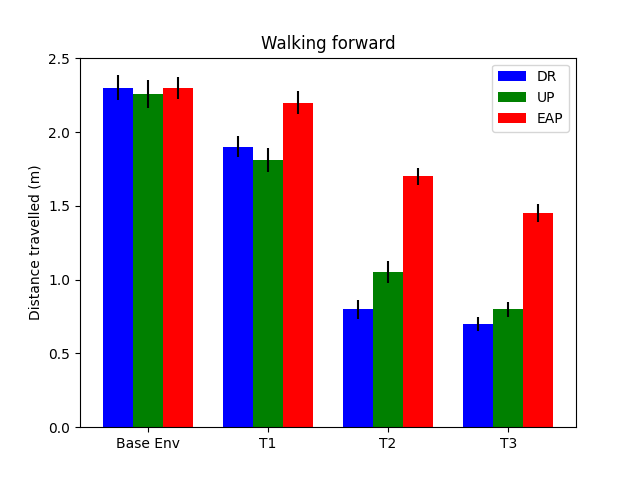}
        \caption{Walking forward task}
        \label{fig:realworldwalk}
    \end{subfigure}
    \caption{Evaluation of real world results.}
\end{figure*}

\section{Conclusion}

We presented a novel approach to train an error-aware policy (EAP) that transfers effectively to unseen target environments in a zero-shot manner. Our method learns an EAP for an assistive wearable device to help a human recover balance after an external push is applied. We show that a single trained EAP is able to assist different human agents with unseen biomechanical characteristics. We also validate our approach by comparing EAP to common baselines like Universal Policy and Domain randomization to show our hypothesis that a policy which explicitly takes future state error as input can enable better decision making. Our approach outperforms the baselines in all the tasks. We also evaluated the performance of our algorithm through a series of ablation studies that sheds some light on the importance of parameters such as error horizon length, error representation, choice of observable parameters and choice of reference dynamics. We find that EAP is not sensitive to either the choice of observable parameters or the reference dynamics, and outperforms the baselines with variations in these quantities as well. Further, we validated the ability of EAP to transfer to real world scenario using A1 quadrupedal robot. The preliminary results presented for simple tasks such as in-place walking and walking forward shows promise of applying the proposed approach to real world scenarios.

Our work has a few limitations. At the core, our algorithm relies on the error function to make predictions of the expected state errors. The accuracy of this prediction can be improved by better function approximators such as recurrent neural networks (RNN) that takes a history of states as input, we leave this for future work.

    \chapter{Conclusion and future work}

In conclusion, we presented of a set of learning based algorithms that address the important challenge of safety in bipedal locomotion. Our contributions are along two closely related directions 1) Fall prevention and safe falling for bipedal robots and 2) Fall prevention for humans during walking using assistive devices. With bipedal robots, we showed the effectiveness of reinforcement learning (RL) based approaches to learn control policies that demonstrate a wide range of falling and balancing strategies. Our algorithmic contributions also included improving efficiency of learning by incorporating abstract dynamical models as priors, curriculum learning, imitation learning and a novel method of building a graph of policies into the standard RL framework. In the chapters 4 and 5 of the thesis, we shifted focus towards an important healthcare related topic of fall prevention for humans using assistive devices. To this end, our contributions include using imitation learning approach to create virtual human walking agents whose bio-mechanical gait characteristics are similar to real world humans. Using these simulated agents we showed that, assistive devices could indeed help prevent falls when pushed whereas human agents without assistants fail. Finally, since humans exhibit a wide range of variations in gait characteristics, we developed a novel algorithm to enable zero-shot generalization of fall prevention policies when tested on new human subjects. Although inspired by assistive devices, the algorithm is applicable for any robotic system and we validated the proposed approach on a real world quadrupedal A1 robot.

\section{Safe locomotion for bipedal robots}

In chapter 3, we developed an algorithm to generate safe falling motion for a bipedal robot. Unlike steady state walking, which is periodic in nature, optimizing for a falling motion is challenging because it involves reasoning about discrete contact sequence planning (which body part should hit the ground and in what sequence) as well a continuous joint motion to minimize impact upon ground contact. We showed that we could solve this problem using an offline policy learning algorithm. Our approach significantly improved on the computation time required by prior dynamic programming methods while also ensuring that a wide variety of falling strategies naturally emerge from the algorithm. Towards the same goal of safety for bipedal locomotion, inspired by strategies humans use to recover balance as a response to external pushes, in our second work we presented a training methodology for a control policy which learns residual control signals to those generated by traditional model-based controllers to maintain balance. By learning residual control signals, we not just simplify learning the task because the model-based controllers act as strong priors, but also improve upon methods that use either just model-based or naive RL approaches. Our algorithm also included a novel sampling method that adaptively adjusts the difficulty in training samples to encourage efficient learning. By doing so, we achieved better performance at a lower sample cost compared to baseline naive RL approach. 

We also addressed the common issue of sample inefficiency while solving complex task such as locomotion with model free reinforcement learning algorithms. RL has shown promising results in learning complex motor skills in high dimensional systems, however the control policies typically perform well only from a small set of initial states and take millions of samples to complete the task. In the final section of chapter 3, we take a divide and conquer approach to solve this by breaking down a complex task into multiple sub-tasks and learn a directed-graph of control policies to achieve the same task but by consuming less samples. The edges of our directed graph contains policies and the nodes contain state distributions, so a policy can take the robotic system from the one set of states to another set of goal states via rolling out policy/policies. Starting from the first policy that attempts to achieve the task from a small set of initial states, the algorithm automatically discovers the next subtask with increasingly more difficult initial states until the last subtask matches the initial state distribution of the original task. We showed that our approach takes lesser samples than common baselines such as naive RL, curriculum learning, manually generated sub-tasks while ensuring that the task can be completed from a wider range of initial states.

\subsubsection{Future work}

The limitation with our work on push recovery, be it either fall prevention or safe falling, is that the control policies work best when the robot's initial center of mass velocity is close to zero when pushed. However, in real world scenarios the ability to recover when the robot's motion is more dynamic, such as walking or running, would be crucial to ensure safety under all circumstances. Developing algorithms that can ensure safety for a wider range of scenarios would be an interesting extension to our work.  

\section{Fall prevention using assistive devices}

In chapter 4, we shifted our focus towards learning fall prevention control policies for assistive devices. We developed an approach to automate the process of augmenting an assistive walking device with the ability to prevent falls. Our method has three key components : A human walking policy, fall predictor and a recovery policy. In a simulated environment we showed that an assistive device can indeed help recover balance from a wider range of external perturbations. We introduced stability region as a quantitative metric to show the benefit of using a recovery policy, larger area of stability region would imply better recovery motion. In addition to this, stability region can also be used to analyze different sensor and actuator design choices for an assistive device. We evaluated the stability region of six different policies using various sensor and actuator configurations to shed some light on appropriate design choices to help develop effective assistive devices.

A significant bottleneck in making assistive devices more ubiquitous in society is due to the fact that humans exhibit remarkable variations in gait characteristics, making design of control policies that generalize for a large population extremely challenging. A common approach has been to tune controllers for each individual user, however this can be tedious and a time consuming process. We addressed this issue by viewing it through the lens of transfer learning problem. We developed a novel algorithm that enables a policy to transfer in a zero-shot manner for partially observed dynamical systems like humans. Our work introduces an Error-aware policy (EAP) which explicitly takes as input a predicted state error generated in the target environment to produce a corrected action that successfully completes the task. We show that a single trained EAP is able to assist different human agents with unseen biomechanical characteristics. We also validate our approach by comparing EAP to common baselines like Universal Policy (UP) and Domain randomization (DR) to show our hypothesis that a policy which explicitly takes future state error as input can enable better decision making. Our approach outperforms the baselines in all the tasks. We also evaluated the performance of our algorithm on a real world quadrupedal robot, which strengthens the case for applicability of EAP on a real system. We show that for tasks such as walking forward, EAP is able to outperform baselines such as DR and UP when tested on novel environments with unique ground contact characteristics.

\subsubsection{Future work}

A key component of our work with assistive devices is virtual human walking agent. Simulated agents that generate bio-mechanically accurate motion  \cite{falisse2020physics,santos2021predictive,weng2021natural} can play a crucial role in not just improving our understanding of human motion but also help in developing better assistive and rehabilitation strategies for people with disability. Algorithms that can generate such human-like motion for not just periodic steady state walking gaits, but for more dynamic motion such as running, push-recovery, slip/ trip recovery can be a very exciting research direction. Incorporating musculo-tendon human models could also be a contributing factor in closing the gap between real world and simulated human agents. The major bottleneck for such applications is the shortage of real-world data for validation purposes, recent advancements in machine learning methods to learn from limited data as well as a collaborative efforts between researchers working in bio-mechanics and machine learning can help overcome some of these challenges. 

In this work, our primary contributions have been on developing control policies for robots, however in the spirit of the famous phrase "form follows function", co-optimization of robot design along with policy learning can be a powerful approach to create functional robots in the future. In chapter 4, we provided some preliminary results by shedding light on the best sensor and actuator design choices for assistive devices based on the policy's performance. Extending this to optimize for robot structural design can be an interesting direction.  

Another closely related topic to our work with assistive devices is modelling human-robot interaction. In our current, work we assume that the human perfectly adapts to the forces applied by the exoskeleton. However, in reality this does not necessarily hold true. Better predictive modelling of the interaction between the robot and the human is a promising research direction. Modelling human intention while collaborating with a robot could also be key towards wide-spread use of these robots. Contributions in this area can have a lasting impact not just with assistive devices but also numerous other applications such as human-robot collaboration in a factory setting, self-driving cars and robots that are built for home environments. 

As highlighted in Chapter 5, sim-2-real transfer of policies has gained a lot of interest in the research community. However, the primary focus from researchers has been through the lens of algorithmic contributions, an exciting future direction will be to develop methods that improves simulation environments by bringing them closer to reality. This could be done through leveraging machine learning methods to improve contact models, focusing efforts on differentiable simulators to enable end-to-end learning or by improving computational efficiency of more traditional methods such as finite-element methods (FEM) to improve accuracy of simulation, especially for deformable bodies.   


    \makeBibliography

\begin{vita}

Visak Kumar was born in Bangalore, India in 1991. He graduated from National Public School in 2009 and attended M.S.Ramaiah Institute of Technology from 2009 to 2013 where he majored in Mechanical Engineering. After completing undergraduate degree, Visak worked at Indian Institute of Science, Bangalore, under supervision of Prof. Ananthasuresh from 2013 - 2014, as a research assistant where he worked on developing a micro-scale mechanical pump device for drug delivery.

Visak obtained his Masters degree from University of Washington, Seattle in 2016. After obtaining his masters degree, he started PhD program in robotics at Georgia Tech, under supervision of Dr. Karen Liu working on developing learning algorithms for robot control. In 2017, Visak worked as a research intern at Disney Research, Pittsburgh under supervision of Dr. Katsu Yamane and Dr. Sehoon Ha. In 2019, Visak interned at Nvidia Research, Seattle under supervision of Dr. Stan Birchfield and Dr. Jonathan Tremblay working on dextrous manipulation. He returned as an intern to Nvidia in summer of 2020.

\end{vita}
\end{thesisbody}

\end{document}